\documentclass[journal]{IEEEtran}

\usepackage[T1]{fontenc}
\usepackage[utf8]{inputenc}
\usepackage{lmodern}
\usepackage{microtype}

\usepackage{amsmath,amssymb,amsthm,amsfonts}
\usepackage{mathtools}
\usepackage{bm}

\usepackage{graphicx}
\usepackage{tikz}
\usetikzlibrary{positioning,arrows.meta,shapes.geometric,fit,calc,trees,backgrounds}
\usepackage{booktabs}
\usepackage{array}
\usepackage{multirow}
\usepackage{makecell}
\usepackage{tabularx}
\usepackage{adjustbox}

\newcommand{\xmark}{\ensuremath{\times}}

\usepackage[hidelinks,breaklinks=true,hypertexnames=false]{hyperref}
\usepackage{orcidlink}
\usepackage{url}
\usepackage{cite}
\usepackage{algorithm}
\usepackage{algpseudocode}
\usepackage{enumitem}

\newtheorem{definition}{Definition}

\theoremstyle{definition}
\newtheorem{axiom}{Axiom}

\newtheorem{result}{Result}
\theoremstyle{remark}
\newtheorem{remark}{Remark}
\newtheorem{principle}{Principle}

\newcommand{\R}{\mathbb{R}}
\newcommand{\E}{\mathbb{E}}
\newcommand{\Prob}{\mathbb{P}}
\newcommand{\indicator}{\mathbf{1}}
\newcommand{\featureset}{N}
\newcommand{\coalition}{S}
\newcommand{\baseline}{x'}
\newcommand{\attribution}{\bm{\phi}}
\newcommand{\valuefn}{v}
\newcommand{\mask}{\bm{z}}

\newcommand{\model}{f}

\DeclareMathOperator{\IG}{IG}

\title{Local Additive Feature Attribution:\\
A Mathematical Taxonomy and Reporting Checklist}

\author{Rebecca~Afriyie~Sarpong\,\orcidlink{0009-0001-7530-926X}
and Daniel~Commey\,\orcidlink{0000-0001-5759-918X}%
}

\begin{document}

\maketitle

\begin{abstract}
Feature-attribution methods are central to explainable artificial
intelligence. Their assumptions are expressed in several mathematical
languages: cooperative-game values, path integrals,
gradient operators, perturbation distributions, and backpropagation
rules. This survey proposes a common framework for local additive
feature attribution. It organizes Shapley, path-based,
gradient/backpropagation, perturbation, and CAM-style methods around
five specification choices: value function, reference, path,
perturbation distribution, and conservation rule. It then compares
these methods through an axiom-by-method matrix and links common
failure modes, including baseline sensitivity, off-manifold
perturbations, sanity-check failures, adversarial manipulation, and
method disagreement, to the assumptions that produce them. Finally,
the survey proposes a ten-item reporting checklist for studies that
use local additive attributions. The central message is that
attribution results are meaningful only relative to the mathematical
assumptions under which they are defined, and that those assumptions
should be reported.
\end{abstract}


\section{Introduction}
\label{sec:introduction}

\IEEEPARstart{F}{eature} attribution has become one of the dominant
interfaces between complex predictive models and their human users.
Faced with a model that maps a $d$-dimensional input $x \in \R^d$ to a prediction
$f(x) \in \R$, a user asks: \emph{which features mattered, and by how
much?} A feature-attribution method answers with a vector
$\attribution(f,x) \in \R^d$ that distributes credit (or blame) for the
prediction among the input features. Over the last decade, dozens of such
methods have been proposed, including Integrated Gradients~\cite{sundararajan2017axiomatic},
SHAP~\cite{lundberg2017unified}, LIME~\cite{ribeiro2016lime},
Grad-CAM~\cite{selvaraju2017gradcam}, LRP~\cite{bach2015lrp},
DeepLIFT~\cite{shrikumar2017deeplift}, and many others, each motivated
by a different intuition about what an explanation should be.

The rapid growth of attribution methods has not been matched by
comparable clarity about their assumptions. Two
methods applied to the same prediction frequently produce explanations
that agree only on coarse features and disagree, sometimes sharply, on
their relative ordering. Krishna et al.~\cite{krishna2024disagreement}
documented this phenomenon at scale across six attribution
methods and four tasks, the median Spearman rank correlation between
explanations of the same prediction was below $0.5$. They termed it
the \emph{disagreement problem}. Bilodeau et al.~\cite{bilodeau2024impossibility}
established a complementary negative result: no feature-attribution
method can simultaneously satisfy a small set of reasonable
desiderata, so disagreement among methods is mathematically
unavoidable. Earlier critiques along similar lines
include~\cite{adebayo2018sanity,kindermans2019unreliability}.
Empirical evaluations on the same benchmarks reach incompatible
conclusions about which method is most ``faithful''~\cite{tomsett2020sanity,
hooker2019roar}. And it is now well documented that adversarial
manipulations can produce arbitrary attribution maps with negligible
changes to the prediction~\cite{dombrowski2019manipulated,ghorbani2019fragile,
slack2020fooling}. These pathologies often trace to \emph{implicit
mathematical choices} that different methods make and that users
rarely see.

\begin{figure}[t]
\centering
\begin{tikzpicture}[font=\footnotesize,
  box/.style={draw=black!60, rounded corners=2pt, align=center,
    inner sep=4pt, minimum height=7mm},
  spec/.style={box, fill=blue!7, text width=70mm},
  io/.style={box, fill=gray!8},
  outc/.style={box, fill=orange!10, text width=33mm},
  rep/.style={box, fill=teal!8, text width=70mm},
  arrow/.style={-{Latex[length=2mm]}, semithick, black!70}
]
\node[io] (model) at (0,0) {Model $\model$, input $x$, scalar output $\model_c(x)$};
\node[io, below=5mm of model] (attr)
  {Attribution method $\attribution(\model, x, \baseline) \in \R^d$};
\node[spec, below=5mm of attr] (spec) {\textbf{Hidden specification choices}\\[0.8mm]
  value function $\valuefn$ \,$\cdot$\, reference $\baseline$
  \,$\cdot$\, path $\gamma$\\
  perturbation distribution $p_{\mask}$ \,$\cdot$\, conservation rule};
\node[outc, anchor=north west] (ax)
  at ($(spec.south west)+(0,-0.5)$)
  {Axioms satisfied\\ {\scriptsize (Sec.~\ref{sec:axioms}, Tab.~\ref{tab:axiom-matrix})}};
\node[outc, anchor=north east] (fail)
  at ($(spec.south east)+(0,-0.5)$)
  {Failure modes\\ {\scriptsize (Sec.~\ref{sec:failure})}};
\node[rep, anchor=north] (rep)
  at ($(ax.south)!0.5!(fail.south)+(0,-0.5)$)
  {Reporting items R1--R10 {\scriptsize (Sec.~\ref{sec:reporting})}};
\draw[arrow] (model) -- (attr);
\draw[arrow] (attr) -- node[right, font=\scriptsize] {determined by} (spec);
\draw[arrow] (ax.north -| ax.center) ++(0,0.5) -- (ax.north);
\draw[arrow] (fail.north -| fail.center) ++(0,0.5) -- (fail.north);
\draw[arrow] (ax.south) -- (ax.south |- rep.north);
\draw[arrow] (fail.south) -- (fail.south |- rep.north);
\end{tikzpicture}
\caption{Attribution methods as specifications of hidden mathematical
choices. A local additive attribution is determined by five
specification choices (value function, reference, path, perturbation
distribution, conservation rule); these choices fix which axioms the
method satisfies and which failure modes it is exposed to, and the
proposed reporting checklist asks studies to state them.}
\label{fig:overview}
\end{figure}
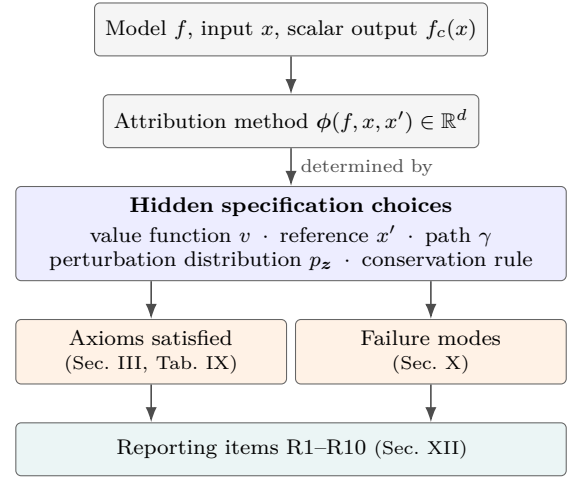

This survey uses those mathematical choices as its organizing principle
and compares feature-attribution methods through their underlying
mathematical objects: the \emph{value function}~$v$ that defines
``feature presence'', the \emph{baseline} or reference~$\baseline$
against which contributions are measured, the \emph{path}~$\gamma$ along
which integration occurs, the \emph{perturbation distribution} that
defines local linearity, and the \emph{conservation rule} that
distributes a quantity through a network. Many important disagreements
and failure modes can be traced to differences in one or more of these
objects. Figure~\ref{fig:overview} summarizes how this framing
organizes the paper.

\begin{table}[t]
\centering
\caption{Five mathematical choices used as the organizing frame for
the survey.}
\label{tab:five-axis-preview}
\footnotesize
\begin{tabularx}{\columnwidth}{@{}p{0.30\columnwidth}p{0.34\columnwidth}X@{}}
\toprule
Object & Question answered & Examples affected \\
\midrule
Value function & What does feature absence mean? & SHAP, KernelSHAP, TreeSHAP \\
Reference & Compared to what? & IG, DeepLIFT, GradientSHAP \\
Path & Along what trajectory? & IG, Guided IG, Blur IG \\
Perturbation distribution & Which neighbourhood is local? & LIME, occlusion, RISE \\
Conservation rule & What quantity is propagated? & LRP, DeepLIFT \\
\bottomrule
\end{tabularx}
\end{table}

\subsection{Why Axioms?}

A second organizing principle of this survey is the use of
\emph{axioms} as the primary tool of comparison. Axioms are precise
properties an attribution method may or may not satisfy: completeness
(the attributions sum to the prediction-minus-baseline), implementation
invariance (functionally equivalent models receive identical
attributions), sensitivity (a feature that changes the prediction must
receive nonzero attribution), symmetry, dummy, and others. We adopt the
axiomatic perspective for three reasons:

\begin{enumerate}[leftmargin=*]
\item \textbf{Axioms are model-agnostic.} A claim like ``Integrated
Gradients satisfies completeness'' is a theorem about the method, not a
hypothesis about a particular dataset or architecture. This makes axioms
the right invariants for a survey: they remain true across
benchmarks~\cite{sundararajan2017axiomatic,lundstrom2025ig}.

\item \textbf{Axioms expose disagreement at its source.} When two methods
produce different explanations, the question ``which is correct?'' is
usually ill-posed; it amounts to asking which set of axioms one
prefers~\cite{sundararajan2020manyshap,kumar2020problems}. The
function-approximation perspective of Han
et al.~\cite{han2022which} and the impossibility result of Bilodeau
et al.~\cite{bilodeau2024impossibility} make the trade-off explicit:
choosing a method is choosing which axioms to retain. Making the
axioms explicit converts an unresolvable empirical dispute into a
modeling decision.

\item \textbf{Axioms enable equivalence and reduction theorems.}
Several apparently distinct methods can be shown to coincide under
appropriate axiomatizations. KernelSHAP, exact Shapley, and certain
forms of weighted linear regression compute the same
quantity~\cite{lundberg2017unified,covert2021kernelshap}. DeepLIFT with
the Rescale rule converges to Integrated Gradients in the limit of small
input increments~\cite{ancona2018gradient}. These equivalences are
invisible from a heatmap-comparison perspective; they appear only
through the axioms.
\end{enumerate}

\subsection{Attribution vs.\ Interpretability, Explanation, and Causality}

The terms \emph{interpretability}, \emph{explanation}, and \emph{attribution}
are often used interchangeably in the XAI literature, but the underlying
objects are mathematically distinct. \emph{Interpretability} is a
property of a model, defined here as the extent to which a human can predict, audit,
or modify its behaviour from its structure
alone~\cite{lipton2018mythos,doshivelez2017rigorous}. \emph{Explanation}
is a broader category that includes example-based, counterfactual,
concept-based, and rule-based justifications. \emph{Attribution} is the
narrow problem of decomposing a single prediction into per-feature
contributions, typically as a real-valued vector. This survey concerns
the third object: \emph{local}, \emph{additive}, \emph{post-hoc} feature
attribution for differentiable and tree-structured models.

We further distinguish attribution from \emph{causal} feature
analysis. A causal feature effect asks how the prediction would change
under an external intervention on a feature, in the sense of Pearl's
do-calculus. Most attribution methods, including all
gradient-based methods, compute associational quantities relative to
a chosen reference distribution. They become causal only under
assumptions about the data-generating process that are rarely stated and
even more rarely satisfied~\cite{janzing2020causal,frye2020manifold}.
Conflating the two is a major source of misinterpretation in
high-stakes domains, and we return to it in
Section~\ref{sec:failure}.

\subsection{Scope}

\textbf{Included.} Local, post-hoc feature attribution for predictive
models. This covers Shapley-value methods, path-integral methods,
gradient and backpropagation methods, perturbation-based and
surrogate methods, and CAM-style visual attribution insofar as it is
derived from gradients. We treat attribution as a mathematical object
and devote substantial space to axioms, value functions, baselines,
paths, and conservation rules. We include the evaluation theory of
faithfulness, infidelity, sanity checks, and ROAR-style retraining that
has emerged in parallel with the methods themselves.

\textbf{Excluded or briefly treated.} Inherently interpretable models
(linear models, decision lists, rule sets), global rule extraction,
counterfactual explanations, example-based explanations, mechanistic
interpretability, concept-bottleneck models, and visualization tools
without attribution semantics. We comment on the boundary with concept
methods (TCAV~\cite{kim2018tcav}, network
dissection~\cite{bau2017networkdissection}) where useful, but treat them
as a complementary research programme outside the attribution methods
surveyed here.

\subsection{Position Relative to Prior Surveys}
\label{sec:prior-surveys}

Several broad XAI surveys preceded this one, including the
foundational surveys of Guidotti et al.~\cite{guidotti2018survey},
Adadi and Berrada~\cite{adadi2018peeking}, Gilpin
et al.~\cite{gilpin2018explaining}, the book-length treatment of
Molnar~\cite{molnar2022iml}, the global-interpretation survey of
Saleem et al.~\cite{saleem2022global}, the gradient-method technical
review of Wang et al.~\cite{wang2024gradientsurvey}, the additive
feature-attribution review of
Cremades et al.~\cite{cremades2024additive}, the Shapley-specific
survey of Li et al.~\cite{li2024shapleysurvey}, and the systematic
evaluation review of Nauta et al.~\cite{nauta2023evalsurvey}.
Table~\ref{tab:related-surveys} compares this paper against those
references along eight dimensions. The present survey differs from this
literature in two respects:

\begin{enumerate}[leftmargin=*]
\item Existing surveys have treated several of these families in
depth, but the cross-family axiom structure remains fragmented. This
article jointly cross-references Shapley, path-based,
gradient/backpropagation, and perturbation methods through a single
\emph{axiom-by-method matrix}. Existing surveys are either family-specific
(Shapley~\cite{li2024shapleysurvey},
gradient~\cite{wang2024gradientsurvey},
additive~\cite{cremades2024additive}) or
breadth-prioritised~\cite{guidotti2018survey,adadi2018peeking,gilpin2018explaining}.
\item The assumption sensitivity of attribution methods has not yet
been consolidated into a reporting checklist for papers that use local
additive attributions. This paper proposes such a checklist in
Section~\ref{sec:reporting} and maps it to the analytic corpus in
Appendix~\ref{app:compliance-audit}.
\end{enumerate}

\paragraph{Comparison criteria for Table~\ref{tab:related-surveys}.}
We assigned \checkmark{} when the surveyed reference dedicates at
least one section, subsection, or comparable structural unit to the
column dimension; $\circ$ when the dimension is treated in
fewer than two paragraphs, in an appendix, or only as part of a
broader framing; and blank when the dimension is not addressed.
The table is a scope comparison, not a bibliometric ranking. Ambiguous
cases are scored in favour of the prior survey, and the interpretation
of each mark is recorded in the supplementary scoring sheet.

\begin{table*}[t]
\centering
\caption{Position of this survey against prior XAI/attribution
surveys. \checkmark = covered substantively; $\circ$ = brief or
partial; blank = not covered. ``Shap.'' Shapley methods; ``IG'' path
methods; ``CAM'' Grad-CAM family; ``Axiom matrix'' an explicit
axiom-by-method cross reference; ``VF tax.'' value-function
taxonomy; ``Path/baseline tax.'' baseline and path taxonomy;
``Eval/Failure'' formal evaluation and failure-mode
treatment; ``Checklist'' reporting checklist as a deliverable. The
final row records the intended scope of the present work, not an
independent quality ranking.}
\label{tab:related-surveys}
\small
\begin{tabular}{@{}lcccccccc@{}}
\toprule
Survey & Shap. & IG & CAM & Axiom matrix & VF tax. & Path/baseline tax. & Eval/Failure & Checklist \\
\midrule
Guidotti et al.~2018~\cite{guidotti2018survey} & $\circ$ & $\circ$ & & & & & $\circ$ & \\
Adadi \& Berrada~2018~\cite{adadi2018peeking} & $\circ$ & $\circ$ & $\circ$ & & & & $\circ$ & \\
Gilpin et al.~2018~\cite{gilpin2018explaining} & $\circ$ & $\circ$ & $\circ$ & & & & & \\
Molnar~2022~\cite{molnar2022iml} & \checkmark & \checkmark & $\circ$ & & $\circ$ & $\circ$ & $\circ$ & \\
Linardatos et al.~2021~\cite{linardatos2021review} & $\circ$ & $\circ$ & $\circ$ & & & & $\circ$ & \\
Saleem et al.~2022~\cite{saleem2022global} & $\circ$ & & & & & & $\circ$ & \\
Nauta et al.~2023~\cite{nauta2023evalsurvey} & $\circ$ & $\circ$ & $\circ$ & & & & \checkmark & \\
Li et al.~2024~\cite{li2024shapleysurvey} & \checkmark & & & & $\circ$ & & $\circ$ & \\
Wang et al.~2024~\cite{wang2024gradientsurvey} & & $\circ$ & $\circ$ & & & $\circ$ & $\circ$ & \\
Cremades et al.~2024~\cite{cremades2024additive} & \checkmark & \checkmark & & & $\circ$ & $\circ$ & $\circ$ & \\
\midrule
This work & \checkmark & \checkmark & \checkmark & \checkmark & \checkmark & \checkmark & \checkmark & \checkmark \\
\bottomrule
\end{tabular}
\end{table*}

\subsection{Survey Methodology}
\label{sec:methodology}

This paper is a structured narrative survey and taxonomy-driven
review. The search aimed to assemble
the methodological lineage needed to compare local attribution
operators mathematically; estimating the size of the XAI literature
or achieving exhaustive bibliometric coverage was out of scope. The
analytic corpus contains 105 unique cited works spanning
method-defining papers, axiomatic characterizations, evaluation
studies, failure-mode analyses, boundary cases, and prior surveys.
Papers were selected for their role in the taxonomy: defining an
attribution operator, stating an axiom or reduction, evaluating
attribution behaviour, documenting a failure mode, or clarifying the
boundary between attribution and adjacent forms of explanation.

\paragraph{Databases.} Google Scholar, Semantic Scholar (S2-API),
the ACL Anthology, and the proceedings indices of NeurIPS, ICML,
ICLR, AAAI, IJCAI, ECCV, ICCV, CVPR, ACL, EMNLP, NAACL, and KDD.

\paragraph{Time range.} Searches were conducted between January and
April 2025. The planned publication window ran from 1953 (Shapley's
$n$-person-games paper) through the end of 2024. We also included the
directly relevant 2025 axiomatic characterization of Integrated
Gradients by Lundstrom and Razaviyayn~\cite{lundstrom2025ig}, identified
during final verification.

\paragraph{Query strings.} We ran the following queries verbatim on
each database; the conjunctions were taken as Boolean AND, the
disjunctions as Boolean OR.
\begin{itemize}[leftmargin=*]
\item \texttt{("feature attribution" OR "Shapley value" OR
"Integrated Gradients" OR LRP OR DeepLIFT OR Grad-CAM OR SHAP OR
LIME) AND (axiom OR axiomatic OR completeness OR consistency)}
\item \texttt{("Aumann-Shapley" OR Banzhaf OR "Owen value" OR
"quantitative input influence" OR "game theory") AND
("feature attribution" OR "individual prediction" OR explanation)}
\item \texttt{("attribution" OR "saliency") AND (faithfulness OR
infidelity OR "sanity check" OR ROAR OR insertion-deletion)}
\item \texttt{("attribution" OR "explanation") AND
(adversarial OR fragile OR manipulation OR Lipschitz)}
\end{itemize}

\paragraph{Selection process.} Candidate records were deduplicated by
title, DOI, and arXiv identifier where available. Papers were retained
when they satisfied at least one of the following roles: (i) introduced
a method included in the taxonomy, (ii) supplied an axiomatic
characterization or reduction used in the comparison matrix,
(iii) proposed an evaluation metric or benchmark used in
Section~\ref{sec:evaluation}, (iv) documented a failure mode used in
Section~\ref{sec:failure}, or (v) provided a prior survey against which
the present article is positioned. Historical sources were retained
when they are needed for the mathematical genealogy of Shapley values,
cooperative-game attribution, or early saliency methods.

\begin{table}[t]
\centering
\caption{Review corpus summary. Role counts are non-exclusive because
one cited paper may introduce a method, state an axiom, and provide an
evaluation result.}
\label{tab:corpus-summary}
\footnotesize
\begin{tabularx}{\columnwidth}{@{}Xr@{}}
\toprule
Role in analytic corpus & Included papers \\
\midrule
Method-introducing papers & 56 \\
Axiomatic characterization papers & 26 \\
Evaluation / benchmark papers & 20 \\
Failure-mode papers & 15 \\
Prior surveys & 10 \\
Boundary-case papers & 15 \\
\midrule
Total unique cited papers in analytic corpus & 105 \\
\bottomrule
\end{tabularx}
\end{table}

\paragraph{Boundary and exclusion criteria.} The primary corpus is
restricted to local additive attribution for predictive models.
Global-interpretability methods, counterfactual and example-based
explanations, concept-based explanations, and mechanistic
interpretability are treated only when they clarify a boundary case
for the taxonomy. Duplicate versions of the same work were represented
by the most complete or most widely cited version.

\paragraph{Corpus construction.} The corpus is organized by analytic
role. This choice is appropriate
for a taxonomy whose unit of comparison is the attribution operator
and its mathematical assumptions. Reproducibility is supported through the
query strings, inclusion roles, boundary criteria, and the final
analytic corpus underlying the tables. The corpus emphasizes
English-language sources from major machine-learning, computer-vision,
natural-language-processing, and AI venues, together with foundational
mathematical sources required for the Shapley and axiomatic lineage.

\subsection{Contributions and Roadmap}

The paper contributes three concrete artifacts: a taxonomy, an axiom
matrix, and a reporting checklist.

\begin{enumerate}[leftmargin=*]
\item \textbf{A unified mathematical taxonomy
(Sections~\ref{sec:formulation}--\ref{sec:perturbation}).}
Section~\ref{sec:formulation} fixes a common notation for $f$, $x$,
$\baseline$, the coalition algebra over $\featureset =
\{1,\dots,d\}$, value functions $\valuefn$, masks $\mask$, and paths
$\gamma(\alpha)$. Section~\ref{sec:axioms} catalogues the axioms
under which different methods are characterized.
Sections~\ref{sec:shapley}--\ref{sec:perturbation} survey the four
method families in this common frame.

\item \textbf{An axiom-by-method matrix and failure-mode formalization
(Sections~\ref{sec:comparison}--\ref{sec:failure}).}
Section~\ref{sec:comparison} presents the main comparison table of
the paper, the axiom-by-method matrix
(Table~\ref{tab:axiom-matrix}), together with the complexity
landscape and several known reductions between methods (KernelSHAP
and exact Shapley in
expectation~\cite{lundberg2017unified,covert2021kernelshap}; DeepLIFT
and Integrated Gradients in the small-increment
limit~\cite{ancona2018gradient}; Grad-CAM and class-conditional
gradient projections~\cite{selvaraju2017gradcam}; LRP-$\epsilon$ and
gradient-$\times$-input on bias-free ReLU
networks~\cite{ancona2018gradient}). Section~\ref{sec:failure}
recasts the most-cited critiques of attribution
methods~\cite{adebayo2018sanity,kindermans2019unreliability,
ghorbani2019fragile,slack2020fooling,kumar2020problems} as
\emph{mathematical} problems tied to explicit modelling choices.

\item \textbf{A proposed reporting checklist
(Section~\ref{sec:reporting}).} Feature-attribution methods encode
modelling choices. Section~\ref{sec:reporting} expresses this point as
a ten-item checklist for papers that report or rely on attribution
results.
\end{enumerate}

The remainder of the paper is organized as follows. Section~\ref{sec:formulation}
fixes notation. Section~\ref{sec:axioms} states the axioms. Sections~\ref{sec:shapley}--\ref{sec:perturbation}
survey the four method families. Section~\ref{sec:comparison}
presents the unifying comparison framework.
Section~\ref{sec:evaluation} reviews evaluation theory.
Section~\ref{sec:failure} analyzes failure modes.
Section~\ref{sec:applications} surveys applications across model
families. Section~\ref{sec:reporting} presents the proposed reporting
checklist. Section~\ref{sec:limitations} states the scope boundaries of
the survey. Section~\ref{sec:open} closes with open problems and a
research agenda.

\subsection{Central Claim}

The argument of the paper is summarized by the following claim:

\begin{quote}\itshape
There is no assumption-free feature-attribution method. Every local
additive attribution method defines feature importance through choices
about value functions, references, paths, perturbation distributions,
or conservation rules. Trustworthy use therefore requires reporting
a heatmap or ranking together with the assumptions under which the
attribution was computed and interpreted.
\end{quote}

\noindent The checklist supports reproducible reporting of the
assumptions required to compute and interpret an attribution. The
remainder of the paper develops this claim formally and shows how
common failure cases arise when these assumptions are left implicit.

\section{Problem Formulation and Notation}
\label{sec:formulation}

This section fixes a single notation used throughout the paper. Every
method in Sections~\ref{sec:shapley}--\ref{sec:perturbation} is
described in the symbols introduced here, making differences between
methods visible as differences in specific mathematical objects across
papers with otherwise distinct notation.

\subsection{Model and Input Space}

We consider a predictive model
\[
\model \colon \mathcal{X} \to \R,
\]
where $\mathcal{X} \subseteq \R^d$ is the input space and $d$ is the
number of input features. For multi-class problems, $\model$ is the
scalar logit or pre-softmax score associated with a target class $c$;
when the dependence on $c$ matters, we write $\model_c$.
We make no structural assumption on $\model$: it may be a
deep network, a tree ensemble, or any other function from
$\mathcal{X}$ to $\R$. Where differentiability is required, we will say
so explicitly.

The data distribution on $\mathcal{X}$ is denoted $p_X$, with
$X \sim p_X$ a random input. An individual input is written
$x = (x_1, \dots, x_d) \in \R^d$, with the $i$-th coordinate $x_i$.

\subsection{Baselines and References}

Every additive attribution method requires, implicitly or explicitly, a
\emph{reference point} against which the prediction $\model(x)$ is
compared. We denote a single reference point as
$\baseline \in \R^d$ and a reference distribution as
$\mathcal{D}_{\baseline}$. Common choices include
\begin{itemize}[leftmargin=*]
\item a fixed point such as the all-zeros vector
$\baseline = \bm{0}$, used in many gradient implementations;
\item the population mean $\baseline = \E[X]$;
\item a sample from the training distribution, $\baseline \sim p_X$,
giving an \emph{expected baseline}~\cite{erion2021expectedgradients};
\item a structured reference (e.g.,~a blurred image, a paraphrase of a
sentence, an isoelectric protein sequence) appropriate to the input
modality.
\end{itemize}
We use $\Delta f(x; \baseline) \coloneqq \model(x) - \model(\baseline)$
for the prediction-minus-baseline difference, which most additive
methods target as the quantity to decompose.

\subsection{Coalitions and Value Functions}

Let $\featureset = \{1, 2, \dots, d\}$ index the features. A
\emph{coalition} is a subset $\coalition \subseteq \featureset$ of
features deemed ``present''; its complement $\bar{\coalition} = \featureset \setminus \coalition$ is the
set of ``absent'' features. The number of features in $\coalition$ is
$|\coalition|$, and $|\featureset| = d$.

A \emph{value function} is a set function
\[
\valuefn \colon 2^{\featureset} \to \R, \qquad \coalition \mapsto \valuefn(\coalition),
\]
intended to represent ``the prediction when only features in
$\coalition$ are observed''. In Shapley-based attribution, the value
function is often the most consequential modelling choice, and the
ambiguity in defining it is the source of much downstream disagreement.
Three canonical choices appear repeatedly:

\begin{description}[style=nextline,leftmargin=1.5em]
\item[Marginal (interventional) value.] Replace absent features with
their reference values:
\begin{equation}
\valuefn^{\mathrm{int}}_{\baseline}(\coalition) =
\E_{\baseline \sim \mathcal{D}_{\baseline}}\!\left[ \model\!\left( x_{\coalition}, \baseline_{\bar{\coalition}} \right) \right],
\label{eq:value-interventional}
\end{equation}
where $(x_{\coalition}, \baseline_{\bar{\coalition}})$ denotes the
vector with feature $i$ set to $x_i$ if $i \in \coalition$ and to
$\baseline_i$ otherwise. This corresponds to a hard intervention.

\item[Conditional value.] Condition on observed features:
\begin{equation}
\valuefn^{\mathrm{cond}}(\coalition) =
\E\!\left[ \model(X) \,\big|\, X_{\coalition} = x_{\coalition} \right].
\label{eq:value-conditional}
\end{equation}
This restricts to the data manifold but requires estimating
$d$-dimensional conditional expectations~\cite{aas2021conditional,frye2020manifold}.

\item[Single-reference value.] The deterministic limit of the
interventional value with a fixed baseline:
\begin{equation}
\valuefn^{\mathrm{ref}}_{\baseline}(\coalition) =
\model\!\left( x_{\coalition}, \baseline_{\bar{\coalition}} \right).
\label{eq:value-ref}
\end{equation}
\end{description}

The interventional and conditional value functions agree only when
features are mutually independent under $p_X$. In all other cases they
yield different Shapley
values~\cite{janzing2020causal,sundararajan2020manyshap,merrick2020explanationgame}.

\subsection{Paths and Perturbations}

Path-based methods integrate gradients along a continuous curve in input
space. A \emph{path} from a baseline $\baseline$ to an input $x$ is a
differentiable map
\[
\gamma \colon [0,1] \to \R^d, \qquad \gamma(0) = \baseline, \quad \gamma(1) = x.
\]
The canonical choice is the straight line
$\gamma(\alpha) = \baseline + \alpha (x - \baseline)$, used by
Integrated Gradients. Alternatives include adaptive
paths~\cite{kapishnikov2021guidedig}, paths through image scale
space~\cite{xu2020blurig}, and paths defined by a generative
model~\cite{chang2019counterfactual}.

Perturbation-based methods do not integrate but instead evaluate
$\model$ at masked inputs. A \emph{mask} is a vector
$\mask \in [0,1]^d$ that interpolates between the input and a
reference; we use $\mask \odot x + (1 - \mask) \odot \baseline$ to
denote a masked input. When $\mask \in \{0,1\}^d$ the mask is binary,
and we identify $\mask$ with the coalition
$\coalition = \{i : \mask_i = 1\}$.

\subsection{Attribution Vectors and Explanation Surrogates}

An \emph{attribution vector} for the prediction $\model(x)$ relative to
a baseline $\baseline$ is a vector
\[
\attribution(\model, x, \baseline) \in \R^d
\]
whose $i$-th entry $\phi_i$ is the signed contribution of feature $i$.
Most methods we discuss are \emph{additive}, meaning they aim to satisfy
the \emph{local accuracy} (or completeness, or efficiency) property:
\begin{equation}
\sum_{i=1}^{d} \phi_i \;=\; \model(x) - \model(\baseline)
\;=\; \Delta f(x; \baseline).
\label{eq:completeness}
\end{equation}
Equation~\eqref{eq:completeness} is the closest the field has to a
universal target; we will return to it as Axiom~\ref{ax:completeness}
in Section~\ref{sec:axioms}.

Lundberg and Lee~\cite{lundberg2017unified} formalized the class of
\emph{additive feature attribution methods} as those that admit an
explanation surrogate
\begin{equation}
g(\mask) = \phi_0 + \sum_{i=1}^{d} \phi_i \, \mask_i, \qquad \mask \in \{0,1\}^d,
\label{eq:additive-surrogate}
\end{equation}
where $\phi_0 = \model(\baseline)$ is the offset. Setting $\mask = \bm{1}$ recovers
\eqref{eq:completeness}. This linear-in-mask surrogate is the common
language of LIME, SHAP, DeepLIFT, LRP, and Integrated Gradients; their
differences lie in \emph{how} the coefficients $\phi_i$ are computed,
not in the form of $g$.

\subsection{A Running Example: Disagreement Without Error}
\label{sec:running-example}

A two-feature interaction already shows why attribution methods can
disagree without either method being erroneous. Let
\[
\model(x_1,x_2) = x_1 x_2, \qquad x = (1,1).
\]
With the zero baseline $\baseline_0=(0,0)$, straight-line IG follows
$\gamma(\alpha)=(\alpha,\alpha)$ and gives
\[
\IG_1(x;\baseline_0)=\IG_2(x;\baseline_0)
=\int_0^1 \alpha\,d\alpha = \frac{1}{2}.
\]
The corresponding single-reference Shapley value with
$\valuefn^{\mathrm{ref}}_{\baseline_0}$ also assigns
$(1/2,1/2)$, because neither feature has value without the other.

Now change only the reference point to $\baseline_1=(1,0)$. The same
model and same prediction give
\[
\IG(x;\baseline_1)=(0,1),
\]
because the path changes only the second coordinate and decomposes
$\model(1,1)-\model(1,0)=1$. Both answers are valid relative to their
baselines; they answer different comparison questions. If, in addition,
the data distribution has $X_2=X_1$, then the coalition inputs
$(1,0)$ and $(0,1)$ used by interventional Shapley or occlusion are
off the data manifold. A conditional or manifold-aware value function
therefore changes the interpretation again, not because the model
changed but because the meaning of ``feature absence'' changed.

\begin{table}[t]
\centering
\caption{Running example for $\model(x_1,x_2)=x_1x_2$ at $x=(1,1)$.
Different specifications answer different questions.}
\label{tab:running-example}
\footnotesize
\begin{tabularx}{\columnwidth}{@{}p{0.36\columnwidth}p{0.22\columnwidth}X@{}}
\toprule
Specification & Attribution & Interpretation \\
\midrule
IG, $\baseline=(0,0)$ & $(1/2,1/2)$ & interaction split evenly \\
IG, $\baseline=(1,0)$ & $(0,1)$ & credit for changing $x_2$ only \\
Single-reference Shapley, $\baseline=(0,0)$ & $(1/2,1/2)$ & coalition interaction split evenly \\
Off-manifold perturbation under $X_2=X_1$ & specification-dependent & masked inputs leave $\mathrm{supp}(p_X)$ \\
\bottomrule
\end{tabularx}
\end{table}

\subsection{Notation Summary}

Table~\ref{tab:notation} summarizes the notation used throughout the
rest of the paper.

\begin{table}[t]
\centering
\caption{Notation used throughout the survey.}
\label{tab:notation}
\small
\begin{tabular}{@{}ll@{}}
\toprule
Symbol & Meaning \\
\midrule
$\model$ & Predictive model $\mathcal{X} \to \R$ \\
$x, x_i$ & Input vector and $i$-th feature \\
$d$ & Number of features \\
$\featureset$ & Feature index set $\{1,\dots,d\}$ \\
$\baseline$ & Baseline / reference input \\
$\mathcal{D}_{\baseline}$ & Reference distribution \\
$\Delta f(x; \baseline)$ & $\model(x) - \model(\baseline)$ \\
$\coalition \subseteq \featureset$ & Coalition of present features \\
$\valuefn(\coalition)$ & Value function on coalitions \\
$\gamma(\alpha)$ & Path from baseline to input \\
$\mask$ & Perturbation mask in $[0,1]^d$ \\
$\attribution = (\phi_i)$ & Attribution vector in $\R^d$ \\
$g$ & Additive explanation surrogate \\
$p_X$ & Data distribution \\
$\nabla \model$ & Gradient of model w.r.t.\ input \\
\bottomrule
\end{tabular}
\end{table}

\section{Axiomatic Foundations}
\label{sec:axioms}

This section catalogues the axioms used to characterize feature
attribution methods. The axioms originate in two distinct traditions:
the \emph{cooperative-game} tradition derived from Shapley's
1953~paper~\cite{shapley1953value}, and the \emph{path-method} tradition
introduced by Sundararajan, Taly, and Yan in their 2017~paper on
Integrated Gradients~\cite{sundararajan2017axiomatic}. A central
mathematical observation of this survey, made explicit in
Section~\ref{sec:comparison}, is that these two axiom systems are not
independent: many path-method axioms have direct game-theoretic
analogues, and some methods (notably DeepLIFT and SHAP variants) can be
characterized in either language.

\subsection{Shapley Axioms}

Let $\valuefn \colon 2^{\featureset} \to \R$ be a value function with
$\valuefn(\emptyset) = 0$. A \emph{value} $\phi$ assigns to each player
$i \in \featureset$ a real number $\phi_i(\valuefn)$.

\begin{axiom}[Efficiency / Completeness]
\label{ax:efficiency}
$\sum_{i=1}^d \phi_i(\valuefn) = \valuefn(\featureset)$.
\end{axiom}

\begin{axiom}[Symmetry]
\label{ax:symmetry}
If players $i, j$ are interchangeable in $\valuefn$, meaning
$\valuefn(\coalition \cup \{i\}) = \valuefn(\coalition \cup \{j\})$ for
every $\coalition \subseteq \featureset \setminus \{i,j\}$, then
$\phi_i(\valuefn) = \phi_j(\valuefn)$.
\end{axiom}

\begin{axiom}[Dummy / Null Player]
\label{ax:dummy}
If $\valuefn(\coalition \cup \{i\}) = \valuefn(\coalition)$ for every
$\coalition \subseteq \featureset \setminus \{i\}$, then
$\phi_i(\valuefn) = 0$.
\end{axiom}

\begin{axiom}[Additivity / Linearity]
\label{ax:linearity}
For any two value functions $\valuefn_1, \valuefn_2$ and any
$\alpha, \beta \in \R$,
$\phi_i(\alpha\, \valuefn_1 + \beta\, \valuefn_2) = \alpha\, \phi_i(\valuefn_1) + \beta\, \phi_i(\valuefn_2)$.
\end{axiom}

\begin{result}[Shapley, 1953~\cite{shapley1953value}]
\label{thm:shapley}
\emph{Assumptions:} a finite player set $\featureset = \{1,\dots,d\}$;
a real-valued set function $\valuefn \colon 2^{\featureset} \to \R$
with $\valuefn(\emptyset) = 0$. Under these assumptions, there is a
unique value $\phi$ on $\featureset$ satisfying
Axioms~\ref{ax:efficiency}--\ref{ax:linearity} for all such
$\valuefn$, given by
\begin{equation}
\phi_i(\valuefn) =
\sum_{\coalition \subseteq \featureset \setminus \{i\}}
\frac{|\coalition|!\,(d - |\coalition| - 1)!}{d!}
\bigl[\valuefn(\coalition \cup \{i\}) - \valuefn(\coalition)\bigr].
\label{eq:shapley-value}
\end{equation}
\end{result}

\noindent The quantity in brackets, $\valuefn(\coalition \cup \{i\}) -
\valuefn(\coalition)$, is the \emph{marginal contribution} of feature
$i$ to coalition $\coalition$. The weight
$|\coalition|!\,(d-|\coalition|-1)!/d!$ is the probability that, under
a uniformly random ordering of the $d$ players, exactly the members of
$\coalition$ precede $i$; the Shapley value is thus the expected
marginal contribution of $i$ over random arrival orders.

The uniqueness half of Result~\ref{thm:shapley} follows from a basis
argument that is worth recording because it recurs in later
characterizations. The \emph{unanimity games}
$u_T(\coalition) = \indicator[T \subseteq \coalition]$, for nonempty
$T \subseteq \featureset$, form a basis of the
$(2^d-1)$-dimensional space of value functions with
$\valuefn(\emptyset)=0$. On a unanimity game, the dummy axiom forces
$\phi_i(u_T) = 0$ for $i \notin T$, and symmetry with efficiency
forces $\phi_i(u_T) = 1/|T|$ for $i \in T$. Linearity then determines
$\phi$ on every $\valuefn = \sum_T c_T u_T$, and evaluating the
resulting expression coalition by coalition
yields~\eqref{eq:shapley-value}. The full argument is in the original
paper~\cite{shapley1953value}.

\subsection{Additive Feature Attribution}

Recall the additive surrogate of \eqref{eq:additive-surrogate}. Lundberg
and Lee~\cite{lundberg2017unified} showed that a single triple of
axioms forces any method expressible as a linear-in-mask surrogate to
coincide with Shapley values.

\begin{axiom}[Local Accuracy]
\label{ax:completeness}
$g(\bm{1}) = \model(x)$, i.e.\ $\sum_i \phi_i + \phi_0 = \model(x)$.
\end{axiom}

\begin{axiom}[Missingness]
\label{ax:missingness}
For any $i$ with $x_i = \baseline_i$, $\phi_i = 0$.
\end{axiom}

\begin{axiom}[Consistency~\cite{lundberg2017unified,lundberg2020treeshap}]
\label{ax:consistency}
If $\model'$ is a model such that the marginal contribution of feature
$i$ in $\model'$ is at least its marginal contribution in $\model$ for
every coalition, then $\phi_i(\model', x) \geq \phi_i(\model, x)$.
\end{axiom}

\begin{result}[Lundberg and Lee, 2017~\cite{lundberg2017unified}, Thm.~1]
\emph{Assumptions:} the attribution takes the additive surrogate
form~\eqref{eq:additive-surrogate} for a binary mask
$\mask \in \{0,1\}^d$; a value function $\valuefn_{f,x}$ over
coalitions of features is fixed. Under these assumptions, the
unique solution satisfying
Axioms~\ref{ax:completeness}--\ref{ax:consistency} coincides with the
Shapley value $\phi_i = \phi_i^{\mathrm{Shap}}(\valuefn_{f,x})$. The
proof is Theorem~1 of~\cite{lundberg2017unified}.
\end{result}

This is the formal sense in which SHAP \emph{unifies} the additive
attribution family. The unification, however, leaves open the choice of
$\valuefn_{f,x}$, and as we saw in Section~\ref{sec:formulation} that
choice is itself consequential.

\subsection{Alternative Cooperative-Game Values}

The finite-player Shapley value is the dominant attribution index in
XAI, but it is not the only cooperative-game value relevant to
explanation. The Banzhaf value~\cite{banzhaf1965weighted} averages
marginal contributions under a different coalition weighting scheme,
placing equal weight on coalitions, while the Shapley value averages player arrival
orders. Owen's multilinear extension~\cite{owen1972multilinear}
connects finite games to polynomial extensions on the unit cube and is
one route by which sampling and interaction calculations can be
studied analytically. For continuous populations and cost-sharing
problems, the Aumann-Shapley value~\cite{aumann1974values} replaces
finite coalitions with pathwise marginal rates. Specialized to a
differentiable cost function $\model$ on $\R^d$ with reference point
$\baseline$, the Aumann-Shapley charge to coordinate $i$ is the
diagonal-path integral
\begin{equation}
\phi^{\mathrm{AS}}_i(x; \baseline)
= (x_i - \baseline_i) \int_0^1
\frac{\partial \model}{\partial x_i}
\bigl(\baseline + \alpha (x - \baseline)\bigr)\, d\alpha,
\label{eq:aumann-shapley}
\end{equation}
which is exactly the Integrated Gradients operator of
Section~\ref{sec:path}. This connection is substantive: it identifies
IG as a fixed-path, feature-level cost-sharing rule, with the baseline
and the straight-line path replacing the cooperative game's population
model.

\subsection{Path-Method Axioms}

The axiom set of Sundararajan et al.~\cite{sundararajan2017axiomatic}
targets attribution methods defined for differentiable models.

\begin{axiom}[Sensitivity-(a)]
\label{ax:sensa}
If the input $x$ and baseline $\baseline$ differ in exactly one feature
and the predictions differ ($\model(x) \neq \model(\baseline)$), then
that feature must receive a nonzero attribution.
\end{axiom}

\begin{axiom}[Sensitivity-(b)]
\label{ax:sensb}
If $\model$ does not mathematically depend on feature $i$, then
$\phi_i = 0$. (This is the differentiable analogue of dummy.)
\end{axiom}

\begin{axiom}[Implementation Invariance]
\label{ax:invariance}
If two networks $\model_1$ and $\model_2$ compute the same function
($\model_1(x) = \model_2(x)$ for all $x$), their attributions are
identical.
\end{axiom}

\begin{axiom}[Completeness]
\label{ax:path-completeness}
$\sum_i \phi_i = \model(x) - \model(\baseline)$.
\end{axiom}

\begin{axiom}[Linearity]
$\phi$ is linear in $\model$.
\end{axiom}

\begin{axiom}[Symmetry-Preserving]
If $x$ and $\baseline$ are symmetric in two features $i, j$
($x_i = x_j$ and $\baseline_i = \baseline_j$) and $\model$ is
symmetric in those features, then $\phi_i = \phi_j$.
\end{axiom}

\begin{result}[Sundararajan et al., 2017~\cite{sundararajan2017axiomatic}; further axiomatic characterizations by Lundstrom and Razaviyayn, 2025~\cite{lundstrom2025ig}]
\label{thm:ig-unique}
\emph{Assumptions:} $\model$ is differentiable along the
straight-line path $\gamma(\alpha) = \baseline + \alpha(x - \baseline)$;
the gradient is integrable on $\alpha \in [0,1]$. Under these
assumptions and within the attribution classes considered in the cited
papers, Integrated Gradients along the straight-line path is singled
out by Axioms~\ref{ax:sensa}--\ref{ax:invariance} together with
completeness, linearity, and symmetry-preservation. Lundstrom and
Razaviyayn~\cite{lundstrom2025ig} give three additional independent
axiomatic characterizations under varied axiom subsets.
\end{result}

That several different axiom combinations identify the same
straight-line operator strengthens the case for IG as a canonical
member of the path family, but it does not make the operator
assumption-free: each characterization fixes the attribution class and
the path in advance.

\subsection{Conservation and Backpropagation Axioms}

Layer-wise relevance propagation (LRP)~\cite{bach2015lrp} and
DeepLIFT~\cite{shrikumar2017deeplift} are characterized not by a
coalition or a path but by a \emph{conservation rule} that distributes
a quantity through the network's computation graph.

\begin{axiom}[Layer-wise Conservation~\cite{bach2015lrp}]
\label{ax:lrp-conservation}
For every layer $\ell$, the sum of relevances assigned to its inputs
equals the sum of relevances received from its outputs:
\begin{equation}
\sum_{i \in \mathrm{in}(\ell)} R_i^{(\ell)} = \sum_{j \in \mathrm{out}(\ell)} R_j^{(\ell+1)}.
\end{equation}
At the output layer, $R^{(L)} = \model(x)$. At the input layer,
$\sum_i R_i^{(0)} = \model(x)$, so layer-wise conservation implies
completeness.
\end{axiom}

\begin{axiom}[Summation-to-Delta~\cite{shrikumar2017deeplift}]
\label{ax:deeplift-completeness}
DeepLIFT attributions $C_{\Delta x_i \Delta y}$ satisfy
$\sum_i C_{\Delta x_i \Delta y} = \Delta y$, where $\Delta y = \model(x) - \model(\baseline)$.
\end{axiom}

Although LRP and DeepLIFT are sometimes presented as alternatives to
Integrated Gradients, all three satisfy a completeness axiom of the
form $\sum_i \phi_i = \Delta y$. They differ in \emph{how} the
distribution is performed, not in \emph{what} is preserved.

\subsection{Robustness and Stability Axioms}

Beyond the classical axioms, several authors have proposed properties
that capture an attribution method's robustness to small perturbations.

\begin{axiom}[Lipschitz Stability~\cite{alvarezmelis2018robustness}]
\label{ax:lipschitz}
There exists $L < \infty$ such that
$\|\attribution(x_1) - \attribution(x_2)\| \leq L \, \|x_1 - x_2\|$
for all $x_1, x_2$.
\end{axiom}

\begin{axiom}[Continuity]
\label{ax:continuity}
$\attribution$ is continuous in $x$ (a strictly weaker condition than
Axiom~\ref{ax:lipschitz}).
\end{axiom}

These properties are not satisfied by raw gradient saliency in
practice~\cite{ghorbani2019fragile,dombrowski2019manipulated}, and
their failure is one of the principal motivations for SmoothGrad and
similar averaging methods (Section~\ref{sec:gradient}).

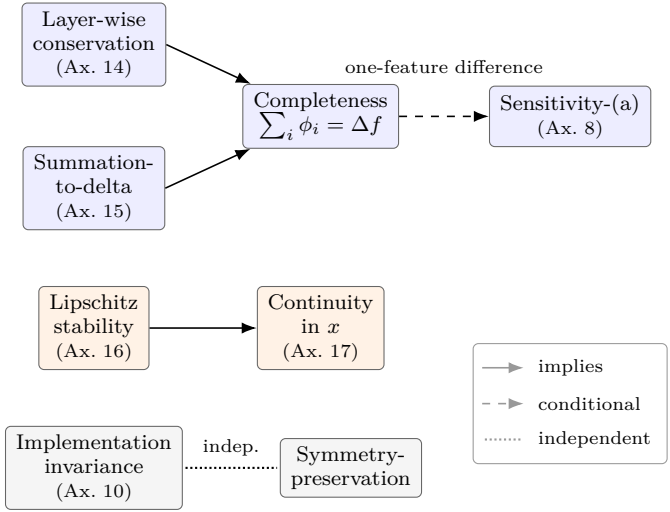
\begin{figure}[t]
\centering
\begin{tikzpicture}[
  font=\footnotesize,
  ax/.style={draw=black!60, rounded corners=2pt, inner xsep=4pt,
    inner ysep=3pt, minimum height=7mm, align=center, fill=blue!7},
  axr/.style={ax, fill=orange!10},
  axn/.style={ax, fill=gray!8},
  impl/.style={-{Latex[length=2mm]}, semithick},
  conds/.style={-{Latex[length=2mm]}, semithick, dashed},
  indep/.style={densely dotted, thick, shorten >=1pt, shorten <=1pt}
]
\node[ax] (lwc)  at (0,1.45)    {Layer-wise\\ conservation\\ {\scriptsize (Ax.~\ref{ax:lrp-conservation})}};
\node[ax] (s2d)  at (0,-0.45)   {Summation-\\ to-delta\\ {\scriptsize (Ax.~\ref{ax:deeplift-completeness})}};
\node[ax] (comp) at (3.0,0.5)   {Completeness\\ $\sum_i \phi_i = \Delta f$};
\node[ax] (sa)   at (6.3,0.5)   {Sensitivity-(a)\\ {\scriptsize (Ax.~\ref{ax:sensa})}};
\draw[impl]  (lwc.east)  -- (comp);
\draw[impl]  (s2d.east)  -- (comp);
\draw[conds] (comp) -- node[above=4.5mm, font=\scriptsize]
  {one-feature difference} (sa);
\node[axr] (lip)  at (0,-2.3)  {Lipschitz\\ stability\\ {\scriptsize (Ax.~\ref{ax:lipschitz})}};
\node[axr] (cont) at (3.0,-2.3) {Continuity\\ in $x$\\ {\scriptsize (Ax.~\ref{ax:continuity})}};
\draw[impl] (lip) -- (cont);
\node[axn] (ii)  at (0,-4.15)  {Implementation\\ invariance\\ {\scriptsize (Ax.~\ref{ax:invariance})}};
\node[axn] (sym) at (3.4,-4.15) {Symmetry-\\ preservation};
\draw[indep] (ii) -- node[above=1pt, font=\scriptsize] {indep.} (sym);
\matrix[draw=black!40, rounded corners=2pt, inner sep=3pt, font=\scriptsize,
  column 2/.style={anchor=west}, column sep=2pt, row sep=2pt]
  at (6.3,-3.3) {
  \draw[impl] (0,0) -- (0.55,0); & \node{implies}; \\
  \draw[conds] (0,0) -- (0.55,0); & \node{conditional}; \\
  \draw[indep] (0,0) -- (0.55,0); & \node{independent}; \\
};
\end{tikzpicture}
\caption{Inter-axiom relationships. Layer-wise conservation
(Axiom~\ref{ax:lrp-conservation}) and summation-to-delta
(Axiom~\ref{ax:deeplift-completeness}) each imply completeness.
Completeness implies Sensitivity-(a) only in the Sensitivity-(a) setup,
where $x$ and $\baseline$ differ in exactly one feature. Lipschitz
stability is strictly stronger than continuity. Implementation
invariance and symmetry-preservation are logically independent
(Remark~\ref{prop:implications}); the converses of all displayed
implications fail in general.}
\label{fig:axioms}
\end{figure}

\subsection{Inter-axiom Relationships}

Many axioms imply or are implied by others. We collect the most useful
relationships here for reference; they will recur in
Section~\ref{sec:comparison}.

\begin{remark}[Inter-axiom implications]
\label{prop:implications}
The following implications follow from the definitions of the axioms;
we list them as observations.
\begin{enumerate}[leftmargin=*,label=(\roman*)]
\item Completeness $\Rightarrow$ Sensitivity-(a), under the
Sensitivity-(a) setup: if $x$ and $\baseline$ differ in exactly one
feature and $\model(x) \neq \model(\baseline)$, the sum constraint
forces that changed feature to receive nonzero attribution. Without
the one-feature-difference condition, completeness implies only that
at least one feature has nonzero attribution.
\item Layer-wise Conservation $\Rightarrow$ Completeness: summing the
per-layer conservation from output to input yields
$\sum_i R_i^{(0)} = \model(x)$.
\item Lipschitz Stability $\Rightarrow$ Continuity in $x$ by
definition.
\item Implementation Invariance and symmetry-preservation are
logically independent; neither implies the other in general.
\end{enumerate}
\end{remark}

\noindent The converses generally fail: a method may be continuous
without being Lipschitz, may satisfy completeness without being
implementation invariant (e.g., a discrete-gradient method on a ReLU
network), and may satisfy Sensitivity-(a) without satisfying
completeness (e.g.,~raw saliency).

The path forward in the next four sections is to take each major method
family and identify, with reference to this axiom catalogue, exactly
which axioms it does and does not satisfy. The resulting matrix,
collected in Section~\ref{sec:comparison}, is a central comparison
table of this survey.

\section{Shapley and Cooperative-Game Attribution}
\label{sec:shapley}

The Shapley value~\cite{shapley1953value}, originally developed to
distribute the joint payoff of a cooperative game among its players, is
the most extensively studied attribution method. Its appeal is
axiomatic: it is the unique solution to a small set of intuitive
constraints (Result~\ref{thm:shapley}). Its difficulty is
computational: exact evaluation has cost exponential in the number of
features. The methods surveyed in this section can be understood as
different points on the trade-off curve between fidelity to the
axiomatic Shapley value and the cost of approximating it.

\subsection{Exact Shapley Values}

For a value function $\valuefn$ on $2^{\featureset}$, the Shapley
value $\phi_i$ is given by~\eqref{eq:shapley-value}. Equivalently,
\begin{equation}
\phi_i(\valuefn) = \frac{1}{d!} \sum_{\pi \in \Pi(\featureset)} \bigl[\valuefn(\mathrm{Pre}_i^{\pi} \cup \{i\}) - \valuefn(\mathrm{Pre}_i^{\pi})\bigr],
\label{eq:shapley-permutation}
\end{equation}
where $\Pi(\featureset)$ is the set of permutations of $\featureset$ and
$\mathrm{Pre}_i^{\pi}$ is the set of features that precede $i$ in
$\pi$. The two forms agree by a counting argument: a fixed coalition
$\coalition \subseteq \featureset \setminus \{i\}$ arises as
$\mathrm{Pre}_i^{\pi}$ for exactly $|\coalition|!\,(d-|\coalition|-1)!$
of the $d!$ permutations (order the members of $\coalition$, place
$i$, order the rest), so grouping the sum
in~\eqref{eq:shapley-permutation} by the value of
$\mathrm{Pre}_i^{\pi}$ recovers the weights
of~\eqref{eq:shapley-value}. The
form~\eqref{eq:shapley-permutation} is more convenient for
sampling-based approximation: drawing permutations uniformly at random
gives an unbiased Monte Carlo estimator whose variance decays at the
standard $O(1/M)$ rate in the number of sampled permutations $M$.

The principal cost of exact Shapley is the
$2^d$ evaluations of $\valuefn$, plus the
work of evaluating $\valuefn$ itself, which under either the
interventional or conditional definitions
(\eqref{eq:value-interventional}--\eqref{eq:value-conditional}) requires
in turn a model call or an expectation. For $d > 20$ exact evaluation
is generally infeasible, and the methods below trade an approximation
error for computational tractability.

\subsection{Pre-SHAP Game-Theoretic Feature Contributions}

Modern SHAP terminology can obscure an older line of work that already
treated local explanation as cooperative-game credit allocation.
\v{S}trumbelj and Kononenko~\cite{strumbelj2010efficient} formulated
individual classification explanations as feature-value contributions
whose sum equals the change from an expected output to the model's
prediction, and proposed a sampling approximation to avoid enumerating
all feature subsets. This places black-box individual explanation in
the Shapley lineage before the later SHAP unification.

Quantitative Input Influence (QII)~\cite{datta2016qii} developed a
related but distinct transparency framework: it asks how much an input
or group of inputs influences an output under specified interventions
or perturbation distributions. QII is important in this survey because
it separates the \emph{influence query} from the \emph{estimator}. That
separation anticipates the value-function distinction that now
dominates Shapley explanations: a numerical attribution is meaningful
only after the intervention, conditioning, or perturbation semantics
have been fixed.

\subsection{KernelSHAP}

KernelSHAP~\cite{lundberg2017unified} reformulates Shapley value
estimation as weighted linear regression. Sampling masks
$\mask \sim p_{\mathrm{Shap}}(\mask)$ with the \emph{Shapley kernel}
\begin{equation}
\pi_{\mathrm{Shap}}(\mask) = \frac{d - 1}{\binom{d}{|\mask|}\, |\mask| (d - |\mask|)},
\label{eq:shapley-kernel}
\end{equation}
KernelSHAP fits the additive surrogate $g(\mask)$ of
\eqref{eq:additive-surrogate} by weighted least squares with weights
$\pi_{\mathrm{Shap}}$:
\begin{equation}
\hat{\attribution} = \operatorname*{arg\,min}_{\phi_0, \dots, \phi_d}
\sum_{\mask \in \{0,1\}^d}
\pi_{\mathrm{Shap}}(\mask)
\Bigl( \valuefn(\mask) - \phi_0 - \sum_{i=1}^d \phi_i \mask_i \Bigr)^2,
\label{eq:kernelshap-wls}
\end{equation}
where $\valuefn(\mask)$ is the model evaluated on the masked input
under the chosen value function. The kernel
weight~\eqref{eq:shapley-kernel} is infinite at $|\mask| \in \{0, d\}$;
those two masks are handled as the hard constraints
$\phi_0 = \valuefn(\bm{0})$ and
$\sum_i \phi_i + \phi_0 = \valuefn(\bm{1})$, the latter being
completeness. Lundberg and Lee proved that the
solution of~\eqref{eq:kernelshap-wls} over the full mask distribution
coincides exactly with the Shapley values
of~\eqref{eq:shapley-value}. In practice, $M \ll 2^d$ masks are
sampled, yielding an estimator with bias and variance that have since
been characterized explicitly by Covert and
Lee~\cite{covert2021kernelshap} and refined by the unbiased estimator
of~\cite{chen2023algorithms}.

KernelSHAP is \emph{model-agnostic}: it requires only black-box access
to $\model$, making it applicable to gradient-free models. Its
correctness, however, depends on the value function used in the
regression's prediction targets. By default KernelSHAP marginalizes
over a fixed reference (the interventional value function), and as a
result it inherits all of the assumptions about feature independence
discussed in Section~\ref{sec:formulation}.

\subsection{TreeSHAP}

For tree ensembles, the structural recursion of the model can be
exploited to compute Shapley values exactly in polynomial time. Lundberg
et al.~\cite{lundberg2020treeshap} introduced TreeSHAP, which runs in
$O(T L D^2)$ time on a tree ensemble with $T$ trees, $L$ leaves, and
depth $D$. The algorithm maintains, at each node, a polynomial in
``coalition mass'' that encodes the contribution of each subtree to all
possible coalitions; this polynomial is propagated through the tree and
combined at the leaves.

TreeSHAP supports two value functions:
\emph{path-dependent} (the empirical conditional expectation along
training-time tree splits) and \emph{interventional} (the
marginal value over a reference dataset). The two yield different
attributions whenever features are correlated, a discrepancy now widely
recognized in the practical SHAP literature~\cite{janzing2020causal,
sundararajan2020manyshap}. TreeSHAP also satisfies the consistency
axiom (Axiom~\ref{ax:consistency}), a property that fails for naive
gain-based feature importance.

\subsection{DeepSHAP and GradientSHAP}

DeepSHAP~\cite{lundberg2017unified} extends DeepLIFT-Rescale rules to
approximate Shapley values for deep networks. The key insight is that
DeepLIFT's per-neuron multipliers can be interpreted as expectations
over a baseline distribution, and aggregated by linear combination to
yield a Shapley-style attribution. The approximation is exact for
linear models and for the composition of linear and ReLU layers with a
single fixed baseline; for more complex architectures it is an empirical
heuristic.

GradientSHAP averages Integrated Gradients computations over a
distribution of baselines drawn from the data~\cite{lundberg2017unified,
erion2021expectedgradients}. Specifically,
\begin{equation}
\begin{aligned}
\phi_i^{\mathrm{GradSHAP}}
&= \E_{\baseline \sim p_X,\, \alpha \sim U[0,1]}
\!\Biggl[
(x_i - \baseline_i) \\
&\qquad\qquad\cdot
\frac{\partial \model(\baseline + \alpha(x - \baseline))}
{\partial x_i}
\Biggr].
\end{aligned}
\label{eq:gradshap}
\end{equation}
The expectation over $\baseline$ is the link to Shapley reasoning, as
discussed in detail by Erion et al.~\cite{erion2021expectedgradients};
the expectation over $\alpha$ is the standard Integrated Gradients
path integral. The combination yields a sampling-based estimator of the
expected Shapley value with respect to the data distribution.

\subsection{SAGE (boundary case): Global Shapley Effects}

\textbf{Scope note.} SAGE produces a \emph{global} feature-importance
measure, one level above the local-attribution scope of this survey.
We include it because it shares the Shapley axiomatic foundation
with KernelSHAP and TreeSHAP and because its loss-based value
function illustrates how the same axiomatic machinery extends
beyond per-prediction attribution. We mark SAGE as a boundary case
in the axiom matrix (Table~\ref{tab:axiom-matrix}) and treat its
remaining global-explanation siblings as out of scope
(Section~\ref{sec:limitations}).

The methods above attribute a single prediction. SAGE
(Shapley Additive Global ExplanationS)~\cite{covert2020sage} extends
the framework to a global feature-importance measure by replacing the
per-input value function with a loss-based one:
\begin{equation}
\valuefn^{\mathrm{SAGE}}(\coalition) = -\E_{(X,Y)}\!\left[\ell(\model(X_{\coalition}, X_{\bar{\coalition}}'), Y)\right] + \mathrm{const}.
\end{equation}
Here $X_{\bar{\coalition}}'$ marginalizes over the missing features.
Shapley values of $\valuefn^{\mathrm{SAGE}}$ measure how much of the
model's predictive performance is attributable to each feature globally.
SAGE inherits the same independence concerns as KernelSHAP but resolves
them at the dataset level; KernelSHAP resolves them per prediction.
It is tempting to treat any global Shapley score as a feature-selection
criterion, but this is a separate modelling decision. Fryer,
Str{\"u}mke, and Nguyen~\cite{fryer2021featureselection} show through
counterexamples that the classical Shapley axioms do not by themselves
guarantee suitability for subset selection; the game formulation must
match the statistical objective of the selected feature set.

\subsection{Shapley-Taylor Interaction Indices}

Standard Shapley values attribute the prediction to single features. In
many applications, including protein--protein interactions, drug combinations,
and NLP feature interactions, joint effects are central.
Dhamdhere et al.~\cite{dhamdhere2020shapleytaylor} introduced the
Shapley-Taylor interaction index, generalizing Shapley to subsets:
\begin{equation}
\phi^{\mathrm{ST},k}_T(\valuefn) = \sum_{\coalition \subseteq \featureset \setminus T} w^k_{|\coalition|,d}\, \Delta_T \valuefn(\coalition),
\end{equation}
where $T \subseteq \featureset$ with $|T| \leq k$, $\Delta_T$ is the
$|T|$-th discrete derivative, and $w^k$ is a generalization of the
Shapley weight. The index is uniquely characterized by an axiom set
that extends the classical Shapley axioms (Theorems~3 and~4
of~\cite{dhamdhere2020shapleytaylor}), and reduces to Shapley values
when $k = 1$ and $|T| = 1$.

Janizek, Sturmfels, and Lee~\cite{janizek2021integratedhessians} give a
continuous analogue via second-order path
integrals (\emph{Integrated Hessians}), which we cover in
Section~\ref{sec:path}; the discrete and continuous interaction frames
are related by Theorem~5 of~\cite{janizek2021integratedhessians}.
Interaction discovery can also be approached directly from learned
model structure. Tsang et al.~\cite{tsang2018interactiondet} detect
statistical interactions from neural-network weights, making the
interaction object explicit without deriving it solely from local
feature perturbations. At the model-element level, Neuron
Shapley~\cite{ghorbani2020neuronshapley} applies Shapley valuation to
neurons or filters as the players. Both works are boundary
cases for this survey's local input-attribution scope, but they sharpen
an important point: the ``players'' in a cooperative explanation game
need not be raw input coordinates.

\subsection{Interventional vs.\ Conditional Value Functions}

In Shapley-based attribution, the value function is often the most
consequential modelling choice. The choice between
$\valuefn^{\mathrm{int}}$ (\eqref{eq:value-interventional}) and
$\valuefn^{\mathrm{cond}}$ (\eqref{eq:value-conditional}) determines
the answers to questions that look identical at the level of an
explanation but differ at the level of inference.

\textbf{Interventional Shapley} answers the question: ``if I forced
feature $i$ to take its baseline value, how would the prediction
change?'' This is the natural quantity for debugging, mechanism
discovery, and causal reasoning under the implicit assumption that the
features can be intervened on independently.

\textbf{Conditional Shapley} answers: ``conditional on observing
$x_i$, how does the expected prediction change?'' This is the natural
quantity for predictive importance under the observed data
distribution.

The two coincide only when features are mutually independent. In the
presence of correlated features, which is the usual case, they
differ, and the gap can be large. Janzing, Minorics, and
Bl{\"o}baum~\cite{janzing2020causal} argued that the conditional
value function conflates association with causation and recommends the
interventional value for almost all use cases. Aas, Jullum, and
L{\o}land~\cite{aas2021conditional} took the opposite position for
risk-management applications, arguing that an interventional value
function evaluates the model at points the data manifold never visits,
producing implausibly large attributions for highly correlated features.
Frye et al.~\cite{frye2020manifold} proposed manifold-aware Shapley
values that compute the interventional quantity but only over
on-manifold coalition completions.

Sundararajan and Najmi~\cite{sundararajan2020manyshap} catalogued no
fewer than four distinct ``Shapley values for model explanation'': the
conditional expectation, conditional expectation w.r.t.\ the model, the
baseline expectation, and the random baseline expectation. They
showed that they disagree on simple, low-dimensional examples. Their
recommendation, which we endorse, is that any paper presenting Shapley
attributions should explicitly name which of these is being computed.

\subsection{Estimation: Variance, Bias, and Recent Algorithms}

Sampling-based Shapley estimators introduce both bias and variance.
Chen et al.~\cite{chen2023algorithms} surveyed the estimator landscape,
distinguishing
\begin{enumerate}[leftmargin=*]
\item \emph{Permutation sampling}, which gives an unbiased Monte Carlo
estimator of~\eqref{eq:shapley-permutation};
\item \emph{KernelSHAP-style} weighted regression, which is more
sample-efficient on well-conditioned value functions;
\item \emph{Unbiased KernelSHAP}~\cite{covert2021kernelshap}, which
corrects the bias introduced by paired sampling.
\end{enumerate}
For tree models, TreeSHAP is exact; for differentiable models,
GradientSHAP is approximate but inexpensive; for arbitrary models,
KernelSHAP is the default. The choice is again a value-function
question: each estimator is optimal for a slightly different definition
of the underlying Shapley value, as Figure~3 of Chen et al.\ makes
explicit.

\subsection{Summary}

Table~\ref{tab:shapley-variants} summarizes the Shapley-family methods
discussed above. The table makes the central point of this section:
the methods agree on what they aim to compute, namely a Shapley
decomposition of the prediction, but differ in their value function,
their approximation algorithm, and the assumptions under which they are
exact.

\begin{table*}[t]
\centering
\caption{Shapley-value variants for feature attribution. The
``Value function'' column refers to the definitions in
\eqref{eq:value-interventional}--\eqref{eq:value-ref}. ``Exact'' means
the method recovers the Shapley value without sampling, under the
stated value function.}
\label{tab:shapley-variants}
\footnotesize
\begin{tabularx}{\textwidth}{@{}p{0.17\textwidth}p{0.20\textwidth}p{0.19\textwidth}p{0.14\textwidth}X@{}}
\toprule
Method & Value function & Estimator & Complexity & Feature dependence \\
\midrule
Exact Shapley & any & exhaustive & $O(2^d)$ & handled by choice of $\valuefn$ \\
KernelSHAP & interventional (default) & weighted regression & $O(M d)$ per sample & ignored \\
\makecell[l]{Unbiased\\KernelSHAP~\cite{covert2021kernelshap}} & interventional & paired sampling & $O(M d)$ & ignored \\
TreeSHAP~\cite{lundberg2020treeshap} & path-dep.\ or interventional & exact recursion & $O(T L D^2)$ & path-dependent option \\
DeepSHAP~\cite{lundberg2017unified} & multiplier-based & DeepLIFT propagation & $O(\mathrm{net})$ & approximation \\
GradientSHAP~\cite{erion2021expectedgradients} & expected baseline & path + baseline avg.\ & $O(K M)$ & via baseline distribution \\
SAGE~\cite{covert2020sage} & loss-based, global & permutation sampling & $O(M d^2)$ & via marginalization \\
Shapley-Taylor~\cite{dhamdhere2020shapleytaylor} & generalizes interventional & enumeration up to order $k$ & $O\!\left(\binom{d}{k}\right)$ & ignored at order $k$ \\
\bottomrule
\end{tabularx}
\end{table*}

\section{Path-Based Attribution}
\label{sec:path}

Path-based methods compute attributions by integrating the gradient of
$\model$ along a continuous path in input space from a baseline
$\baseline$ to the target input $x$. They are the continuous analogue
of cooperative-game methods: where Shapley values average marginal
contributions over discrete coalitions, path methods average gradients
over a continuous trajectory. The canonical instance, Integrated
Gradients, was introduced by Sundararajan, Taly, and
Yan~\cite{sundararajan2017axiomatic} and remains the most widely used
member of the family.

\subsection{Integrated Gradients}

Let $\model$ be differentiable. Integrated Gradients (IG) attributes to
feature $i$ the quantity
\begin{equation}
\IG_i(x; \baseline) = (x_i - \baseline_i) \int_0^1 \frac{\partial \model\bigl(\baseline + \alpha (x - \baseline)\bigr)}{\partial x_i} \, d\alpha.
\label{eq:ig}
\end{equation}
The integral is taken along the straight-line path
$\gamma(\alpha) = \baseline + \alpha (x - \baseline)$, evaluated by
Riemann sums in practice (typically $K \in [20, 300]$ steps).
Equation~\eqref{eq:ig} satisfies completeness exactly:
\begin{equation}
\sum_{i=1}^d \IG_i(x; \baseline) = \model(x) - \model(\baseline).
\label{eq:ig-completeness}
\end{equation}
Writing $g(\alpha) = \model(\gamma(\alpha))$, the chain rule gives
$g'(\alpha) = \sum_i \partial_i \model(\gamma(\alpha)) \,
(x_i - \baseline_i)$ because
$\gamma_i'(\alpha) = x_i - \baseline_i$ on the straight line, and so
\begin{equation}
\model(x) - \model(\baseline)
= g(1) - g(0)
= \int_0^1 g'(\alpha)\, d\alpha
= \sum_{i=1}^d \IG_i(x;\baseline),
\label{eq:ig-ftc}
\end{equation}
the middle equality being the fundamental theorem of calculus and the
last exchanging the finite sum with the integral.

\begin{result}[Sundararajan et al.~\cite{sundararajan2017axiomatic}]
\label{thm:ig-axioms}
\emph{Assumptions:} $\model$ is differentiable on the straight-line
path between $\baseline$ and $x$; the gradient
$\partial \model / \partial x_i$ is integrable on $[0,1]$. Under
these assumptions, Integrated Gradients along the straight-line path
satisfies sensitivity-(a), sensitivity-(b), implementation
invariance, completeness, linearity, and symmetry-preservation
(Axioms~\ref{ax:sensa}--\ref{ax:invariance} plus completeness and
symmetry). Failures of differentiability (e.g.,~ReLU kink points) are
handled by the standard interpretation of partial derivatives as
subgradients almost everywhere along the path.
\end{result}

Implementation invariance follows from the fact that
$\partial \model/\partial x_i$ depends only on the input-output behaviour
of $\model$, not on the network's parametric form. Completeness
follows from the chain rule. Sensitivity-(a) follows because the
integrand is nonzero whenever $\model(x) \neq \model(\baseline)$ along
$\gamma$. These properties are robust under any monotone reparameterization
of the path, but not under arbitrary changes to $\gamma$:
implementation invariance, in particular, fails for paths that depend
on the model's parametrization, an observation made precise by Lundstrom
and Razaviyayn~\cite{lundstrom2025ig}.

\subsection{Baselines and Path Sensitivity}

The IG attribution depends on the baseline $\baseline$. This dependence
is mathematically necessary to define $\Delta f$ and is substantively
important. The choice of
$\baseline$ encodes what counts as the ``absence'' of a feature.

Sturmfels, Lundberg, and Lee~\cite{sturmfels2020baselines} catalogued
the most common baselines and their failure modes:

\begin{itemize}[leftmargin=*]
\item The \emph{zero baseline} $\baseline = \bm{0}$ is the default in
many implementations but is often pathological. In image models, black
pixels are themselves features, and Sensitivity-(a) is then satisfied
in a misleading way: every dark pixel in $x$ receives an attribution
of magnitude $|x_i \cdot \partial f / \partial x_i|$ that conflates
``important to the model'' with ``not equal to zero''.

\item The \emph{mean baseline} $\baseline = \E[X]$ has the merit of
being on or near the data manifold but can still be highly atypical.
For ImageNet, the mean is a uniform grey image.

\item The \emph{expected baseline}, due to Erion
et al.~\cite{erion2021expectedgradients}, replaces a fixed
$\baseline$ with an expectation over the training distribution:
\begin{equation}
\begin{aligned}
\mathrm{EG}_i(x)
&= \E_{\baseline \sim p_X,\, \alpha \sim U[0,1]}
\!\Biggl[
(x_i - \baseline_i) \\
&\qquad\qquad\cdot
\frac{\partial \model(\baseline + \alpha(x - \baseline))}
{\partial x_i}
\Biggr].
\end{aligned}
\label{eq:eg}
\end{equation}
This was already encountered as~\eqref{eq:gradshap}; in the
path-integral framing it is the natural extension of IG to a
distribution of baselines.

\item \emph{Adversarial} or \emph{informative} baselines are chosen so
that $\model(\baseline)$ is itself meaningful (e.g., a baseline of the
target class for a counterfactual attribution).
\end{itemize}

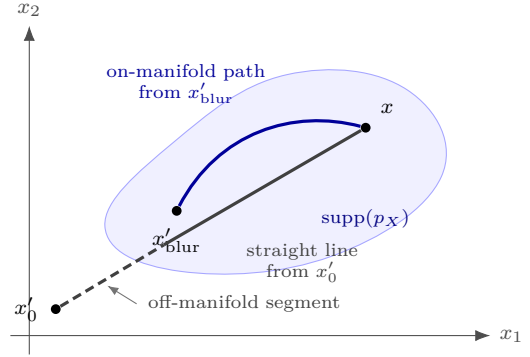
\begin{figure}[t]
\centering
\begin{tikzpicture}[font=\footnotesize,
  axline/.style={-{Latex[length=2mm]}, black!70},
  pt/.style={circle, fill=black, inner sep=1.3pt},
  offpath/.style={very thick, black!75, dash pattern=on 3.5pt off 2pt},
  onpath/.style={very thick, black!75},
  curvepath/.style={very thick, blue!60!black},
  manifold/.style={draw=blue!35, fill=blue!6}
]
\draw[axline] (-0.25,0) -- (6.1,0) node[right] {$x_1$};
\draw[axline] (0,-0.25) -- (0,4.1) node[above] {$x_2$};
\draw[manifold] plot [smooth cycle, tension=0.7] coordinates
  {(1.0,1.45) (1.9,2.55) (3.4,3.45) (5.0,3.25) (5.45,2.1) (4.0,1.05) (2.0,0.85)};
\node[font=\scriptsize, blue!50!black] at (4.45,1.55) {$\mathrm{supp}(p_X)$};
\node[pt, label={[label distance=1pt]left:{$\baseline_{0}$}}] (b0) at (0.35,0.35) {};
\node[pt, label={[label distance=1pt]below:{$\baseline_{\mathrm{blur}}$}}] (bb) at (1.95,1.65) {};
\node[pt, label={[label distance=1pt]above right:{$x$}}] (xx) at (4.45,2.75) {};
\coordinate (entry) at (1.78,1.20);
\draw[offpath] (b0) -- (entry);
\draw[onpath]  (entry) -- (xx);
\node[font=\scriptsize, black!75, align=center, anchor=north west]
  at (2.75,1.35) {straight line\\ from $\baseline_0$};
\node[font=\scriptsize, black!75, align=left, anchor=west]
  at (1.45,0.42) {off-manifold segment};
\draw[-{Latex[length=1.5mm]}, black!55, shorten >=1pt]
  (1.42,0.42) -- (1.0,0.68);
\draw[curvepath]
  (bb) .. controls (2.5,2.7) and (3.5,3.0) .. (xx);
\node[font=\scriptsize, blue!60!black, align=center, anchor=south east]
  at (3.25,2.95) {on-manifold path\\ from $\baseline_{\mathrm{blur}}$};
\end{tikzpicture}
\caption{Baseline and path dependence for path-based attribution in a
two-feature input space. The shaded region is the support of the data
distribution. The straight-line path from a zero baseline $\baseline_0$
spends its early segment (dashed) outside $\mathrm{supp}(p_X)$, where
$\model$ extrapolates and gradients are unconstrained by training
data. A path from an in-distribution baseline
$\baseline_{\mathrm{blur}}$ that stays inside the support queries
$\model$ only where it was trained. Both decompose the same total,
$\sum_i \phi_i = \model(x) - \model(\baseline)$ for their respective
$\baseline$, but the per-feature split depends on the baseline and the
path.}
\label{fig:baseline-path}
\end{figure}

\subsection{Adaptive and Region-Based Paths}

The straight-line path is mathematically convenient but may be poorly
matched to the data manifold. It traverses regions of $\mathcal{X}$ where the model may behave
erratically, a well-known manifestation of gradient saturation in deep networks
means $\partial \model / \partial x_i \approx 0$ along long stretches
of the path~\cite{shrikumar2017deeplift,ancona2018gradient}, washing
out attributions for important features.

\textbf{Guided Integrated Gradients} (Guided
IG)~\cite{kapishnikov2021guidedig} replaces the straight-line path with
an adaptive trajectory that steers around low-gradient regions. At each
step, the path follows the direction of steepest absolute partial
derivative, restricted to features that have not yet been ``saturated''.
Guided IG retains completeness (it is still a path integral) but loses
the symmetry-preservation property: a path that depends on the gradient
landscape is no longer permutation-equivariant.

\textbf{XRAI}~\cite{kapishnikov2019xrai} attributes credit to
\emph{image regions}, with individual pixels grouped
by a segmentation procedure; each region is then assigned the sum of IG
attributions of its constituent pixels, divided by region area to give
an importance score. XRAI is best understood as an aggregation layer
over IG; it does not change the underlying attribution but improves
the human readability of image-domain explanations.

\textbf{Blur Integrated Gradients}~\cite{xu2020blurig} replaces the
path in input space with a path in \emph{scale space}: $\gamma(\alpha)$
is a sequence of progressively less blurred versions of $x$, with the
heavily blurred input as the baseline. This avoids the pathologies of
zero-baseline IG for images while keeping the path integral
interpretable as a decomposition of $\Delta f$. The trade-off is that
the resulting attribution is no longer additive in the original pixel
coordinates; it lives in scale space.

\subsection{Higher-Order Path Methods: Integrated Hessians}

To attribute joint effects of feature pairs, Janizek, Sturmfels, and
Lee~\cite{janizek2021integratedhessians} introduced Integrated Hessians
(IH). Where IG integrates $\partial \model / \partial x_i$ along a
single path from $\baseline$ to $x$, IH integrates the second mixed
partial $\partial^2 \model / \partial x_i \partial x_j$ along a
\emph{nested} pair of paths and yields an interaction attribution
$\Gamma_{ij}$ satisfying
\begin{equation}
\sum_{i, j} \Gamma_{ij}(x; \baseline) = \model(x) - \model(\baseline).
\end{equation}

\begin{result}[Janizek et al.~\cite{janizek2021integratedhessians}, Thm.~1]
\emph{Assumptions:} $\model$ is twice-differentiable along the nested
path; the mixed partials $\partial^2 \model / \partial x_i \partial x_j$
exist and are integrable. Under these assumptions, Integrated Hessians
is the unique attribution of second-order feature interactions
consistent with the IG axioms when applied to the gradient-of-IG
operator, as stated in the cited paper.
\end{result}

IH connects directly to the Shapley-Taylor interaction
index~\cite{dhamdhere2020shapleytaylor} (Section~\ref{sec:shapley}):
the discrete Shapley-Taylor index of order~2 is a coalition-sampling
approximation of the IH continuous integral.

\subsection{Conductance: Internal Path Attribution}

The path integral~\eqref{eq:ig} attributes credit to \emph{input}
features. Conductance~\cite{dhamdhere2018conductance} extends the same
construction to internal neurons:
\begin{equation}
\mathrm{Cond}_h(x; \baseline) = (x - \baseline)^\top \int_0^1 \frac{\partial \model}{\partial h}\bigg|_{\gamma(\alpha)} \frac{\partial h}{\partial x}\bigg|_{\gamma(\alpha)} d\alpha,
\end{equation}
where $h$ is the activation of a chosen hidden neuron. By construction,
conductances of all neurons in a layer sum to $\Delta f$, giving a
layer-wise decomposition consistent with the LRP conservation axiom
(Axiom~\ref{ax:lrp-conservation}) without imposing layer-specific
rules.

\subsection{Path Dependence: When Does the Path Matter?}

A central theoretical question is: how sensitive are IG and its
variants to the choice of path? Every method in this section is an
instance of the \emph{generalized path attribution}
\begin{equation}
\IG^{\gamma}_i(x; \baseline)
= \int_0^1
\frac{\partial \model}{\partial x_i}\bigl(\gamma(\alpha)\bigr)\,
\gamma_i'(\alpha)\, d\alpha,
\label{eq:path-general}
\end{equation}
for some differentiable $\gamma$ with $\gamma(0) = \baseline$ and
$\gamma(1) = x$; the straight line recovers~\eqref{eq:ig} and the
Aumann-Shapley value~\eqref{eq:aumann-shapley}. Summing
\eqref{eq:path-general} over $i$ and applying the chain rule as
in~\eqref{eq:ig-ftc} shows that every path-integral method of this
form satisfies completeness, provided $\model$ is differentiable along
$\gamma$ with integrable gradient and the path has the stated
endpoints: the total
$\int_0^1 \nabla \model(\gamma(\alpha))^\top \gamma'(\alpha)\, d\alpha
= \model(x) - \model(\baseline)$ depends only on the endpoints. The
\emph{per-feature} terms of~\eqref{eq:path-general}, however, are not
endpoint-determined: the decomposition that IG, EG, Guided IG, and
Blur IG return is path-dependent.

Two paths between the same $\baseline$ and $x$ that disagree on the
per-feature decomposition correspond to two different ways of
attributing the same total credit. This is not necessarily an error; it
reflects the fact that attribution is underdetermined without
additional assumptions. Under the assumptions and attribution class
used by Sundararajan et al.~\cite{sundararajan2017axiomatic}, the
straight-line path is singled out by symmetry-preservation and related
axioms, but other choices are defensible under different axiom
sets~\cite{lundstrom2025ig}.

\subsection{Summary}

Table~\ref{tab:path-methods} summarizes the path-based methods. The
common thread is that each method makes one of two changes to
IG~\eqref{eq:ig}: either it changes the baseline (EG, GradientSHAP), or
it changes the path (Guided IG, Blur IG, XRAI), or both
(IH adds an additional integration dimension). The choice of path or
baseline determines which axioms hold, which features are highlighted,
and how robust the attribution is to gradient saturation.

\begin{table*}[t]
\centering
\caption{Path-based attribution methods. All preserve completeness;
they differ in the choice of path and baseline, with corresponding
effects on the remaining axioms.}
\label{tab:path-methods}
\footnotesize
\begin{tabularx}{\textwidth}{@{}p{0.18\textwidth}p{0.16\textwidth}p{0.17\textwidth}p{0.15\textwidth}p{0.13\textwidth}X@{}}
\toprule
Method & Path & Baseline & Completeness & Interactions & Suited to \\
\midrule
Integrated Gradients~\cite{sundararajan2017axiomatic} & straight line & fixed $\baseline$ & yes & 1st order & tabular, image \\
Expected Gradients~\cite{erion2021expectedgradients} & straight line & $\E_{\baseline \sim p_X}$ & yes & 1st order & tabular, image \\
Guided IG~\cite{kapishnikov2021guidedig} & adaptive & fixed $\baseline$ & yes & 1st order & image (noisy IG) \\
Blur IG~\cite{xu2020blurig} & scale space & blurred input & yes & 1st order & image \\
XRAI~\cite{kapishnikov2019xrai} & inherits IG path & inherits IG baseline & yes (per region) & 1st order & image (segmented) \\
Integrated Hessians~\cite{janizek2021integratedhessians} & nested paths & fixed $\baseline$ & yes (sum over pairs) & 2nd order & feature interactions \\
Conductance~\cite{dhamdhere2018conductance} & straight line in $x$ & fixed $\baseline$ & yes (sum over neurons) & 1st order & internal neurons \\
\bottomrule
\end{tabularx}
\end{table*}

\section{Gradient and Backpropagation Attribution}
\label{sec:gradient}

Gradient and backpropagation methods derive an attribution from a
single backward pass through the model. They are the cheapest family of
attribution methods, typically one or two backward passes, and
historically were the first to be applied to deep networks. They are
also the family with the most diverse axiomatic profiles: some satisfy
completeness, some do not; some are implementation-invariant, some
are not; some are continuous in $x$, and a notable subset is so
discontinuous as to be operationally
unstable~\cite{ghorbani2019fragile,dombrowski2019manipulated}.

\subsection[Raw Saliency and Gradient-times-Input]{Raw Saliency and Gradient $\times$ Input}

The earliest gradient-based attribution is the
\emph{saliency map}~\cite{simonyan2013saliency,baehrens2010explain}:
\begin{equation}
\phi_i^{\mathrm{sal}}(x) = \left|\frac{\partial \model(x)}{\partial x_i}\right| \quad \text{or} \quad \phi_i^{\mathrm{sal}}(x) = \frac{\partial \model(x)}{\partial x_i}.
\label{eq:saliency}
\end{equation}
Saliency is the first-order Taylor coefficient of $\model$ at $x$. It
satisfies sensitivity-(b) and implementation invariance but \emph{not}
completeness or sensitivity-(a): a feature can change the prediction
substantially yet receive a gradient near zero whenever $\model$ is
locally flat (gradient saturation~\cite{shrikumar2017deeplift}).

The \emph{gradient $\times$ input} variant,
\begin{equation}
\phi_i^{\mathrm{gxi}}(x) = x_i \cdot \frac{\partial \model(x)}{\partial x_i},
\label{eq:grad-times-input}
\end{equation}
is the first-order term of the Taylor expansion of $\model$ around the
origin. It is one of the cheapest sensible attributions, and is the
$K = 1$ approximation of Integrated Gradients with $\baseline = \bm{0}$.
It satisfies completeness only when $\model$ is linear.

\subsection{SmoothGrad and Noise Averaging}

SmoothGrad~\cite{smilkov2017smoothgrad} smooths the saliency map by
averaging over noisy versions of the input:
\begin{equation}
\phi_i^{\mathrm{SG}}(x) = \E_{\epsilon \sim \mathcal{N}(0, \sigma^2 I)}\!\left[\left|\frac{\partial \model(x + \epsilon)}{\partial x_i}\right|\right].
\label{eq:smoothgrad}
\end{equation}
The averaging effectively replaces $\model$ with the Gaussian-smoothed
model
$\tilde{\model}_\sigma = \model * \mathcal{N}(0, \sigma^2 I)$, i.e.\
$\tilde{\model}_\sigma(x) = \E_{\epsilon}[\model(x + \epsilon)]$. For
$\model$ locally integrable with gradient of at most polynomial
growth, differentiation and expectation interchange, so
\begin{equation}
\nabla \tilde{\model}_\sigma(x)
= \E_{\epsilon \sim \mathcal{N}(0,\sigma^2 I)}
\bigl[\nabla \model(x + \epsilon)\bigr]:
\label{eq:smoothgrad-identity}
\end{equation}
the \emph{signed} SmoothGrad average is an exact gradient of a smoothed
model, and inherits whatever axioms hold for gradients of
$\tilde{\model}_\sigma$. The absolute-value form
in~\eqref{eq:smoothgrad} breaks this identity because $\E[|\cdot|] \neq
|\E[\cdot]|$, so absolute SmoothGrad is not the gradient of any
smoothed model. SmoothGrad
substantially reduces the high-frequency artefacts visible in raw
saliency maps for image classifiers, at the cost of a hyperparameter
$\sigma$ that controls the trade-off between locality and stability.

\textbf{VarGrad}~\cite{adebayo2018sanity} computes the variance rather
than the expectation:
$\phi_i^{\mathrm{VG}}(x) = \mathrm{Var}_{\epsilon}[\partial \model(x + \epsilon)/\partial x_i]$.
VarGrad highlights features for which the gradient is locally
\emph{unstable}, often providing complementary information to
SmoothGrad.

Both methods can be viewed as Monte Carlo estimators of the gradient
of a noise-smoothed model, and as such inherit the axiomatic profile of
that smoothed model. They satisfy implementation invariance but
typically not completeness, since the smoothed gradient does not
integrate to $\model(x)$ in general.

\subsection{Guided Backpropagation and Deconvolution}

Guided backpropagation~\cite{springenberg2015guidedbp} and deconvolution
networks~\cite{zeiler2014visualizing} modify the backward pass through
ReLU units. Standard backpropagation uses positive forward activations;
these methods additionally suppress gradients of opposite sign.

\begin{result}[Empirical finding of Adebayo et al.~\cite{adebayo2018sanity}]
\label{thm:gbp-sanity}
On standard image classifiers, guided backpropagation produces
attribution maps that are visually largely independent of the model's
parameters: replacing the weights with random values yields visually
similar attributions. We report this as an empirical result from the
cited paper, not as a formal theorem.
\end{result}

Result~\ref{thm:gbp-sanity} (the model-randomization sanity check) is
one of the strongest empirical critiques of guided backpropagation: an
attribution that is largely insensitive to model parameters cannot, by
itself, support strong claims about model-specific reasoning. We return
to the sanity-check framework in Section~\ref{sec:evaluation}.

\subsection{DeepLIFT}

DeepLIFT~\cite{shrikumar2017deeplift} replaces gradients with
\emph{multipliers} computed against a reference activation. For each
neuron $h$ with reference value $h_{\mathrm{ref}}$ and difference
$\Delta h = h - h_{\mathrm{ref}}$, the DeepLIFT contribution
$C_{\Delta x \Delta h}$ propagates through the network analogously to
the chain rule:
\begin{equation}
C_{\Delta x_i \Delta y} = \sum_{h} m_{x_i h} \cdot C_{\Delta h \Delta y},
\end{equation}
where $m_{x_i h} = C_{\Delta x_i \Delta h} / \Delta x_i$ is a
multiplier. DeepLIFT satisfies the summation-to-delta axiom
(Axiom~\ref{ax:deeplift-completeness}) by construction. Two distinct
rules, \emph{Rescale} (for monotone nonlinearities) and
\emph{RevealCancel} (for pairs of opposing effects), determine
the multiplier at each layer.

\begin{result}[Ancona et al.~\cite{ancona2018gradient}, \S 3]
\label{thm:deeplift-ig}
\emph{Assumptions:} the network is composed only of linear layers and
monotone elementwise nonlinearities (e.g., ReLU, sigmoid, tanh);
there are no skip connections, attention layers, concatenations, or
non-monotone operators; a single fixed baseline $\baseline$ is used.
Under these assumptions, DeepLIFT-Rescale and Integrated Gradients
coincide in the small-input-increment limit. When the assumptions
fail (Add, GroupNorm, attention, concatenation), the equivalence
breaks and the methods can yield different attributions.
\end{result}

Result~\ref{thm:deeplift-ig} establishes a key bridge: under restricted
architectures and propagation rules, DeepLIFT-Rescale and Integrated
Gradients can be viewed as closely related approximations of the same
path-based quantity. Outside those assumptions, they should be treated
as distinct attribution operators. They differ in computational profile
(one forward+backward pass for DeepLIFT vs.\ $K$ for IG) and in their
handling of non-monotone nonlinearities (the RevealCancel rule
addresses a case IG does not).

\subsection{Layer-Wise Relevance Propagation}

LRP~\cite{bach2015lrp} propagates a quantity called \emph{relevance}
backward through the network, subject to the conservation axiom
(Axiom~\ref{ax:lrp-conservation}). The relevance of an output neuron
is initialized to $\model(x)$; at each layer, it is decomposed
into contributions from the preceding layer using a propagation rule.
The most common rules are:
\begin{description}[style=nextline,leftmargin=1.5em]
\item[$z$-rule] $R_i = \sum_j \frac{z_{ij}}{\sum_{i'} z_{i'j}} R_j$,
where $z_{ij} = x_i w_{ij}$ is the pre-activation contribution.
Numerically unstable near zero.
\item[$\epsilon$-rule] adds a small constant $\epsilon$ to the
denominator to suppress numerically unstable terms.
\item[$\gamma$-rule] amplifies positive contributions with
multiplier $\gamma$, used for the upper layers of deep classifiers.
\end{description}
The \emph{LRP composition} strategy of Montavon
et al.~\cite{montavon2017deeptaylor} applies different rules to
different layers, producing attribution maps that are
empirically more faithful to the model's behaviour than any single rule
alone.

LRP is implementation-invariant only for the standard rules applied to
specific architecture families; bespoke rules can break the property.
It satisfies conservation by construction, hence completeness. It does
not in general satisfy Sensitivity-(b): an input feature that is
formally a dummy can still receive nonzero relevance under certain
rules.

\subsection{FullGrad}

FullGrad~\cite{srinivas2019fullgrad} extends gradient-based attribution
to incorporate \emph{bias} terms, which earlier methods ignored. The
key observation is that for a network with biases, the gradient
$\nabla_x \model$ alone does not satisfy completeness; the bias
contributions must be added explicitly. FullGrad computes
\begin{equation}
\phi^{\mathrm{FG}}_i(x) = x_i \frac{\partial \model}{\partial x_i} + \sum_{\ell} f_\ell^{\mathrm{bias}}(x),
\end{equation}
where $f_\ell^{\mathrm{bias}}$ is the contribution from biases at
layer $\ell$. FullGrad satisfies completeness exactly:
$\sum_i \phi_i^{\mathrm{FG}} = \model(x)$. This target includes the bias
contribution at $\baseline = \bm{0}$ and therefore differs from $\Delta f$.

\subsection{The CAM Family: From Class Activation to Shapley-CAM}

Class Activation Mapping (CAM)~\cite{zhou2016cam} was originally
developed for global-average-pooled CNNs as a coarse spatial map of
which regions activate a given class. Grad-CAM~\cite{selvaraju2017gradcam}
generalized CAM to arbitrary CNN architectures by using gradients of
the class score with respect to the final convolutional feature map:
\begin{equation}
L^{\mathrm{GC}}_{c,(i,j)} = \mathrm{ReLU}\!\left(\sum_k \alpha_k^c \, A^k_{(i,j)}\right), \quad \alpha_k^c = \frac{1}{Z} \sum_{i,j} \frac{\partial \model_c}{\partial A^k_{(i,j)}}.
\label{eq:gradcam}
\end{equation}
Here $A^k$ is the $k$-th feature map of the target convolutional layer
and $Z$ is the number of spatial positions. The weights $\alpha_k^c$
average the class-score gradient over space, and the final map is a
ReLU-rectified sum of the feature maps.

\textbf{Grad-CAM++}~\cite{chattopadhyay2018gradcampp} replaces the
average pooling with a weighted average that emphasizes positive
contributions:
\begin{equation}
\begin{aligned}
\alpha_k^{c, \mathrm{++}}
&= \sum_{i,j}
\frac{g^{(2)}_{kij}}
{2g^{(2)}_{kij} + \sum_{a,b} A^k_{(a,b)} g^{(3)}_{kij}},\\
g^{(m)}_{kij}
&= \frac{\partial^m \model_c}{\partial (A^k_{(i,j)})^m}.
\end{aligned}
\end{equation}
The motivation is to handle multiple instances of the target class.

\textbf{Score-CAM}~\cite{wang2020scorecam} eliminates gradients entirely
by computing weights through perturbation: each feature map is
upsampled into a soft mask, applied to the input, and the resulting
class score determines the weight. This avoids gradient saturation but
adds $O(K)$ forward passes per attribution, where $K$ is the number of
channels.

\textbf{Ablation-CAM}~\cite{desai2020ablationcam} computes weights by
explicitly zeroing one channel at a time:
$\alpha_k^c = (\model_c - \model_c^{\setminus k}) / \model_c$, where
$\model_c^{\setminus k}$ is the prediction with channel $k$ ablated.

\textbf{LayerCAM}~\cite{jiang2021layercam} aggregates Grad-CAM-style
maps across multiple layers, including layers before the final convolution,
giving finer spatial resolution.

\textbf{Eigen-CAM}~\cite{muhammad2020eigencam} uses the principal
components of the feature map matrix as weights, removing the
class-conditional gradient entirely.

\textbf{HiResCAM}~\cite{draelos2020hirescam} corrects a known failure
of Grad-CAM: the ReLU-and-average step in~\eqref{eq:gradcam} can
produce attribution maps that do not faithfully reflect the model's
prediction. HiResCAM eliminates the spatial pooling and applies the
gradient directly, restoring faithfulness at the cost of some spatial
smoothness.

\textbf{Shap-CAM}~\cite{zheng2022shapcam} reframes the CAM weighting as
a Shapley value over channels, providing an axiomatic
justification for Score-CAM's perturbation-based weighting.

What unites the CAM family is a \emph{spatial projection} of
attribution onto the last convolutional layer. In their usual form, CAM
variants are not pixel-level complete decompositions of
$\model(x)-\model(\baseline)$; they are intermediate-layer attributions
that are then upsampled to image resolution.

\subsection{Transformer Attribution: Attention Is Not Enough}

For transformer architectures, the most-cited attribution candidate is
the attention map itself. The mathematical critique
of~\cite{jain2019attentionnot,serrano2019attention} showed that
attention weights are not generally consistent with model
behaviour: alternative attention distributions can produce identical
or near-identical predictions, so attention is not uniquely identified
by the model's output.

Wiegreffe and Pinter~\cite{wiegreffe2019notnot} treat attention as one
plausible explanatory hypothesis without claiming that it is unique.
They propose testing whether the attention pattern is plausible under
the model.

Two attribution methods have since become standard for transformers:

\textbf{Attention rollout}~\cite{abnar2020flow} aggregates attention
weights across layers by treating each layer's attention matrix as a
transition probability and computing the rollout
$\tilde{A}^L = \prod_{\ell=1}^{L} (A^{(\ell)} + I) / 2$. The product
captures the iterated effect of attention across the depth of the
model.

\textbf{Transformer relevance
propagation}~\cite{chefer2021transformer} extends LRP to transformers,
distributing the class score backward through both attention and
feed-forward components. It satisfies a conservation axiom at each
layer, restoring the property that attention alone does not.

\subsection{Summary}

Gradient and backpropagation methods range from raw saliency (one
forward+backward pass, no axioms beyond sensitivity-(b)) to LRP and
DeepLIFT (multiple rules, conservation, implementation invariance under
the right conditions). They are united by their differentiability
assumption and their computational efficiency, but divided on which
axioms they satisfy and on how they behave under known sanity
checks. Table~\ref{tab:gradient-methods} summarizes the family.

\begin{table*}[t]
\centering
\caption{Gradient and backpropagation attribution methods. ``Cost''
is in backward passes; ``Sanity'' is the model-randomization sanity
check of~\cite{adebayo2018sanity}: pass means the attribution changes
substantially when model weights are randomized.}
\label{tab:gradient-methods}
\footnotesize
\begin{tabular}{@{}llllllr@{}}
\toprule
Method & Derivative & Completeness & Impl.\ Inv. & Sanity & Noise smoothing & Cost \\
\midrule
Saliency~\cite{simonyan2013saliency} & 1st & no & yes & pass & no & 1 \\
Gradient $\times$ Input & 1st & only if linear & yes & pass & no & 1 \\
SmoothGrad~\cite{smilkov2017smoothgrad} & 1st (smoothed) & no & yes & pass & yes & $N$ \\
VarGrad~\cite{adebayo2018sanity} & 1st (variance) & no & yes & pass & yes (variance) & $N$ \\
Guided BP~\cite{springenberg2015guidedbp} & 1st (rule-modified) & no & no & \emph{fail} & no & 1 \\
Deconvnet~\cite{zeiler2014visualizing} & 1st (rule-modified) & no & no & \emph{fail} & no & 1 \\
DeepLIFT~\cite{shrikumar2017deeplift} & multipliers & yes & yes (Rescale) & pass & no & 1 \\
LRP~\cite{bach2015lrp} & rule-based & yes (conservation) & yes (standard rules) & pass & no & 1 \\
FullGrad~\cite{srinivas2019fullgrad} & 1st + bias & yes (vs.\ $0$ baseline) & yes & pass & no & 1 \\
Grad-CAM~\cite{selvaraju2017gradcam} & 1st of last conv & no (region only) & layer-dep. & partial & no & 1 \\
Grad-CAM++~\cite{chattopadhyay2018gradcampp} & 1st--3rd of last conv & no & layer-dep. & partial & no & 1 \\
Score-CAM~\cite{wang2020scorecam} & none (forward only) & no & layer-dep. & pass & no & $K$ fwd \\
HiResCAM~\cite{draelos2020hirescam} & 1st (no pooling) & yes (per layer) & layer-dep. & pass & no & 1 \\
\bottomrule
\end{tabular}
\end{table*}

\section{Perturbation and Occlusion Attribution}
\label{sec:perturbation}

Perturbation methods sidestep the question of value functions, paths,
and gradients by directly measuring how the prediction changes when
parts of the input are modified. They are conceptually simple
(``occlude a feature and see what happens'') and fully
model-agnostic, requiring only black-box access to $\model$. Their
mathematical content lies in how perturbations are chosen, how
predictions on perturbed inputs are aggregated, and how their
attributions should be interpreted given the off-manifold inputs they
typically produce.

\subsection{Occlusion Sensitivity and Prediction Difference}

The simplest perturbation method is \emph{occlusion}, introduced for
CNNs by Zeiler and Fergus~\cite{zeiler2014visualizing}: slide a fixed
patch across the input image, record the change in the prediction at
each position, and use the resulting map as an attribution. Formally,
let $\mathcal{P}_p(x)$ denote $x$ with a patch of features
$p \subseteq \featureset$ replaced by a reference value:
\begin{equation}
\phi^{\mathrm{occ}}_p(x) = \model(x) - \model(\mathcal{P}_p(x)).
\label{eq:occlusion}
\end{equation}
Occlusion is a single-coalition Shapley estimate: it computes the
marginal contribution of the patch $p$ to the full feature set
$\featureset$, ignoring all other coalitions. It is unbiased when the
features outside $p$ are independent of those inside, and biased
otherwise.

The \emph{Prediction Difference
Analysis} of Zintgraf et al.~\cite{zintgraf2017pda} replaces the
deterministic patch with a conditional expectation: features are
``removed'' by marginalizing over their conditional distribution given
the rest of the input. This addresses the off-manifold issue at the
cost of requiring a conditional density model.

\subsection{LIME: Local Linear Surrogates}

LIME~\cite{ribeiro2016lime} fits a local linear surrogate to
$\model$ in a neighbourhood of $x$. Given a similarity kernel
$\pi_x(\mask)$ and a perturbation distribution over masks, LIME solves
\begin{equation}
g^* = \arg\min_{g \in \mathcal{G}} \sum_{\mask} \pi_x(\mask) \bigl(\model(\mathcal{P}_{\mask}(x)) - g(\mask)\bigr)^2 + \Omega(g),
\label{eq:lime}
\end{equation}
where $\mathcal{G}$ is a family of interpretable models (typically
sparse linear) and $\Omega$ is a complexity penalty. The coefficients
of $g^*$ are the attributions.

LIME is a general additive surrogate method: any choice of
mask distribution and kernel yields a method in the
class~\eqref{eq:additive-surrogate}. KernelSHAP is the special case in
which the kernel is $\pi_{\mathrm{Shap}}$ of~\eqref{eq:shapley-kernel}
and the regression is unregularized; in that case the surrogate
coefficients coincide with the Shapley
values~\cite{lundberg2017unified}. Different kernels yield different
attributions, and the LIME kernel (exponential of inverse cosine
distance, in the original paper) does \emph{not} satisfy the Shapley
axioms.

MAPLE~\cite{plumb2018maple} is a supervised-neighbourhood variant of
the same local-surrogate idea. Instead of sampling an unsupervised
neighbourhood around $x$, MAPLE uses tree-ensemble structure to weight
training points and then fits a local linear model under that induced
neighbourhood. This gives MAPLE a clearer statistical object than a
generic proximity kernel: the local explanation is tied to the
predictive neighbourhood learned by random forests~\cite{breiman2001randomforests}
or boosted trees~\cite{friedman2001gbm}. It also illustrates a
general lesson for local surrogates: locality is not a purely geometric
choice, but a modelling assumption about which perturbations should
stand in for nearby counterfactuals.

\subsection{Anchors: Rule-Based Local Explanations}

\textbf{Scope note.} Anchors lie at the boundary of the
local-additive scope of this survey: their output is a rule
predicate, not a per-feature attribution vector, so the linear-in-mask
surrogate of~\eqref{eq:additive-surrogate} does not apply directly.
We include Anchors in this section because the underlying
\emph{perturbation distribution} is closely related to that of LIME
and because Anchors is widely used as a baseline against
attribution methods; we mark it as a boundary case in the axiom
matrix (Table~\ref{tab:axiom-matrix}).

Anchors~\cite{ribeiro2018anchors} step away from the additive surrogate
form entirely and produce a
set of rules $A$ such that, conditional on the rules holding,
$\model(x') = \model(x)$ with high probability:
\begin{equation}
\Prob\bigl(\model(x') = \model(x) \,\big|\, A(x') = 1\bigr) \geq 1 - \delta,
\end{equation}
for some tolerance $\delta$ and a coverage condition on the rule. The
rules are themselves the explanation; they have no real-valued
attribution per feature. Anchors and LIME together span the local-surrogate
design space: a local linear model versus a local
rule.

\subsection{Meaningful Perturbations and Extremal Masks}

Fong and Vedaldi~\cite{fong2017meaningful} introduced
\emph{meaningful perturbations}, which optimize a smooth mask
$\mask \in [0,1]^d$ that maximally
suppresses the prediction subject to a sparsity constraint:
\begin{equation}
\min_{\mask} \; \model(\mathcal{P}_{\mask}(x)) + \lambda \|\mask\|_1 + \mu \, \mathrm{TV}(\mask),
\label{eq:meaningful-pert}
\end{equation}
where $\mathrm{TV}$ is a total-variation regularizer. The optimal
mask identifies the regions of $x$ whose removal most disrupts the
prediction, yielding a localized form of attribution.

The follow-up Extremal
Perturbations~\cite{fong2019extremal} replaces the unconstrained
$\ell_1$ objective with a hard area constraint: ``find the smallest
mask of area $a$ that maximizes/minimizes the prediction''. This avoids
sensitivity to the Lagrange multiplier $\lambda$ and yields more
interpretable masks.

\subsection{RISE and Random-Mask Methods}

RISE~\cite{petsiuk2018rise} averages over a large number of random masks:
\begin{equation}
\phi^{\mathrm{RISE}}_i(x) = \frac{1}{N} \sum_{n=1}^{N} \model(\mathcal{P}_{\mask^{(n)}}(x)) \, \mask^{(n)}_i \, / \, p,
\label{eq:rise}
\end{equation}
where $p$ is the probability that feature $i$ is unmasked. RISE is
\emph{embarrassingly parallel} and requires no gradient access, making
it appealing for true black-box settings. It is a Monte Carlo
estimator of the marginal expectation of $\model$ with respect to
the mask distribution, normalized to give a per-feature score.

\subsection{Counterfactual Generation: FIDO-CA (boundary case)}

\textbf{Scope note.} FIDO-CA produces counterfactual generations and
does not yield per-feature attribution scores; we include it here
because the underlying optimisation formulation
(mask + generative completion) lies on the same axis as the
perturbation-distribution choice made by LIME and RISE, and because
its on-manifold mask is a constructive answer to the off-manifold
failure mode of mask-based attribution. Strictly speaking, the
contribution sits in the counterfactual-explanation literature
catalogued in Section~\ref{sec:limitations}.

Chang et al.~\cite{chang2019counterfactual} introduced FIDO-CA, which
generates counterfactual perturbations using a generative model. Where
RISE and meaningful perturbations replace masked features with
constants or blurred values, FIDO-CA fills them with samples from a
conditional generator $G(\mask, x)$:
\begin{equation}
\phi^{\mathrm{FIDO}}_i(x) = \arg\min_{\mask} \; -\log \model_c\bigl(G(\mask, x)\bigr) + \lambda \|\mask\|_1.
\end{equation}
The motivation is that the resulting perturbed inputs are on-manifold,
addressing one of the principal failure modes of mask-based
methods (Section~\ref{sec:failure}).

\subsection{Real-Time Saliency: Learned Mask Predictors}

Dabkowski and Gal~\cite{dabkowski2017realtime} trained a separate
network $g_\psi$ to predict the optimal mask of~\eqref{eq:meaningful-pert}:
\begin{equation}
g_\psi(x) \approx \arg\max_{\mask} \; \model_c(\mathcal{P}_{\mask}(x)) - \lambda \|\mask\|_1.
\end{equation}
The learned mask predictor gives near-instant attributions at inference
time but introduces an additional model whose own validity must be
verified.

\subsection{Information-Bottleneck Attribution}

Schulz et al.~\cite{schulz2020information} cast mask-based attribution
as restricting the information flow through the network. They add
Gaussian noise to intermediate activations and optimize the
noise variance per spatial location to maximize the suppression of
information about the input while preserving the prediction. The
resulting per-location noise scale is the attribution. The framing
exposes a duality between mask-based attribution and information
bottlenecks that is independently interesting; the cost is one
optimization per attribution.

\subsection{Insertion / Deletion Evaluation}

Petsiuk et al.~\cite{petsiuk2018rise} also introduced the
\emph{insertion / deletion} curves that have since become a standard
evaluation tool (Section~\ref{sec:evaluation}). Features are added or
removed in order of attribution magnitude; the area under the resulting
prediction-versus-step curve quantifies how well the attribution
identifies features that matter to the model. These curves are not an
attribution method per se but a meta-evaluation built on any underlying
attribution.

\subsection{Summary}

Perturbation methods are the natural complement to gradient methods.
They make no differentiability assumption and require no
backpropagation, but pay for that generality in two ways: many forward
passes per attribution, and a strong dependence on the perturbation
distribution. The choice of how to ``remove'' a feature (by zeroing,
blurring, sampling from $p_X$, sampling from $p(X \mid x_{\bar{\coalition}})$,
or generating with $G$) determines whether the resulting attribution
is interventional, conditional, or on-manifold, and is a direct
counterpart of the value-function choice in Shapley methods
(Section~\ref{sec:shapley}). The same axes therefore organize both
families.

\section{Mathematical Comparison Framework}
\label{sec:comparison}

\begin{figure*}[t]
\centering
\begin{tikzpicture}[
  font=\small,
  root/.style={draw=black!60, rounded corners=2pt, inner xsep=8pt,
    inner ysep=5pt, fill=gray!12},
  family/.style={draw=black!55, rounded corners=2pt, text width=3.6cm,
    align=center, inner sep=5pt},
  object/.style={text width=3.6cm, align=center, font=\scriptsize\itshape},
  edge/.style={-{Latex[length=2mm]}, semithick, black!60}
]
\node[root] (root) at (0,0) {Local additive feature attribution
  $\attribution(\model, x, \baseline) \in \R^d$};
\node[family, fill=blue!7]   (shap) at (-6.6,-2.45)
  {\textbf{Shapley /\\ cooperative game}\\[1.2mm]
  \scriptsize Exact Shapley\\ KernelSHAP\\ TreeSHAP\\ DeepSHAP \,/\, GradientSHAP};
\node[family, fill=teal!8]   (path) at (-2.2,-2.45)
  {\textbf{Path integral}\\[1.2mm]
  \scriptsize Integrated Gradients\\ Expected Gradients\\ Guided IG \,/\, Blur IG\\ Integrated Hessians};
\node[family, fill=orange!10] (grad) at (2.2,-2.45)
  {\textbf{Gradient /\\ backpropagation}\\[1.2mm]
  \scriptsize Saliency \,/\, SmoothGrad\\ DeepLIFT\\ LRP \,/\, FullGrad\\ Grad-CAM$^{\ast}$};
\node[family, fill=purple!8] (pert) at (6.6,-2.45)
  {\textbf{Perturbation /\\ surrogate}\\[1.2mm]
  \scriptsize Occlusion\\ LIME\\ RISE\\ Extremal perturbations};
\node[object, below=1.5mm of shap]
  {value function $\valuefn(\coalition)$,\\ Sec.~\ref{sec:shapley}};
\node[object, below=1.5mm of path]
  {path $\gamma$ and baseline $\baseline$,\\ Sec.~\ref{sec:path}};
\node[object, below=1.5mm of grad]
  {derivative / conservation rule,\\ Sec.~\ref{sec:gradient}};
\node[object, below=1.5mm of pert]
  {perturbation distribution $p_{\mask}$,\\ Sec.~\ref{sec:perturbation}};
\draw[edge] (root.south) -- ++(0,-0.45) -| (shap.north);
\draw[edge] (root.south) -- ++(0,-0.45) -| (path.north);
\draw[edge] (root.south) -- ++(0,-0.45) -| (grad.north);
\draw[edge] (root.south) -- ++(0,-0.45) -| (pert.north);
\end{tikzpicture}
\caption{Taxonomy of local additive feature attribution methods used
throughout this survey, with the mathematical object that each family
primarily manipulates (italics): a cooperative-game value function, a
path integral with its baseline, a backpropagation-style derivative or
conservation rule, or a black-box perturbation distribution. Methods
marked $^{\ast}$ (CAM family) are gradient-based but produce spatial
attributions through an intermediate-layer projection.
Section~\ref{sec:comparison} places all four families in a common
axiom-by-method comparison.}
\label{fig:taxonomy}
\end{figure*}
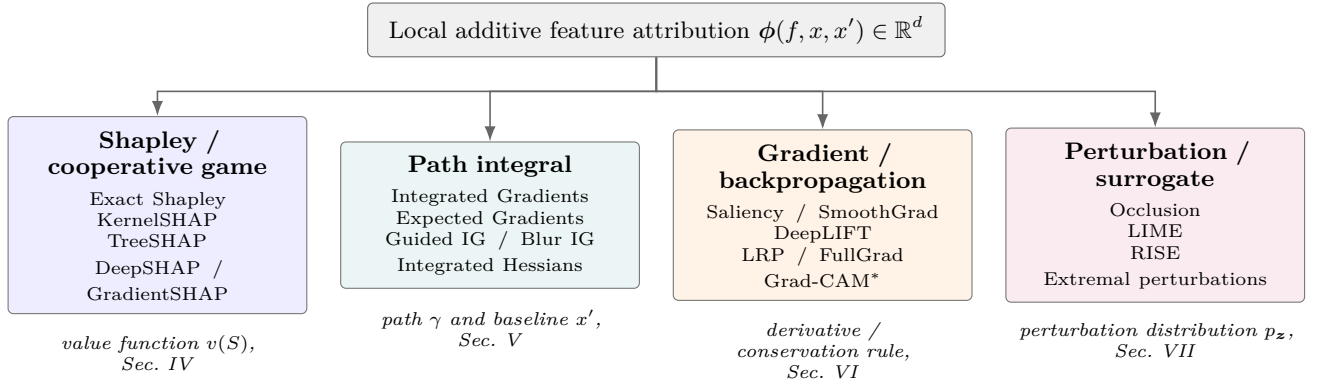

This section presents the main comparison table of the survey. Having
surveyed Shapley, path-based, gradient/backpropagation, and
perturbation methods, we now place them in a single mathematical frame.
We compare them along five axes: axioms satisfied, value function
choice, baseline distribution, computational complexity, and known
equivalences/reductions. The centerpiece is the axiom-by-method
matrix (Table~\ref{tab:axiom-matrix}) recording which axioms each
method satisfies. The matrix illustrates that many of the field's central
disagreements, examined as failure modes in
Section~\ref{sec:failure}, arise from incompatible axiom subsets, not from
implementation defects.

\subsection{Axes of Comparison}

We compare methods along the following axes:

\begin{description}[style=nextline,leftmargin=1.5em]
\item[Axioms satisfied.] Catalogued in Section~\ref{sec:axioms};
collected in Table~\ref{tab:axiom-matrix}.
\item[Value function choice.] Interventional, conditional, or
single-reference (Section~\ref{sec:formulation}).
\item[Baseline / path / perturbation.] The specific reference, path, or
perturbation distribution used.
\item[Computational complexity.] In flops, forward/backward passes, or
samples, as a function of $d$ and any approximation parameter.
\item[Differentiability assumption.] Required for gradient/path methods,
not for Shapley or perturbation methods.
\item[Causal assumptions.] Whether the method's output admits a causal
interpretation, and under what conditions.
\item[Stability / scalability.] Sensitivity to small input changes;
behaviour as $d \to \infty$.
\end{description}

\subsection{Axiom Satisfaction Matrix}

Table~\ref{tab:axiom-matrix} cross-references methods (rows) against
axioms (columns). A \checkmark{} means the method satisfies the axiom
unconditionally under the assumptions of its original paper; an
$\circ$ marks a conditional claim (specific architecture, sampling
limit, propagation rule, or method variant); a blank means the
axiom is not satisfied in general; \emph{N/A} marks axioms that are
inapplicable to the method's output type. The matrix records
\emph{mathematical} axioms only. Empirical evidence about
sanity-check behaviour is implementation- and architecture-dependent
and lies outside the axiomatic claims. It is reported
separately in Table~\ref{tab:empirical-evidence}, following the
critique that sanity checks are properties of an
implementation tested on a dataset, not of a method
formula~\cite{adebayo2018sanity,tomsett2020sanity}.

\begin{table*}[t]
\centering
\caption{Axiom satisfaction matrix. \checkmark = satisfied
unconditionally under the stated original assumptions; $\circ$ =
conditional (architecture, limit, rule, or method variant); blank =
not satisfied; N/A = inapplicable to the method's output
type. Abbreviations: Comp = completeness; S(a)/S(b) =
Sensitivity-(a)/(b); II = implementation invariance; Cons =
consistency; Mono = monotonicity; Sym = symmetry-preservation; Lin
= linearity; Cont = continuity in $x$. Per-cell justifications are
in Appendix~\ref{app:axiom-justification}. The entries refer to
canonical formulations unless otherwise stated; software
implementations may differ.}
\label{tab:axiom-matrix}
\small
\begin{tabular}{@{}lccccccccc@{}}
\toprule
Method & Comp & S(a) & S(b) & II & Cons & Mono & Sym & Lin & Cont \\
\midrule
\multicolumn{10}{l}{\textit{Shapley family}} \\
Exact Shapley~\cite{shapley1953value} & \checkmark & \checkmark & \checkmark & \checkmark & \checkmark & \checkmark & \checkmark & \checkmark & $\circ$ \\
KernelSHAP~\cite{lundberg2017unified} & \checkmark & \checkmark & \checkmark & \checkmark & $\circ$ & $\circ$ & \checkmark & \checkmark & $\circ$ \\
TreeSHAP~\cite{lundberg2020treeshap} & \checkmark & \checkmark & \checkmark & \checkmark & \checkmark & \checkmark & \checkmark & \checkmark & $\circ$ \\
DeepSHAP~\cite{lundberg2017unified} & $\circ$ & $\circ$ & \checkmark & $\circ$ & $\circ$ & & \checkmark & \checkmark & $\circ$ \\
GradientSHAP~\cite{erion2021expectedgradients} & \checkmark & \checkmark & \checkmark & \checkmark & $\circ$ & & \checkmark & \checkmark & $\circ$ \\
\midrule
\multicolumn{10}{l}{\textit{Path family}} \\
Integrated Gradients~\cite{sundararajan2017axiomatic} & \checkmark & \checkmark & \checkmark & \checkmark & & & \checkmark & \checkmark & \checkmark \\
Expected Gradients~\cite{erion2021expectedgradients} & \checkmark & \checkmark & \checkmark & \checkmark & & & \checkmark & \checkmark & \checkmark \\
Guided IG~\cite{kapishnikov2021guidedig} & \checkmark & \checkmark & \checkmark & $\circ$ & & & & \checkmark & \checkmark \\
Blur IG~\cite{xu2020blurig} & \checkmark & \checkmark & \checkmark & \checkmark & & & & \checkmark & \checkmark \\
Integrated Hessians~\cite{janizek2021integratedhessians} & \checkmark & \checkmark & \checkmark & \checkmark & & & \checkmark & \checkmark & \checkmark \\
\midrule
\multicolumn{10}{l}{\textit{Gradient \& backpropagation family}} \\
Saliency~\cite{simonyan2013saliency} & & & \checkmark & \checkmark & & & & \checkmark & $\circ$ \\
Grad $\times$ Input & $\circ$ & $\circ$ & \checkmark & \checkmark & & & & \checkmark & $\circ$ \\
SmoothGrad~\cite{smilkov2017smoothgrad} & & & \checkmark & \checkmark & & & & \checkmark & \checkmark \\
Guided BP~\cite{springenberg2015guidedbp} & & & & & & & & & $\circ$ \\
Deconvnet~\cite{zeiler2014visualizing} & & & & & & & & & $\circ$ \\
DeepLIFT~\cite{shrikumar2017deeplift} & \checkmark & \checkmark & \checkmark & $\circ$ & & & \checkmark & \checkmark & \checkmark \\
LRP~\cite{bach2015lrp} & \checkmark & $\circ$ & $\circ$ & $\circ$ & & & & \checkmark & \checkmark \\
FullGrad~\cite{srinivas2019fullgrad} & \checkmark & \checkmark & \checkmark & \checkmark & & & & \checkmark & \checkmark \\
Grad-CAM~\cite{selvaraju2017gradcam} & & $\circ$ & $\circ$ & $\circ$ & & & & \checkmark & \checkmark \\
HiResCAM~\cite{draelos2020hirescam} & $\circ$ & \checkmark & \checkmark & $\circ$ & & & & \checkmark & \checkmark \\
Score-CAM~\cite{wang2020scorecam} & & \checkmark & $\circ$ & $\circ$ & & & & \checkmark & \checkmark \\
\midrule
\multicolumn{10}{l}{\textit{Perturbation family}} \\
Occlusion~\cite{zeiler2014visualizing} & & \checkmark & $\circ$ & \checkmark & & & \checkmark & \checkmark & \checkmark \\
LIME~\cite{ribeiro2016lime} & $\circ$ & \checkmark & $\circ$ & \checkmark & & & & \checkmark & \checkmark \\
Meaningful Pert.~\cite{fong2017meaningful} & & \checkmark & $\circ$ & \checkmark & & & & & $\circ$ \\
RISE~\cite{petsiuk2018rise} & & \checkmark & $\circ$ & \checkmark & & & \checkmark & \checkmark & \checkmark \\
\midrule
\multicolumn{10}{l}{\textit{Boundary cases (not local additive on $\R^d$)}} \\
Anchors~\cite{ribeiro2018anchors} & N/A & \checkmark & \checkmark & \checkmark & & & & N/A & N/A \\
SAGE~\cite{covert2020sage} (global) & N/A & N/A & \checkmark & \checkmark & $\circ$ & & \checkmark & \checkmark & N/A \\
\bottomrule
\end{tabular}
\end{table*}

\begin{table*}[t]
\centering
\caption{Empirical-evidence companion to Table~\ref{tab:axiom-matrix}.
Each row reports published empirical evidence that the method
passes (\checkmark) or fails (\xmark) the model-randomization sanity
check~\cite{adebayo2018sanity} and the data-randomization sanity
check~\cite{adebayo2018sanity}. Entries marked $\circ$ are
partial or method-variant-dependent (e.g., Grad-CAM
variants differ). Blank evidence cells make no randomization-check
claim for that method; blank source cells are included only for
matrix completeness.
Sanity-check behaviour depends on implementation, architecture, and
dataset; entries here are not axiomatic claims.}
\label{tab:empirical-evidence}
\footnotesize
\begin{tabularx}{\textwidth}{@{}p{0.20\textwidth}p{0.18\textwidth}p{0.18\textwidth}X@{}}
\toprule
Method & Model-rand.\ evidence & Data-rand.\ evidence & Source(s) \\
\midrule
Saliency & \checkmark & \checkmark & \cite{adebayo2018sanity} \\
Grad $\times$ Input & \checkmark & \checkmark & \cite{adebayo2018sanity} \\
SmoothGrad & \checkmark & \checkmark & \cite{adebayo2018sanity} \\
Guided BP & \xmark & \xmark & \cite{adebayo2018sanity} \\
Deconvnet & \xmark & & \cite{adebayo2018sanity} \\
DeepLIFT (Rescale) & \checkmark & & \cite{adebayo2018sanity} \\
LRP-$\epsilon$ & \checkmark & & \cite{adebayo2018sanity} \\
Integrated Gradients & \checkmark & \checkmark & \cite{adebayo2018sanity} \\
Grad-CAM family & $\circ$ & & \cite{adebayo2018sanity,draelos2020hirescam} \\
HiResCAM & \checkmark & & \cite{draelos2020hirescam} \\
KernelSHAP / TreeSHAP & & & \\
LIME & & & \\
RISE & & & source reports perturbation metrics, not randomization checks \\
\bottomrule
\end{tabularx}
\end{table*}

\subsection{Complexity Landscape}

Table~\ref{tab:complexity} collects the asymptotic computational costs.
The methods span seven orders of magnitude: from a single backward
pass for raw saliency to $O(2^d)$ for exact Shapley.

\begin{table}[t]
\centering
\caption{Computational complexity of representative methods.
$\mathrm{net}$ denotes one forward+backward pass through the model.
$T, L, D$ are the number of trees, leaves, and tree depth; $K$ is the
number of path steps; $M$ the number of samples; $N$ the number of
SmoothGrad noise draws.}
\label{tab:complexity}
\footnotesize
\begin{tabularx}{\columnwidth}{@{}p{0.42\columnwidth}X@{}}
\toprule
Method & Complexity \\
\midrule
Saliency / Grad $\times$ Input & $O(\mathrm{net})$ \\
SmoothGrad / VarGrad & $O(N \cdot \mathrm{net})$ \\
DeepLIFT, LRP, FullGrad & $O(\mathrm{net})$ \\
Grad-CAM family & $O(\mathrm{net})$ \\
Score-CAM & $O(K \cdot \mathrm{net}_{\mathrm{fwd}})$ \\
Integrated Gradients & $O(K \cdot \mathrm{net})$ \\
Expected Gradients / GradientSHAP & $O(K \cdot M \cdot \mathrm{net})$ \\
Integrated Hessians & $O(K^2 \cdot \mathrm{net})$ \\
TreeSHAP & $O(T L D^2)$ \\
KernelSHAP (sampled) & $O(M \cdot d \cdot \mathrm{net}_{\mathrm{fwd}})$ \\
Exact Shapley & $O(2^d \cdot \mathrm{net}_{\mathrm{fwd}})$ \\
LIME (sampled) & $O(M \cdot \mathrm{net}_{\mathrm{fwd}})$ \\
RISE & $O(N \cdot \mathrm{net}_{\mathrm{fwd}})$ \\
Anchors & $O(M \cdot \mathrm{net}_{\mathrm{fwd}})$ (rule search) \\
\bottomrule
\end{tabularx}
\end{table}

\subsection{Equivalence and Reduction Theorems}

Several methods that appear distinct in the literature are
mathematically equivalent or related by explicit reductions; the
differences in notation between the original papers tend to hide this.

\begin{result}[KernelSHAP and exact Shapley, Lundberg \& Lee~\cite{lundberg2017unified}, Thm.~2]
\label{prop:kernel-shap}
\emph{Assumptions:} the interventional value function
\eqref{eq:value-interventional} is used; the weighted least-squares
regression in~\eqref{eq:additive-surrogate} is solved exactly on the
full mask distribution (no sampling). Under these assumptions, the
KernelSHAP solution equals the exact Shapley
value~\eqref{eq:shapley-value}. Finite-sample bias and variance are
characterized in~\cite{covert2021kernelshap,chen2023algorithms}.
\end{result}

\begin{result}[DeepLIFT-Rescale and IG, Ancona et al.~\cite{ancona2018gradient}, \S 3]
\label{prop:deeplift-ig}
\emph{Assumptions:} linear layers and monotone elementwise
nonlinearities; no skip connections, concatenations, or non-monotone
operators; single fixed baseline. Under these assumptions, DeepLIFT
with the Rescale rule and Integrated Gradients (with $K \to \infty$)
yield equivalent attributions in the small-input-increment limit.
The reduction is sensitive to architectural assumptions and does not
extend automatically to transformer or residual networks.
\end{result}

\begin{result}[Grad-CAM as class-conditional gradient projection; reformulation, not reduction]
\label{prop:gradcam}
\emph{Assumptions:} CNN with global average pooling followed by a
linear classification head. Under these assumptions, the Grad-CAM
map of~\eqref{eq:gradcam} is a rewriting (not a reduction) of a
class-conditional spatial average of the final-feature-map
gradient~\cite{selvaraju2017gradcam}, and coincides with the
original CAM~\cite{zhou2016cam} on the GAP-then-linear architecture.
We label this a \emph{reformulation} because the two computations
yield the same output by construction; it is not a reduction between
distinct methods.
\end{result}

\begin{result}[LRP-$\epsilon$ and gradient-$\times$-input, Ancona et al.~\cite{ancona2018gradient}, \S 3.2--3.3]
\label{prop:lrp-ig}
\emph{Assumptions:} deep ReLU network with no bias terms; LRP with
the $\epsilon$-rule applied uniformly to every layer; the input is
viewed as the difference $x - \bm{0}$ from the zero baseline. Under
these assumptions, LRP-$\epsilon$ reduces to gradient $\times$ input,
which is the $K = 1$ Riemann approximation of Integrated Gradients
with $\baseline = \bm{0}$. The reduction fails when biases are
present, when LRP composition (different rules at different layers)
is used, or when the baseline is non-zero.
\end{result}

These reductions matter in practice: they imply that empirical
disagreements between methods should be tracked back to differences in
the axioms (which method ignores which), the value function (which
notion of feature absence is in play), or the path (which trajectory
through input space is integrated), not to differences in mathematical
sophistication.

\subsection{Taxonomy Table}

Table~\ref{tab:taxonomy} situates the methods on three orthogonal
axes: their mathematical object (the primary computation), whether they
are local or global, and whether they are model-specific or
model-agnostic.

\begin{table*}[t]
\centering
\caption{Taxonomy of attribution methods.}
\label{tab:taxonomy}
\small
\begin{tabular}{@{}llllll@{}}
\toprule
Family & Math.\ object & Local/Global & Model-specific & Input type & Output \\
\midrule
Shapley / cooperative game & coalition value function & local (mostly) & no & any & vector $\in \R^d$ \\
Path-based & line integral of gradient & local & yes (diff.) & continuous & vector $\in \R^d$ \\
Gradient / backprop & 1st-order derivative / rules & local & yes (diff.) & continuous & vector / heatmap \\
CAM family & spatial gradient projection & local & yes (CNN) & image & 2D heatmap \\
Perturbation / occlusion & black-box query & local & no & any & vector / heatmap \\
Surrogate (LIME) & local linear regression & local & no & any & vector \\
Rule-based (Anchors) & rule predicate & local & no & any & rule set \\
SAGE & global Shapley over loss & global & no & any & vector $\in \R^d$ \\
\bottomrule
\end{tabular}
\end{table*}

\subsection{The Central Disagreement, Geometrically}

A useful way to summarize the comparison is geometric. Each
attribution method is a function
\begin{equation}
\Phi \colon \mathcal{F} \times \mathcal{X} \to \R^d
\end{equation}
mapping a (model, input) pair to an attribution vector, parameterized
by a choice of value function $\valuefn$, baseline distribution
$\mathcal{D}_{\baseline}$, path $\gamma$, and perturbation
distribution $p_{\mask}$. Two methods that disagree on $\attribution(x)$
disagree because they live at different points in
\[
\mathcal{C} = \{\text{value fn}\} \times \{\text{baselines}\} \times \{\text{paths}\} \times \{\text{perturbations}\}.
\]
The empirical question ``which method should I use?'' is therefore not
a question about $\Phi$ but about which neighbourhood of $\mathcal{C}$
matches the user's intended question. We close this section, and the
methods half of the survey, with the following statement.

\begin{principle}
\label{prin:disagreement}
A disagreement between attribution methods $\Phi_1$ and $\Phi_2$ on a
prediction $\model(x)$ is informative about the model only to the
extent that $\Phi_1$ and $\Phi_2$ share a value function, baseline,
path, and perturbation distribution. Disagreement under different
choices reflects the choices, not the model.
\end{principle}

Principle~\ref{prin:disagreement} summarizes the comparison: attribution
choices are modelling choices, and explanation comparison should hold
those choices constant.

\section{Evaluation Theory and Metrics}
\label{sec:evaluation}

Evaluating an attribution method is harder than producing one. There
is no ground-truth attribution against which to measure, and the
purposes for which attributions are used, including model debugging, user
trust, regulatory compliance, and scientific discovery, impose different
evaluation criteria. This section surveys the principal evaluation
frameworks that have emerged. We organize them around the property each
metric purports to measure: faithfulness to the model, stability under
input changes, alignment with ground-truth annotations, and the validity
of the metric itself.

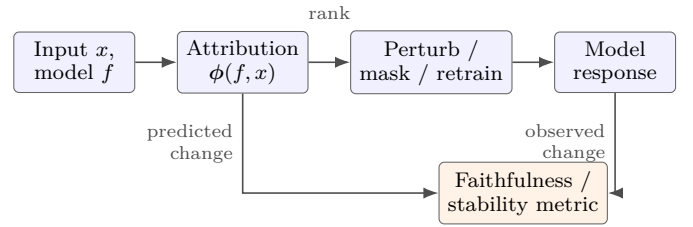
\begin{figure}[t]
\centering
\begin{tikzpicture}[font=\footnotesize,
  box/.style={draw=black!60, rounded corners=2pt, minimum width=16mm,
    minimum height=8mm, align=center, fill=blue!6},
  mbox/.style={box, fill=orange!10},
  arrow/.style={-{Latex[length=2mm]}, semithick, black!70},
  node distance=4mm and 5.5mm
]
\node[box] (input) {Input $x$,\\ model $\model$};
\node[box, right=of input] (attr) {Attribution\\ $\attribution(\model,x)$};
\node[box, right=of attr] (pert) {Perturb /\\ mask / retrain};
\node[box, right=of pert] (resp) {Model\\ response};
\node[mbox, below=9mm of $(pert.south)!0.5!(resp.south)$] (metric)
  {Faithfulness /\\ stability metric};
\draw[arrow] (input) -- (attr);
\draw[arrow] (attr) -- (pert)
  node[midway, above=4.5mm, font=\scriptsize] {rank};
\draw[arrow] (pert) -- (resp);
\draw[arrow] (resp.south) |- (metric.east)
  node[pos=0.25, left, font=\scriptsize, align=right] {observed\\ change};
\draw[arrow] (attr.south) |- (metric.west)
  node[pos=0.25, left, font=\scriptsize, align=right] {predicted\\ change};
\end{tikzpicture}
\caption{Generic evaluation pipeline for attribution methods. The
attribution $\attribution(\model,x)$ ranks or selects features; the
model is queried under perturbations or retrainings indexed by the
ranking; the observed change in the model's response is compared
against the change the attribution predicts, producing a faithfulness
or stability metric. Section~\ref{sec:evaluation} catalogues the
instantiations.}
\label{fig:eval-pipeline}
\end{figure}

\subsection{Faithfulness: Does the Attribution Reflect the Model?}

\emph{Faithfulness} measures whether the attribution reports what the
model actually does, as opposed to what a human would do given
the same input. Faithfulness has been operationalized in several
inequivalent ways.

\begin{description}[style=nextline,leftmargin=1.5em]
\item[Perturbation response.] Remove the top-$k$ features by attribution
magnitude and measure the drop in prediction. Used in
~\cite{samek2017evaluating} and codified by
DeYoung et al.~\cite{deyoung2020eraser} as
``comprehensiveness'' (drop when top features are removed) and
``sufficiency'' (drop when only top features are retained).
\item[Functional faithfulness.] $\sum_i \phi_i \approx \model(x) - \model(\baseline)$,
i.e.\ completeness as an evaluation metric distinct from the axiom. Methods that
satisfy completeness pass trivially; methods that do not are penalized.
\item[Causal faithfulness.] Defined by Jacovi and
Goldberg~\cite{jacovi2020faithfulness}, who distinguish faithfulness
from plausibility and argue that the two are often conflated in
NLP attribution.
\end{description}

\subsection{Infidelity and Sensitivity}

Yeh et al.~\cite{yeh2019infidelity} proposed two metric families that
have become standard.

\begin{definition}[Infidelity]
\label{def:infidelity}
For an attribution $\attribution$ and a perturbation distribution
$\mu_I$,
\begin{equation}
\mathrm{INFD}(\attribution, \model, x) = \E_{I \sim \mu_I}\!\Bigl[\bigl(I^\top \attribution - (\model(x) - \model(x - I))\bigr)^2\Bigr].
\label{eq:infidelity}
\end{equation}
\end{definition}

Infidelity measures the mean-squared error between (i) the
attribution's prediction of how the model output changes under
perturbation $I$ and (ii) the actual change. Methods that satisfy
completeness in expectation under $\mu_I$ minimize~\eqref{eq:infidelity}.
Different choices of $\mu_I$ yield different metrics: a Gaussian
$\mu_I$ tests robustness to noise; a sparse $\mu_I$ tests feature
removal.

\begin{definition}[Max-Sensitivity]
\label{def:sensitivity}
$\mathrm{SENS}_{\mathrm{max}}(\attribution, \model, x, r) = \sup_{\|y - x\| \leq r} \|\attribution(\model, y) - \attribution(\model, x)\|.$
\end{definition}

Max-sensitivity bounds how much the attribution can change when the
input is perturbed within radius $r$. It is closely related to the
Lipschitz stability axiom (Axiom~\ref{ax:lipschitz}); a Lipschitz
attribution has bounded max-sensitivity for every $r$.

Yeh et al.\ also characterize the optimum: for a fixed perturbation
distribution $\mu_I$, the attribution minimizing
infidelity~\eqref{eq:infidelity} is the generalized least-squares
solution
\begin{equation}
\attribution^{\ast}
= \E_{I \sim \mu_I}\!\bigl[I I^{\top}\bigr]^{-1}\,
\E_{I \sim \mu_I}\!\bigl[I\,\bigl(\model(x) - \model(x - I)\bigr)\bigr],
\label{eq:infidelity-opt}
\end{equation}
a smoothed, kernel-weighted gradient of $\model$ around $x$. They
further show that kernel smoothing of a given attribution lowers its
max-sensitivity and, under the conditions stated in their paper, does
not worsen its infidelity; empirically it often improves both.
Stability and fidelity are therefore not strictly opposed. The real
tension is between fidelity to $\model$ at the point $x$ and fidelity
averaged over the neighbourhood that $\mu_I$ defines.

\subsection{ROAR and KAR: Retraining-Based Faithfulness}

Hooker et al.~\cite{hooker2019roar} argued that perturbation-response
metrics like~\eqref{eq:occlusion} are confounded by distribution shift:
removing features may simply push the input off-manifold, in which case
a prediction drop may reflect manifold violation and fail to measure
feature importance.

\textbf{ROAR (RemOve and Retrain)} addresses this by retraining the
model after feature removal. Given an attribution method, mask the
top-$k\%$ of features, train a fresh model on the masked data, and
compare its test accuracy to a baseline that masks $k\%$ randomly. A
faithful attribution should produce a substantially larger accuracy
drop than random.

\textbf{KAR (Keep And Retrain)} is the dual: retain only the top-$k\%$
features. A faithful attribution should still allow the model to learn.

ROAR/KAR is expensive because each evaluation point requires a full
retraining, and so has been applied at small scale, but it is
one of the more defensible faithfulness criteria in the literature
because it explicitly controls for distribution shift.

\subsection{Sanity Checks}

Adebayo et al.~\cite{adebayo2018sanity} proposed the
\emph{model-randomization} and \emph{data-randomization} sanity checks:
\begin{itemize}[leftmargin=*]
\item \emph{Model-randomization:} replace the model's weights
(layer-by-layer or in cascade) with random values; the attribution
should change substantially. An attribution that does not is not
explaining the model.
\item \emph{Data-randomization:} train the model on data with permuted
labels; the attribution should change. An attribution that does not is
not sensitive to what the model has learned.
\end{itemize}
These checks are \emph{necessary} for an attribution method to be
called an explanation, but they are not sufficient: a method can pass
both and still be uninformative for downstream tasks.

The most-discussed empirical finding from Adebayo et al.\ is the
failure of guided backpropagation and deconvolution under the
model-randomization check. We incorporated this finding directly into
Table~\ref{tab:empirical-evidence}.

\subsection{Insertion / Deletion Curves}

\begin{figure}[t]
\centering
\begin{tikzpicture}[font=\footnotesize, scale=1.0]
\draw[-{Latex[length=2mm]}, black!70] (0,0) -- (5.3,0);
\node[black!70, font=\footnotesize] at (2.65,-0.5) {fraction of features $k/d$};
\draw[-{Latex[length=2mm]}, black!70] (0,0) -- (0,3.4)
  node[left=2pt] {$\model_c$};
\foreach \x/\l in {0/0, 4.8/1} \draw (\x,0.06) -- (\x,-0.06)
  node[below, font=\scriptsize] {$\l$};
\fill[blue!8] (0,0) -- plot[domain=0:4.8, samples=50]
  ({\x}, {3.0*(1-exp(-1.4*\x))/(1-exp(-6.72))}) -- (4.8,0) -- cycle;
\draw[very thick, blue!60!black]
  plot[domain=0:4.8, samples=50]
  ({\x}, {3.0*(1-exp(-1.4*\x))/(1-exp(-6.72))});
\node[blue!60!black, font=\scriptsize, anchor=east]
  at (1.35,2.85) {insertion};
\draw[-{Latex[length=1.5mm]}, blue!60!black, shorten >=1pt]
  (1.4,2.85) -- (1.75,2.55);
\node[blue!50!black, font=\scriptsize] at (3.3,1.15) {AUC-Ins};
\draw[very thick, dashed, red!65!black]
  plot[domain=0:4.8, samples=50]
  ({\x}, {3.0*exp(-1.4*\x) + 0.0});
\node[red!65!black, font=\scriptsize, anchor=west]
  at (1.55,0.62) {deletion};
\draw[densely dotted, thick, black!55]
  plot[domain=0:4.8, samples=40] ({\x}, {3.0*\x/4.8});
\node[black!55, font=\scriptsize, rotate=29, anchor=south]
  at (3.6,2.1) {random order};
\end{tikzpicture}
\caption{Insertion and deletion curves~\cite{petsiuk2018rise}.
Features are inserted into a baseline input (solid) or deleted from
$x$ (dashed) in decreasing order of attribution magnitude. A faithful
attribution makes the prediction rise quickly under insertion and fall
quickly under deletion, relative to a random ordering (dotted); the
shaded area is AUC-Ins. Both curves depend on the baseline used for
insertion and on the replacement value used for deletion, which is the
same baseline-sensitivity caveat as in
Section~\ref{sec:failure}.}
\label{fig:insertion-deletion}
\end{figure}
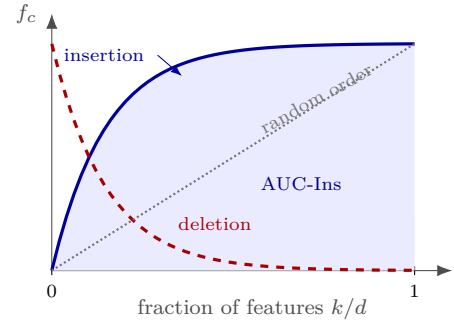

For image attribution, Petsiuk et al.~\cite{petsiuk2018rise}
introduced two curves:
\begin{description}[style=nextline,leftmargin=1.5em]
\item[Insertion.] Start from a baseline (e.g., a blurred image) and
add features in decreasing order of attribution magnitude. Plot the
prediction as a function of the number of features inserted; a faithful
attribution should rise quickly.
\item[Deletion.] Start from $x$ and delete features in decreasing
order of attribution. The prediction should fall quickly.
\end{description}
The area under the resulting curves (AUC-Ins, AUC-Del) summarizes the
method's performance; Figure~\ref{fig:insertion-deletion} sketches the
geometry. The metric has the virtue of being entirely model-internal:
no ground-truth annotation is required.

\subsection{Localization Metrics}

When ground-truth annotations are available, such as bounding boxes for
object detection or rationales for text classification, the
attribution can be compared directly to them:
\begin{description}[style=nextline,leftmargin=1.5em]
\item[Pointing game.] Does the maximum-attribution pixel lie within the
ground-truth bounding box? (Standard in CAM evaluation.)
\item[IoU.] Treat the top-$k$ attributions as a predicted mask and
compute intersection-over-union with the ground-truth mask.
\item[Token F1 / IOU.] In NLP, treat selected tokens as a predicted
rationale and compare against human rationales~\cite{deyoung2020eraser}.
\end{description}
Localization metrics conflate plausibility (alignment with human
annotations) with faithfulness (alignment with model behaviour). A
method can have perfect IoU while being unfaithful (e.g.,~if the model
relies on features outside the ground-truth box but the attribution
defers to the human annotation).

\subsection{Using Attribution During Training}

Another evaluation route is to make explanations actionable during
training. Ross, Hughes, and Doshi-Velez~\cite{ross2017rightreasons}
penalize input gradients at annotated irrelevant dimensions, training
models that are accurate and at the same time constrained away from
known spurious reasons. Rieger et al.~\cite{rieger2020penalizing} extend
this idea through contextual-decomposition explanation penalization,
including feature interactions. These methods do not replace post-hoc
faithfulness tests; they show that an attribution method can be
validated by whether its signal can guide a model away from documented
confounders under a stated training objective.

\subsection{Human-Grounded Evaluation}

Doshi-Velez and Kim~\cite{doshivelez2017rigorous} categorized
evaluations into three tiers: \emph{functionally-grounded}
(model-internal, no human required), \emph{human-grounded} (simplified
tasks with non-expert users), and \emph{application-grounded}
(deployed tasks with domain experts). Most attribution evaluations
have been functionally-grounded; application-grounded evaluations
remain rare and are often the most decision-relevant.

DeYoung et al.~\cite{deyoung2020eraser} introduced the ERASER
benchmark, which provides human rationales for several NLP tasks and
treats attribution evaluation as a token-selection problem. The
benchmark exposes a striking gap: even attribution methods that score
well on functional metrics often fail to identify the tokens that
humans annotate as evidence.

For model debugging in text classification, Bastings et
al.~\cite{bastings2022shortcuts} propose a complementary shortcut-based
protocol: inject known lexical shortcuts, verify that the trained model
uses them, and evaluate whether salience methods rank the shortcut
tokens near the top. This protocol is important because it provides a
controlled ground truth for model reliance, something a human
rationale alone cannot supply.

\subsection{Sanity Checks on the Metrics}

A second-order concern, raised by Tomsett
et al.~\cite{tomsett2020sanity}, is that the evaluation metrics
\emph{themselves} can fail sanity checks. They showed that several
popular faithfulness metrics correlate weakly across methods. A
method ranked highest by one metric is often ranked lowest by another,
and some metrics depend strongly on hyperparameters in ways
the original authors did not document. The methodological implication
is that no single metric should be treated as authoritative; a method
should be evaluated against multiple, ideally diverse, metrics, with
the disagreements among them reported.

\subsection{Summary}

Table~\ref{tab:eval} summarizes the principal evaluation metrics. No
single metric dominates: faithfulness, stability, and plausibility
are different quantities, and a method can score well on one while
failing another. What the
metrics jointly require is stated as items R8--R9 of the reporting
checklist (Section~\ref{sec:reporting}).

\begin{table*}[t]
\centering
\caption{Evaluation metrics for attribution methods. ``Required'' is
what the metric needs beyond the attribution and the model;
``Limitations'' lists the principal known failure modes.}
\label{tab:eval}
\footnotesize
\begin{tabularx}{\textwidth}{@{}p{0.19\textwidth}p{0.23\textwidth}p{0.20\textwidth}X@{}}
\toprule
Metric & What it measures & Required & Principal limitation \\
\midrule
Completeness check & functional faithfulness & none & trivial for methods satisfying it as axiom \\
Comprehensiveness / Sufficiency~\cite{deyoung2020eraser} & top-$k$ perturbation response & none & off-manifold distortion \\
Infidelity~\cite{yeh2019infidelity} & MSE between attribution and $\Delta f$ & noise distribution $\mu_I$ & depends on choice of $\mu_I$ \\
Max-sensitivity~\cite{yeh2019infidelity} & local stability of $\attribution$ & radius $r$ & infidelity/sensitivity trade-off \\
ROAR / KAR~\cite{hooker2019roar} & retraining-based faithfulness & retraining budget & very expensive \\
Sanity (model-rand.)~\cite{adebayo2018sanity} & non-degeneracy w.r.t.\ model & weight randomization & necessary, not sufficient \\
Sanity (data-rand.)~\cite{adebayo2018sanity} & non-degeneracy w.r.t.\ data & label permutation & necessary, not sufficient \\
Insertion / Deletion AUC~\cite{petsiuk2018rise} & ordering quality of attribution & baseline (e.g., blur) & strong baseline dependence \\
Pointing game / IoU & alignment with bounding box & ground-truth box & conflates plausibility, faithfulness \\
ERASER token F1/IoU~\cite{deyoung2020eraser} & alignment with human rationale & human rationale & limited to text tasks \\
Shortcut protocol~\cite{bastings2022shortcuts} & recovery of known model shortcut & controlled shortcut data & task construction must match debugging hypothesis \\
Application study & downstream utility & human users + task & expensive, low statistical power \\
\bottomrule
\end{tabularx}
\end{table*}

\section{Failure Modes and Theoretical Gaps}
\label{sec:failure}

The standard analyses of attribution failure modes by Adebayo
et al.~\cite{adebayo2018sanity}, Kindermans
et al.~\cite{kindermans2019unreliability}, Ghorbani
et al.~\cite{ghorbani2019fragile}, Slack
et al.~\cite{slack2020fooling}, and Kumar et al.~\cite{kumar2020problems}
present these failures as discoveries about specific methods. We
recast them as mathematical phenomena tied to specific choices of
value function, baseline, path, or perturbation distribution. This
recasting clarifies which failures are artefacts of method design and
which are intrinsic to the attribution problem itself.

\subsection{Gradient Saturation and Shattering}

Let $\model$ be differentiable. \emph{Saturation} occurs when
$\partial \model / \partial x_i \approx 0$ in a region around $x$,
even though $\model$ is locally sensitive to $x_i$ in a broader sense.
The canonical example is a ReLU network in which a unit's gradient is
zero on its inactive side; on long inputs the cumulative effect is
that important features receive zero saliency.

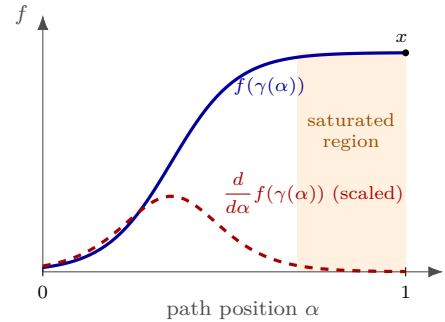
\begin{figure}[t]
\centering
\begin{tikzpicture}[font=\footnotesize]
\fill[orange!12] (3.36,0) rectangle (4.8,2.9);
\node[orange!60!black, font=\scriptsize, align=center]
  at (4.08,1.85) {saturated\\ region};
\draw[-{Latex[length=2mm]}, black!70] (0,0) -- (5.3,0);
\node[black!70, font=\footnotesize] at (2.65,-0.5) {path position $\alpha$};
\draw[-{Latex[length=2mm]}, black!70] (0,0) -- (0,3.4)
  node[left=2pt] {$\model$};
\foreach \x/\l in {0/0, 4.8/1} \draw (\x,0.06) -- (\x,-0.06)
  node[below, font=\scriptsize] {$\l$};
\draw[very thick, blue!60!black]
  plot[domain=0:4.8, samples=70]
  ({\x}, {2.9/(1+exp(-2.3*(\x-1.7)))});
\node[blue!60!black, font=\scriptsize, anchor=west]
  at (2.4,2.45) {$\model(\gamma(\alpha))$};
\draw[very thick, dashed, red!65!black]
  plot[domain=0:4.8, samples=70]
  ({\x}, {2.9*2.3*exp(-2.3*(\x-1.7))/(1+exp(-2.3*(\x-1.7)))^2 / 1.667 * 1.0});
\node[red!65!black, font=\scriptsize, anchor=west]
  at (2.25,1.05) {$\dfrac{d}{d\alpha}\model(\gamma(\alpha))$ (scaled)};
\fill[black] (4.8,2.9*0.999) circle (1.3pt);
\node[font=\scriptsize, anchor=south east] at (4.95,2.92) {$x$};
\end{tikzpicture}
\caption{Gradient saturation along an attribution path. For a
sigmoidal $\model$, the gradient (dashed) is concentrated in the
middle of the path and nearly zero at the input $x$ itself (shaded
band). Saliency evaluated at $x$ misses the feature; the path
integral~\eqref{eq:ig} accumulates the mid-path gradient mass and
recovers the full prediction difference, as in the worked
example~\eqref{eq:saturation-example}.}
\label{fig:saturation}
\end{figure}

For example, take $\model(x) = 1 - e^{-5x}$ at $x = 1$ with baseline
$\baseline = 0$.
The prediction difference is
$\Delta f = 1 - e^{-5} \approx 0.993$, yet the local gradient is
$\model'(1) = 5e^{-5} \approx 0.034$: saliency and gradient
$\times$ input both report the feature as nearly irrelevant. The path
integral repairs this by averaging the gradient over the whole
trajectory and avoiding reliance on the saturated endpoint:
\begin{equation}
\IG_1(1; 0) = (1 - 0)\int_0^1 5 e^{-5\alpha}\, d\alpha
= 1 - e^{-5} = \Delta f.
\label{eq:saturation-example}
\end{equation}
Figure~\ref{fig:saturation} plots the same effect for a sigmoidal
model along its attribution path.

Saturation is a property of the local geometry of $\model$, so any
purely local method, including raw saliency, gradient $\times$ input, and
the $K=1$ Riemann approximation of IG, is vulnerable. The
mathematical fix is to integrate: IG with sufficiently large $K$
reduces saturation effects by averaging over the path and recovering
completeness as the numerical integral converges. Path methods reduce
this failure mode, although the result still depends on the baseline
and path.

\emph{Gradient shattering} is a stronger failure: in deep networks,
gradients become high-frequency and approximately independent across
inputs, with white-noise-like behaviour inconsistent with a smooth
function. The effect grows with depth and is a known property of deep
networks~\cite{balduzzi2017shatteredgradients}. SmoothGrad addresses
shattering by averaging gradients over a noise neighbourhood,
effectively replacing $\model$ with a smoothed version
$\tilde{\model}_\sigma$. The trade-off is between locality (small
$\sigma$) and stability (large $\sigma$).

\subsection{Baseline Sensitivity}

For any method that requires a baseline, including IG, EG, DeepLIFT, LRP, and
Shapley methods with $\valuefn^{\mathrm{int}}$, the attribution depends
on the baseline choice. Sturmfels et al.~\cite{sturmfels2020baselines}
showed how strong this dependence can be: for the same model and
input, IG with a zero baseline and IG with a mean baseline can yield
attributions that disagree on the sign for most features.

This is not strictly speaking a failure because a baseline is a modelling
choice and different baselines answer different questions. However, the
empirical literature often treats it as one. We propose
the following classification:

\begin{remark}
A baseline-dependence ``failure'' is meaningful only if (i) the
researcher does not state which baseline they used, or (ii) the
researcher claims their attribution is canonical when it depends on a
hyperparameter. Otherwise, baseline dependence is the
correct behaviour.
\end{remark}

\subsection{Correlated and Dependent Features}

Real features are correlated, and most attribution methods are
defined assuming independence at one or more points. The
interventional value function~\eqref{eq:value-interventional} produces
inputs with implausibly extreme features when applied to highly
correlated coordinates; the conditional value function~\eqref{eq:value-conditional}
requires estimating $d$-dimensional conditional distributions, which is
intractable except in special cases.

The mathematical statement of the problem, paraphrasing Theorem~3
of Janzing et al.~\cite{janzing2020causal}, is:
\begin{result}[Paraphrase of Janzing et al.~\cite{janzing2020causal}, Thm.~3]
For some $\model$ and some pairs of features $X_i, X_j$ that are
perfectly correlated under $p_X$, the Shapley value
$\phi_i^{\mathrm{Shap}}$ under $\valuefn^{\mathrm{int}}$ and under
$\valuefn^{\mathrm{cond}}$ can differ by an arbitrary factor. The
precise sufficient conditions are in the cited paper.
\end{result} The implication is sharp: a researcher
who reports ``the Shapley value of feature $i$'' without specifying
the value function has reported an underdetermined quantity.

Approximate solutions: Aas et al.~\cite{aas2021conditional} estimate
$\valuefn^{\mathrm{cond}}$ via a multivariate Gaussian or copula model;
Frye et al.~\cite{frye2020manifold} use a learned generative model to
sample on-manifold coalition completions; Janzing
et al.~\cite{janzing2020causal} argue that the question itself should
be reformulated as a causal one.

\subsection{Off-Manifold Perturbations}

Many evaluations perturb the input by setting features to zero, blurring
them, or replacing them with samples from a marginal distribution. The
resulting inputs typically lie far from the data manifold, and the
model's prediction at these inputs is undefined in a deeper sense:
the model was never trained on such inputs and its behaviour there is
extrapolation, not interpretation.

Mathematically, off-manifold perturbation breaks the implicit
assumption of all perturbation-based methods that
$\model$ behaves locally well outside $\mathrm{supp}(p_X)$.
Hooker et al.~\cite{hooker2019roar} showed that the apparent ``faithfulness''
of an attribution measured by perturbation response can be entirely
explained by the model's degradation on off-manifold inputs, not by the
attribution's identification of important features.

The principled fix is to constrain perturbations to the data manifold:
generative-model perturbations
(FIDO-CA~\cite{chang2019counterfactual}, Schulz
et al.~\cite{schulz2020information}) and on-manifold Shapley
(Frye et al.~\cite{frye2020manifold}) both implement this. The cost is
an additional model whose own validity must be verified.

\subsection{Causal Misinterpretation}

A pervasive failure lies in interpretation: reading attribution as a
causal effect. The
following statements have all appeared in published applied papers:
\begin{itemize}[leftmargin=*]
\item ``Feature $i$ caused the prediction $\model(x) > 0$.''
\item ``Increasing feature $i$ by 1 unit would change the prediction
by $\phi_i$ units.''
\item ``Feature $i$ is more important than feature $j$ for the
underlying phenomenon.''
\end{itemize}
None of these is supported by any standard attribution method without
strong additional assumptions about the data-generating process.
Attribution methods compute \emph{associations} (in the loose sense of
sensitivity, marginal contribution, or conditional expectation),
typically with respect to a model trained by maximum likelihood. They do
not compute causal effects on the world.

Janzing et al.~\cite{janzing2020causal} give a constructive bridge: a
class of attribution methods that admit a causal interpretation under
the assumption that the model is a fair approximation of the conditional
expectation and that the user has correctly specified the causal graph.
The conditions are strong; the broader point is that
attribution-as-causation requires a separate, explicit causal model.

\subsection{Adversarial Manipulation of Explanations}

Adversarial attacks on explanations show that the attribution map
$\attribution(\model, x)$ can be made nearly arbitrary by a small,
nearly imperceptible perturbation $\delta$ that preserves the
prediction $\model(x + \delta) \approx \model(x)$.

\begin{result}[Adversarial manipulability of explanations; Ghorbani et al.~\cite{ghorbani2019fragile} and Dombrowski et al.~\cite{dombrowski2019manipulated}]
\label{thm:adversarial}
For standard gradient-based attributions on ReLU networks, including
saliency, gradient $\times$ input, and Grad-CAM, the cited papers
construct, for typical inputs $x$, a perturbation $\delta$ of small
norm $\|\delta\| \leq \epsilon$ such that the prediction is nearly
preserved, $|\model(x + \delta) - \model(x)|$ is at most
$O(\epsilon)$, while the attribution map changes by a substantial
amount $\|\attribution(x + \delta) - \attribution(x)\| \geq C$, where
$C$ can be made arbitrarily large. The constructions exploit the
piecewise-linear geometry of ReLU networks; we state this as a summary
grounded in the cited constructions and do not supply a self-contained proof.
\end{result}

\noindent The geometric intuition due to Dombrowski
et al.~\cite{dombrowski2019manipulated} is that for ReLU networks, the
prediction is piecewise linear but the attribution is piecewise
\emph{constant}; small movements across pieces change the attribution
discontinuously while leaving the prediction nearly unchanged. The
mathematical fix is to enforce Lipschitz stability
(Axiom~\ref{ax:lipschitz}), which SmoothGrad approximates and
expected-gradient methods enforce in expectation.

Slack et al.~\cite{slack2020fooling} extended adversarial attacks to
LIME and SHAP: a model can be trained that behaves discriminatorily on
the data distribution but produces non-discriminatory explanations
under LIME or KernelSHAP. The attack exploits the off-manifold queries
that LIME and KernelSHAP make: a model that is anomalous on those
queries can hide its real behaviour. The fix is the same as for the
off-manifold problem: constrain queries to the data manifold.

\subsection{Disagreement Among Methods}

A practical concern, often the first one users encounter, is that
different attribution methods disagree on the same prediction.
Krishna et al.~\cite{krishna2024disagreement} formalized this as the
\emph{disagreement problem}, measured across six methods and four
benchmarks; their results show median rank correlations below $0.5$
between methods that are widely treated as interchangeable. Bilodeau
et al.~\cite{bilodeau2024impossibility} provided the complementary
theoretical statement: no attribution method can simultaneously
satisfy a small list of intuitive properties (linearity,
completeness, and a faithfulness-style criterion), so disagreement is
not an empirical accident of any particular method.

Section~\ref{sec:comparison}
(Principle~\ref{prin:disagreement}) established that disagreement is informative
only when the methods share a value function, baseline, path, and
perturbation distribution. The most common form of reported
disagreement, for example between vanilla saliency and KernelSHAP,
fails this condition: the methods compute different
quantities, so it would be strange if they agreed.

The implication for practice is that disagreement calls for further
investigation before it supports any conclusion. ``IG and SHAP
disagree'' is not the same finding as ``IG with the same baseline,
value function, and number of samples produces a different ordering on
different runs''; only the latter indicates a method-level
instability. We propose a short protocol for investigating
disagreement: (i) verify that both methods use the same scalar output
$\model_c(x)$ and the same feature granularity; (ii) compare value
functions (interventional vs.\ conditional) and baselines; (iii)
quantify approximation error in each method against a higher-budget
reference; (iv) only if (i)--(iii) match, report the methods as
substantively disagreeing.

\subsection{Specification-to-Failure Map}

Table~\ref{tab:failure-map} summarizes the methodological lesson of
the preceding subsections: many failures are traceable to an unstated
choice in the attribution specification. The table is a diagnostic map
for deciding what a paper must report before an attribution claim can
be interpreted.

\begin{table}[t]
\centering
\caption{Hidden choices and the failure modes they commonly induce.}
\label{tab:failure-map}
\footnotesize
\begin{tabularx}{\columnwidth}{@{}p{0.28\columnwidth}p{0.34\columnwidth}X@{}}
\toprule
Hidden choice & Failure mode & Example \\
\midrule
Off-manifold perturbation & Implausible explanations & occlusion, SHAP, LIME \\
Baseline choice & Baseline sensitivity & IG, DeepLIFT \\
Gradient saturation & Missing important features & saliency, low-$K$ IG \\
Rule-modified backpropagation & Sanity-check failure & guided backprop \\
Sampling distribution & Instability / variance & LIME, KernelSHAP, RISE \\
\bottomrule
\end{tabularx}
\end{table}

\subsection{Summary: Failure Modes Are Specifications of Assumptions}

Table~\ref{tab:failure} summarizes the failure modes. The unifying
view is that each failure occurs when an attribution method is asked a
question it was not designed to answer. Saturation occurs when a method
is asked about a region in which the gradient is locally
uninformative. Baseline sensitivity occurs when a method is asked for
a result without a baseline being specified. Off-manifold failure
occurs when a method is asked to extrapolate. Causal misinterpretation
occurs when a method is asked for a causal answer it does not provide.

Each failure has a structural fix: path integration for saturation,
explicit baselines for sensitivity, manifold-aware perturbations for
off-manifold, and explicit causal modelling for causal claims. None of
those fixes is method-agnostic. The implication is the
following: an attribution paper claiming general superiority over
prior methods should identify which failure mode it addresses and what
assumptions or trade-offs the mitigation introduces.

\begin{table*}[t]
\centering
\caption{Failure modes of attribution methods, the methods principally
affected, the formal cause, and the structural mitigation.}
\label{tab:failure}
\small
\begin{tabular}{@{}p{0.18\linewidth}p{0.20\linewidth}p{0.27\linewidth}p{0.27\linewidth}@{}}
\toprule
Failure mode & Methods affected & Formal cause & Mitigation \\
\midrule
Gradient saturation & saliency, $\nabla \model \times x$, low-$K$ IG & $\nabla \model \approx 0$ on the path & $K \to \infty$, DeepLIFT-Rescale \\
Gradient shattering & many gradient methods on deep nets & high-frequency $\nabla \model$ & SmoothGrad ($\sigma > 0$) \\
Baseline sensitivity & IG, EG, DeepLIFT, LRP, $\valuefn^{\mathrm{int}}$ Shapley & answer depends on $\baseline$ & report $\baseline$; use expected baseline \\
Correlated features & $\valuefn^{\mathrm{int}}$ Shapley, KernelSHAP & off-manifold coalition completions & conditional Shapley / manifold-aware \\
Off-manifold perturbations & LIME, KernelSHAP, RISE, occlusion & $\model$ undefined outside $\mathrm{supp}(p_X)$ & generative perturbations \\
Causal misinterpretation & associational methods & associational $\neq$ causal & causal Shapley + DAG \\
Adversarial manipulation & non-Lipschitz methods & Result~\ref{thm:adversarial} & Lipschitz / smoothed attribution \\
Method disagreement & cross-family comparisons & differing value fn, path, $\mathcal{D}_{\baseline}$ & hold method-parameters constant \\
\bottomrule
\end{tabular}
\end{table*}

\section{Applications and Model Families}
\label{sec:applications}

The methods of Sections~\ref{sec:shapley}--\ref{sec:perturbation} differ
in their suitability across model families and input modalities. This
section is brief by design: our focus is the mathematical analysis,
not the application landscape. We summarize, for each major model
family, which method choices are mathematically appropriate, which are
empirically established, and which open problems remain.

\subsection{Tabular Models}

Tabular models, including gradient-boosted trees~\cite{friedman2001gbm},
random forests~\cite{breiman2001randomforests}, generalized linear
models, and deep tabular networks, are the original domain of SHAP
and remain its strongest fit. TreeSHAP~\cite{lundberg2020treeshap}
gives \emph{exact} Shapley values in polynomial time for tree
ensembles, removing the principal obstacle (exponential cost) that
applies to other Shapley implementations.

For tabular data the choice between interventional and conditional
value functions is unusually consequential: real tabular features are
often highly correlated (e.g., age, tenure, and account balance in
credit scoring), and the two value functions can produce contradictory
explanations. The current recommendation, following
\cite{janzing2020causal,sundararajan2020manyshap}, is to report
interventional TreeSHAP as the default and to discuss conditional
TreeSHAP when correlation between features is structural and meaningful
(e.g., when two columns are functionally related). LIME and KernelSHAP
remain competitive for non-tree tabular models.

\subsection{Convolutional Image Models}

For CNNs, the attribution literature has converged on a small set of
methods: Grad-CAM and HiResCAM for coarse spatial attribution;
Integrated Gradients and Expected Gradients for pixel-level
attribution; LRP and DeepLIFT for backpropagation-style attribution;
RISE and Score-CAM for gradient-free attribution; and XRAI for
region-aggregated attribution.

The key methodological points specific to images are:

\begin{itemize}[leftmargin=*]
\item The zero baseline is almost always wrong for image inputs; a
blurred or mean-image baseline is more
appropriate~\cite{sturmfels2020baselines}.
\item Pixel-level attribution is often dominated by high-frequency
artefacts; SmoothGrad smoothing or XRAI region aggregation are usually
required for human-readable maps.
\item Adebayo et al.~\cite{adebayo2018sanity} sanity-checks should be
run for any image attribution method; failure of the sanity check
implies the heatmap does not depend on what the model has learned.
\end{itemize}

\subsection{Transformers and NLP}

For transformer language models, attention rollout~\cite{abnar2020flow}
and transformer relevance propagation~\cite{chefer2021transformer} are
the principal model-aware attribution methods. For NLP classification
specifically, the ERASER benchmark~\cite{deyoung2020eraser} provides
the standard evaluation; LIME, IG (with a token-embedding baseline),
DeepLIFT, and LRP all have NLP-specific
implementations~\cite{arras2017rnnlrp}.

The chief NLP-specific difficulty is that token-level attribution
ignores compositional structure: a sentence's meaning depends on the
interaction of tokens, not their individual contributions. Hierarchical
methods~\cite{singh2018hierarchical}, contextual
decomposition~\cite{murdoch2018contextual}, and feature-interaction
attribution~\cite{tsang2020interaction} address this directly. The
debate between attention-as-explanation~\cite{jain2019attentionnot,
serrano2019attention,wiegreffe2019notnot} has not been settled but has
clarified that attention is at best a partial input-feature
attribution: it captures which tokens the model attends to but not why
those attentions matter for the prediction.
Shortcut-based evaluation~\cite{bastings2022shortcuts} is especially
useful in NLP because many relevant confounders are lexical or
template-like, and therefore can be injected, verified, and evaluated
under controlled data modifications.

For \emph{large} language models, attribution is largely an open
problem (Section~\ref{sec:open}): the same input may produce different
generations under different sampling, gradient computations are
expensive at trillion-parameter scale, and the meaningful unit of
explanation may be sub-token, token, span, or document. Token-level
attribution for large language models inherits the same issues surveyed
here but adds complications from discrete inputs, subword tokenization,
prompt dependence, retrieval context, and generation-time decoding. We
treat these as extensions within the local attribution problem and
outside the survey's separate targets.

\subsection{Graph Neural Networks}

For GNNs, attribution must respect graph structure: features include
node attributes, edge attributes, and the graph topology itself.
Extensions of LIME (GraphLIME), SHAP (GraphSHAP), and Integrated
Gradients (GNN-IG) to graphs have been developed in dedicated lines of
work, which we do not survey here. The mathematical points relevant
to our axiomatic framing are: (i) the value function must condition
on both the graph adjacency structure and the node features; (ii)
permutation symmetries induced by graph isomorphism interact with the
symmetry axiom; (iii) message-passing computations admit a layer-wise
relevance propagation extension analogous to LRP for feed-forward
networks.

\subsection{Biomedical and High-Stakes Domains}

In biomedical applications, the failure modes of
Section~\ref{sec:failure} are not abstract concerns. A clinical
decision-support system that attributes credit to a feature for
spurious reasons (gradient saturation, off-manifold perturbation,
adversarial weight choice during training) can produce confident,
visually convincing explanations that are systematically misleading.

The methodological consensus emerging in this domain prefers:

\begin{itemize}[leftmargin=*]
\item Methods that satisfy completeness and sensitivity-(a) as axioms
(IG, SHAP variants, LRP, DeepLIFT) over methods that satisfy them only
approximately;
\item Methods with documented sanity-check behaviour;
\item Multiple methods reported in parallel, with disagreement
analyzed under Principle~\ref{prin:disagreement} of
Section~\ref{sec:comparison};
\item Training-time explanation constraints when domain knowledge
identifies invalid reasons for a prediction~\cite{ross2017rightreasons,
rieger2020penalizing};
\item Application-grounded evaluation~\cite{doshivelez2017rigorous}
involving domain experts in addition to functional metrics.
\end{itemize}

These are the same principles that emerge in any high-stakes
decision domain: legal, financial, hiring. The mathematical machinery
of axioms, value functions, and failure modes is the common ground;
domain expertise determines which axioms are required for which
decisions.

\subsection{Application-Domain Matrix}

Table~\ref{tab:applications} summarizes which methods are preferred
across application domains, the principal cautions, and standard
evaluation conventions.

\begin{table*}[t]
\centering
\caption{Application-domain matrix for attribution methods.}
\label{tab:applications}
\footnotesize
\begin{tabularx}{\textwidth}{@{}p{0.09\textwidth}p{0.28\textwidth}p{0.25\textwidth}X@{}}
\toprule
Domain & Preferred methods & Principal cautions & Standard evaluation \\
\midrule
Tabular & TreeSHAP, KernelSHAP, LIME & feature correlation; value-function choice & ROAR, insertion/deletion \\
Image (CNN) & IG, EG, Grad-CAM, HiResCAM, XRAI, RISE & baseline choice; sanity checks; off-manifold & Insertion/Deletion AUC, sanity checks, IoU \\
Text (transformer) & IG (token), DeepLIFT, transformer LRP, attention rollout & attention vs.\ attribution; compositional structure & ERASER token F1/IoU, comprehensiveness/sufficiency \\
Graph (GNN) & GNN-IG, GraphSHAP, GNN-LRP & graph structure as feature; permutation symmetry & ablation by edge / node subset \\
Biomedical & TreeSHAP / IG with documented baselines & off-manifold; adversarial trojans; multi-method reporting & ROAR + application-grounded \\
Sequence & Contextual decomposition, hierarchical, LRP-RNN & long-range dependence; vanishing gradients & ERASER-style + perturbation \\
LLM (open problem) & N/A & cost; sampling randomness; meaningful unit & (see Section~\ref{sec:open}) \\
\bottomrule
\end{tabularx}
\end{table*}

\section{A Proposed Reporting Checklist for Attribution Studies}
\label{sec:reporting}

Attribution methods encode modelling assumptions, and comparison or use
of an attribution requires those assumptions to be stated. This section
converts that observation into a proposed reporting checklist: a paper
or applied study that uses an
attribution method should report enough information that another researcher can
(i) reproduce the attribution from the same inputs and model,
(ii) place the result correctly on the axiom matrix
(Table~\ref{tab:axiom-matrix}) and the failure-mode table
(Table~\ref{tab:failure}), and (iii) decide whether the reported
attribution is appropriate for the user's question.

The checklist has ten items, organized into three blocks:
\emph{specification of the question}, \emph{specification of the
method}, and \emph{specification of the evidence}.

\begin{table*}[t]
\centering
\caption{Ten-item reporting checklist for local additive attribution.
The items specify the minimum information needed to reproduce and
interpret an attribution claim.}
\label{tab:reporting-checklist}
\footnotesize
\begin{tabularx}{\textwidth}{@{}p{0.08\textwidth}p{0.38\textwidth}X@{}}
\toprule
Item & What to report & Why it matters \\
\midrule
R1 & Model, training context, and scalar output & fixes the target of attribution \\
R2 & Feature granularity, grouping, and display transform & prevents comparing pixels, tokens, words, and groups as if identical \\
R3 & Baseline or reference distribution & defines the comparison point \\
R4 & Value function or perturbation distribution & defines feature absence and locality \\
R5 & Path, coalition strategy, or sampling design & determines approximation and axioms \\
R6 & Axioms satisfied and not satisfied & states the mathematical guarantees \\
R7 & Approximation budget, stochasticity, seeds, and uncertainty & makes sampled attributions reproducible \\
R8 & Sanity-check behaviour & tests model and data dependence \\
R9 & Faithfulness or stability metric & evaluates the attribution claim \\
R10 & Known failure modes for the setting & prevents overinterpretation \\
\bottomrule
\end{tabularx}
\end{table*}

\subsection{Block A: Specifying the Question}

\textbf{R1. Model and scalar output.} State (i) the model $\model$
being explained, including its architecture, training data, and whether it is
treated as a black box or as a differentiable function;
(ii) the scalar output $\model_c(x)$ being attributed, including the class
logit, post-softmax probability, regression output, loss, or other
quantity; and (iii) the input $x$ and whether the attribution is
local (per-input) or aggregated.

\textbf{R2. Feature definition, grouping, and display transform.}
State what counts as a feature: raw pixels, super-pixels, tokens,
sub-tokens, structured groups, one-hot categories, or graph
nodes/edges. Feature granularity is part of the attribution problem
definition: attributions over pixels, superpixels, tokens, words, and
domain variables are not directly comparable unless the grouping map is
stated. Also state whether attributions are signed, absolute-valued,
normalized, clipped, smoothed, thresholded, aggregated, or recolored
before visualization.

\textbf{R3. Reference / baseline distribution.} State the baseline
$\baseline$ or reference distribution $\mathcal{D}_{\baseline}$ used.
If a single $\baseline$ is used, state its choice (zero, mean, blurred
image, paraphrase, etc.). If an expected baseline is used, state the
distribution $p_X$ from which references are drawn and the number of
samples.

\subsection{Block B: Specifying the Method}

\textbf{R4. Value function or perturbation distribution.} State
explicitly which value function is used: interventional
(\eqref{eq:value-interventional}), conditional
(\eqref{eq:value-conditional}), single-reference
(\eqref{eq:value-ref}), or other. For perturbation methods, state
the perturbation distribution $p_{\mask}$. This item fixes the meaning
of feature absence and locality.

\textbf{R5. Path or coalition strategy.} For path methods, state the
path $\gamma$ (straight-line, adaptive, scale-space) and the number of
Riemann steps $K$. For coalition methods, state the sampling strategy
(KernelSHAP weighting, permutation sampling, TreeSHAP recursion) and
the sample budget $M$.

\textbf{R6. Axioms satisfied / not satisfied.} State which axioms
from Section~\ref{sec:axioms} the chosen method satisfies, under the
chosen value function and baseline. Where a known reduction or
equivalence applies (e.g., Result~\ref{prop:kernel-shap},
Result~\ref{prop:deeplift-ig}), cite it.

\textbf{R7. Computational approximation and stochasticity.} State the
approximation error budget and the corresponding sample size,
integration steps, or network-pass count. For sampling-based methods,
report the number of samples, random seeds, variance estimates, and
convergence diagnostics where feasible.

\subsection{Block C: Specifying the Evidence}

\textbf{R8. Sanity-check results.} Report the outcome of at least
one of the model-randomization and data-randomization sanity checks
of Adebayo et al.~\cite{adebayo2018sanity}. A method that is largely
insensitive to model parameters should not, by itself, be used to
support strong claims about model-specific reasoning.

\textbf{R9. Faithfulness or stability metric.} Report at least one
metric of faithfulness (infidelity~\eqref{eq:infidelity},
comprehensiveness/sufficiency~\cite{deyoung2020eraser}, ROAR/KAR
accuracy drop~\cite{hooker2019roar}, or insertion/deletion AUC
\cite{petsiuk2018rise}) and at least one metric of stability
(max-sensitivity, empirical Lipschitz, or VarGrad-style variance).
We recommend the Quantus toolkit~\cite{hedstrom2023quantus} as a
reference implementation supporting reproducible computation of these
metrics.

\textbf{R10. Known failure modes that apply.} Map the chosen method
onto Table~\ref{tab:failure} and state which failure modes are
plausibly active for the chosen setting. For example, an
interventional Shapley method on a tabular dataset with correlated
features should explicitly acknowledge the correlated-features
failure mode and the value-function ambiguity.

\subsection{Compliance Examples}

We illustrate the checklist with two short compliance examples; the
purpose is to show how short the report can be when the
methodological choices are unambiguous.

\paragraph{Example 1: TreeSHAP on a credit-risk model.}
\emph{R1}: gradient-boosted tree ensemble, target = probability of
default for input record $x$; local attribution.
\emph{R2}: features = the 14 tabular columns of the credit dataset.
\emph{R3}: interventional baseline distribution = empirical training
distribution, 1000 reference samples.
\emph{R4--R5}: interventional value function with TreeSHAP exact
recursion.
\emph{R6}: satisfies efficiency, symmetry, dummy, additivity,
local accuracy, missingness, consistency (see Table~\ref{tab:axiom-matrix}).
\emph{R7}: exact computation; no sampling error.
\emph{R8}: model-randomization sanity check passes (cosine similarity
$< 0.1$ to attribution under random weights).
\emph{R9}: infidelity = $0.012$ at Gaussian noise $\sigma = 0.1$.
\emph{R10}: features ``income'' and ``debt-to-income ratio'' are
correlated ($\rho = 0.71$); interventional Shapley
under-attributes their joint effect (Table~\ref{tab:failure}, row
``correlated features'').

\paragraph{Example 2: Integrated Gradients on an image classifier.}
\emph{R1}: ResNet-50 ImageNet classifier, target = pre-softmax logit
for the predicted class.
\emph{R2}: pixel-level attribution.
\emph{R3}: blurred-input baseline (Gaussian blur, $\sigma = 30$ pixels).
\emph{R4--R5}: single-reference value function; straight-line path
with $K = 200$ Riemann steps.
\emph{R6}: satisfies sensitivity-(a), sensitivity-(b), implementation
invariance, completeness, linearity, symmetry (Result~\ref{thm:ig-axioms}).
\emph{R7}: integration error empirically below $1\%$ on test images
relative to $K = 2000$ reference.
\emph{R8}: sanity-check pass under cascading randomization.
\emph{R9}: max-sensitivity = $0.18$ at $r = 0.02$; insertion AUC =
$0.74$.
\emph{R10}: gradient saturation possible on inactive ReLU paths
(Table~\ref{tab:failure}); off-manifold artefacts at intermediate
$\alpha$ on the blur path.

\subsection{Summary}

The proposed compact, checkable checklist converts the assumptions
encoded by attribution methods into a reproducible reporting practice.
The checklist does not restrict
which attribution methods may be used. It specifies the information
needed for attribution results to support the conclusions drawn from
them. It is offered to authors and reviewers as a proposal, not as a
community-ratified standard.

\section{Scope and Boundary Conditions}
\label{sec:limitations}

The principal scope boundaries and their rationale keep the
mathematical object of study fixed: local additive attribution
operators for predictive models.

\subsection{Explicit Out-of-Scope}

The following topics are adjacent to the survey but outside its main
scope because they manipulate mathematical objects distinct from the
local additive attribution operator.

\begin{itemize}[leftmargin=*]
\item \textbf{Mechanistic interpretability.} Circuit-level
explanations of neural network internals (induction heads, sparse
autoencoders, polysemantic neurons) target a different question:
``what does the model compute?'' This differs from the question ``what
features mattered for this prediction?'' These methods do not produce a per-feature
attribution vector and are not comparable to the methods we survey
within the frame of Sections~\ref{sec:formulation}--\ref{sec:axioms}.
\item \textbf{Concept-based and concept-bottleneck explanations.}
TCAV~\cite{kim2018tcav}, network
dissection~\cite{bau2017networkdissection}, and concept-bottleneck
models replace input features with named human concepts. They are
treated briefly in Section~\ref{sec:introduction} as the natural
contrast to feature attribution; a full survey of the concept
literature is a separate project.
\item \textbf{Counterfactual and example-based explanations.} The
literature on counterfactual explanations (``what would have to
change about $x$ for the prediction to flip?'') and on
example-based explanations (influence functions, prototypes,
training-data attribution) addresses questions complementary to but
distinct from local additive attribution. We mention FIDO-CA and
counterfactual generators
(Section~\ref{sec:perturbation}) where they intersect the
attribution literature; we do not survey the broader counterfactual
research programme.
\item \textbf{Global interpretability and rule extraction.} Methods
that produce a global summary of a black-box model (decision lists,
partial dependence plots, ALE plots, rule
extraction~\cite{lakkaraju2016decisionsets}) target a different
deliverable than per-prediction attribution. SAGE~\cite{covert2020sage}
is the closest member of the Shapley family to a global method and is
covered for that reason; the broader global-interpretation literature
is surveyed in~\cite{saleem2022global,linardatos2021review}.
\item \textbf{Causal explanation methods.} Methods grounded in
structural causal models or Pearl's do-calculus
(causal Shapley~\cite{janzing2020causal}, do-attribution) are
discussed in their connection to interventional value functions
(Section~\ref{sec:shapley}) and as an open problem
(Section~\ref{sec:open}), but a full survey of causal attribution
would require a separate exposition of causal-graph machinery beyond
the scope of this paper.
\item \textbf{Comprehensive LLM-specific attribution.} The LLM
attribution literature has grown rapidly in 2023--2025; we discuss
its open status in Section~\ref{sec:open}. A full treatment would
need to cover token-level vs.\ span-level attribution for
autoregressive generation, in-context-learning attribution, sparse
attention probes, and emerging mechanistic-interpretability tools for
transformers.
\end{itemize}

\subsection{Topics Treated Briefly}

The following topics receive a section or subsection in this paper and
have dedicated literatures of their own.

\begin{itemize}[leftmargin=*]
\item \emph{Graph neural network attribution} (Section~\ref{sec:applications}):
permutation symmetry, edge-attribution, and graph-conditional value
functions deserve a separate treatment.
\item \emph{Biomedical and high-stakes-domain
applications} (Section~\ref{sec:applications}): the
domain-specific evaluation criteria are surveyed lightly; we focus
on the general methodological consensus.
\item \emph{Evaluation in NLP} (Section~\ref{sec:evaluation}): the
ERASER~\cite{deyoung2020eraser} benchmark is the standard reference,
and the broader debate on plausibility vs.\ faithfulness in NLP is
ongoing; we summarize but do not extend the
faithfulness debate.
\end{itemize}

\subsection{Interpretive Boundaries of the Taxonomy}

The taxonomy is a conceptual comparison and does not provide an
empirical ranking. The scoring of prior surveys in
Table~\ref{tab:related-surveys} depends on judgment about what counts
as substantive coverage; for this reason, the scoring rule is stated
explicitly and ambiguous cases are assigned in favour of the prior
survey. Several attribution methods also have implementation-dependent
variants. An axiom entry may hold for the canonical formulation but
not for every software implementation or propagation rule. Claims about
sanity checks, infidelity, robustness, and method disagreement are
therefore synthesized as reported properties of the cited studies, with
implementation-dependent cases marked separately from mathematical
axioms.

\subsection{Threats to Validity of the Taxonomy}

Four limitations qualify the conclusions that should be drawn from
the taxonomy and its tables.

\begin{itemize}[leftmargin=*]
\item \emph{Corpus limitation.} The corpus prioritizes methodological
and axiomatic relevance over exhaustive bibliometric coverage; counts
and coverage claims should be read accordingly.
\item \emph{Axiom-classification uncertainty.} Several methods exist
in multiple variants (LRP rules, DeepLIFT rules, CAM layers), so
matrix entries depend on which variant and implementation is taken as
canonical; the conditional marks and Appendix~\ref{app:axiom-justification}
record these dependencies, but borderline judgments remain.
\item \emph{Domain dependence.} Attribution behaviour differs across
images, text, tabular data, graphs, and biological sequences;
properties observed in one modality do not automatically transfer.
\item \emph{Evaluation non-equivalence.} Faithfulness, stability,
sanity checks, and human usefulness measure different quantities
(Section~\ref{sec:evaluation}); the taxonomy does not assume that any
one of them subsumes the others.
\end{itemize}

\subsection{Methodological Boundaries}

The survey is structured and taxonomy-driven. Its corpus construction
prioritizes papers that define attribution operators, establish
axioms, evaluate attribution behaviour, or document failure modes.

\begin{itemize}[leftmargin=*]
\item The axiom-by-method matrix (Table~\ref{tab:axiom-matrix})
encodes the authors' reading of each method's axiomatic profile;
where a method's behaviour depends on architectural details or on a
choice of rule (LRP rules, DeepLIFT rules), we mark the entry
conditionally and refer to the cited paper.
\item Several reductions reported in Section~\ref{sec:comparison} as
``Results'' are paraphrases of theorems in the cited papers, with
assumptions made explicit. We have tried to flag every assumption
that is critical to the reduction.
\item Empirical results we cite (e.g.,~the sanity-check failure of
guided backpropagation) are reported as such, not as theorems. Where
quantitative claims are made, they reflect the original
authors' experimental setting, not a meta-analysis.
\end{itemize}

\section{Open Problems and Future Directions}
\label{sec:open}

We close with six directions in which the methods of
Sections~\ref{sec:shapley}--\ref{sec:perturbation} are incomplete.
Each is stated as a question that current theory answers partially or
not at all.

\subsection{Causal Attribution}

The clearest open problem is the rigorous integration of attribution
methods with causal modelling. Janzing
et al.~\cite{janzing2020causal} provided the first systematic
formulation, but the conditions under which Shapley-style attribution
admits a causal interpretation remain restrictive. Open questions
include:

\begin{itemize}[leftmargin=*]
\item How does attribution interact with confounding? An attribution
method may attribute credit to feature $i$ that, in the underlying
causal graph, is mediated by an unobserved confounder.
\item Can attribution methods be characterized as estimating
\emph{some} causal estimand, such as average treatment effect, conditional
ATE, or individualized treatment effect, under the model's
implicit causal assumptions?
\item For interventional Shapley, what is the relationship between the
do-operator and the value function? Recent work suggests a tight
correspondence, but a full theorem remains to be stated.
\end{itemize}

\subsection{Correlated Features and Manifold-Aware Attribution}

The conditional vs.\ interventional value-function debate
(Section~\ref{sec:shapley}) remains live. Manifold-aware
methods~\cite{frye2020manifold,aas2021conditional,schulz2020information}
constrain perturbations or coalition completions to the data
manifold, but they require an additional generative model whose own
faithfulness is now a question. Two open problems:

\begin{itemize}[leftmargin=*]
\item How should the additional generative model be trained,
evaluated, and validated? Standard generative-model evaluation (FID,
likelihood) is not obviously appropriate for an attribution sub-routine.
\item Is there an axiomatic characterization of on-manifold Shapley
analogous to Result~\ref{thm:shapley}? The Lundstrom-Razaviyayn-style
characterization~\cite{lundstrom2025ig} of IG does not yet have a
manifold-aware analogue.
\end{itemize}

\subsection{Uncertainty in Explanations}

Attribution methods are deterministic functions of $(\model, x)$. Yet
the model itself is a sample from a learning algorithm, and the
attribution should plausibly carry uncertainty propagated from the
model's training process. Existing approaches include
Bayesian-network attribution, ensemble disagreement, and
posterior-distribution attribution, but the principled framework remains
open. The natural question is whether the attribution should be a
distribution over $\R^d$, and if so what its
marginals should mean.

\subsection{Attribution for Large Language Models}

LLM attribution is the most active open area in XAI today. Four
specific difficulties:

\begin{itemize}[leftmargin=*]
\item \textbf{What is the prediction?} LLMs produce sequences, not
scalars. Attribution targets include per-token logits, sequence
probabilities, and aggregate metrics like answer correctness; the
choice matters.
\item \textbf{What is the feature?} Sub-tokens, tokens, spans,
sentences, documents, system prompts, and retrieved context are all
candidate units. Hierarchical methods~\cite{singh2018hierarchical}
extend cleanly here but require deciding the granularity in advance.
\item \textbf{Computational cost.} A trillion-parameter forward pass
makes IG-style $K$-step path integration prohibitively expensive;
gradient methods are similarly costly at scale.
\item \textbf{In-context learning attribution.} For few-shot LLM
predictions, the explanation may need to attribute credit to specific
demonstrations in the context. This is a new problem class without an
obvious analogue in the methods we have surveyed.
\end{itemize}

\subsection{Rigorous Benchmarks and Human-Aligned Axioms}

The benchmarks of Section~\ref{sec:evaluation} are largely
functionally-grounded and often disagree with one another. Two
directions seem productive:

\begin{itemize}[leftmargin=*]
\item \textbf{Application-grounded benchmarks.} ERASER~\cite{deyoung2020eraser}
is one such; analogues are needed for vision (clinical image cohorts),
biology (gene-perturbation prediction), and structured prediction
(graph classification). The principal obstacle is the cost of expert
annotation.
\item \textbf{Human-aligned axioms.} The axioms surveyed in
Section~\ref{sec:axioms} are mathematically motivated. A
complementary line of work would establish which axioms humans
actually endorse for an explanation, and whether the satisfaction of
those axioms predicts downstream utility.
\end{itemize}

\subsection{Explanation Robustness}

Result~\ref{thm:adversarial} establishes that gradient-based
attributions are intrinsically fragile. The constructive question is
how to build attributions that are stable by design. Two directions:

\begin{itemize}[leftmargin=*]
\item \textbf{Lipschitz-constrained attribution.} Enforce Lipschitz
stability (Axiom~\ref{ax:lipschitz}) directly, either by smoothing the
model or by smoothing the attribution. SmoothGrad does the latter
in expectation; principled methods that enforce a Lipschitz constant
explicitly would be a substantial advance.
\item \textbf{Certifiable attribution.} Analogous to certifiable
robustness for predictions, the question is whether one can certify
that the attribution map has a bounded change under any
$\|\delta\| \leq \epsilon$. Existing approaches based on randomized
smoothing provide a partial answer, but tight bounds remain open.
\end{itemize}

\subsection{Conclusion}

We began this survey by arguing that the mathematical objects
manipulated by feature-attribution methods provide the appropriate
basis for comparison. The intervening sections have
substantiated that argument: the methods discussed here, from exact
Shapley to Grad-CAM to LIME, can be situated in a common frame
defined by a value function, a baseline distribution, a path or
perturbation distribution, and a conservation rule. The axiomatic
analysis (Section~\ref{sec:axioms}, Table~\ref{tab:axiom-matrix}) and
the failure-mode formalization (Section~\ref{sec:failure},
Table~\ref{tab:failure}) make precise the conditions under which each
method's attribution is well-defined, the questions it can and cannot
answer, and the assumptions it requires of the underlying data and
model.

The thesis stated in Section~\ref{sec:introduction}, restated here as
a closing principle:

\begin{quote}\itshape
There is no assumption-free feature-attribution method. Every local
additive attribution method defines feature importance through choices
about value functions, references, paths, perturbation distributions,
or conservation rules. Trustworthy use therefore requires reporting
a heatmap or ranking together with the assumptions under which the
attribution was computed and interpreted.
\end{quote}

For practice, this means that both method papers and applied papers
should state the assumptions behind the reported attribution: a new
attribution method should state its value function, baseline, path,
and conservation rule, and an applied paper that reports attributions
without specifying these choices has reported an underdetermined
quantity.

The resulting catalogue supports a more explicit style of
attribution reporting: method comparisons should name the mathematical
choices they hold fixed, and applied studies should state the choices
on which their attributions depend.

\section*{Reproducibility Statement}

\textbf{Reproducible artefacts.} (i) The
axiom-by-method matrix
(Table~\ref{tab:axiom-matrix}) is derivable from the per-cell
justifications in Appendix~\ref{app:axiom-justification}; each cell
references either the original method paper, an axiomatic
characterization, or a stated conditional assumption. A reader can
verify each cell against the cited paper without further
implementation. (ii) The empirical-evidence companion
(Table~\ref{tab:empirical-evidence}) cites the published sources for
each sanity-check outcome; these external empirical claims were not
generated in this survey.
(iii) The checklist crosswalk in
Appendix~\ref{app:compliance-audit} records the source literature
that motivates each reporting item and shows how the items apply
across method families.

\textbf{Supplementary materials.} The supplementary material includes a
machine-readable reporting checklist, a review-corpus summary, an
axiom-matrix CSV, and a CSV recording the scoring rationale for
Table~\ref{tab:related-surveys}. These materials provide stable
artefacts for checking the taxonomy, the proposed reporting checklist,
and the survey-positioning table.

\section*{Acknowledgments}

The authors thank the maintainers of the open-source attribution
libraries (Captum, SHAP, iNNvestigate, Quantus, Grad-CAM++) whose
implementations underlie much of the empirical XAI literature.

\appendices
\section{Per-Cell Justification of the Axiom Matrix}
\label{app:axiom-justification}

This appendix supplies, for each row of Table~\ref{tab:axiom-matrix},
a short justification or citation for each non-trivial axiom claim.
We use the following abbreviations: \emph{Cmp} (Completeness),
\emph{Sa/Sb} (Sensitivity-(a)/(b)), \emph{II} (Implementation
Invariance), \emph{Cns} (Consistency), \emph{Mono} (Monotonicity),
\emph{Sym} (Symmetry-preservation), \emph{Lin} (Linearity), and
\emph{Cont} (Continuity in $x$). Entries
in the matrix marked \checkmark{} are justified here unconditionally;
entries marked $\circ$ are conditional, with the condition stated.

\subsection*{Shapley Family}

\textbf{Exact Shapley.} Cmp, Sym, dummy and Lin are
Axioms~\ref{ax:efficiency}--\ref{ax:linearity}; satisfied by
construction via~\cite{shapley1953value}. Cns is a direct consequence
of monotone marginal contributions. II follows from
$\valuefn$ depending only on the input--output function of
$\model$. Cont in $x$ holds for any continuous value function $\valuefn$.

\textbf{KernelSHAP~\cite{lundberg2017unified}.} In the limit of infinite
samples, equals exact Shapley (Result~\ref{prop:kernel-shap});
inherits its axioms. Cns and Mono are conditional ($\circ$) because
finite-sample KernelSHAP can exhibit variance-induced violations,
documented in~\cite{covert2021kernelshap,chen2023algorithms}.

\textbf{TreeSHAP~\cite{lundberg2020treeshap}.} Inherits all Shapley
axioms exactly; Cns is provably satisfied
(Theorem 1 of~\cite{lundberg2020treeshap}). Cont is conditional
because TreeSHAP attributions are piecewise constant in $x$.

\textbf{DeepSHAP~\cite{lundberg2017unified}.} Cmp is approximate (one
forward+backward pass; full Shapley requires multiple). Sa and Cns are
conditional ($\circ$) and II is conditional because both depend on the
DeepLIFT propagation rule used (Rescale vs.\ RevealCancel).

\textbf{GradientSHAP~\cite{erion2021expectedgradients}.} Equation
\eqref{eq:gradshap}: the inner expectation over $\alpha$ is an IG
operator, the outer expectation over $\baseline$ is a sampling step.
Inherits Cmp, Sa, Sb, II, Lin, Sym from IG. Cns conditional on
sampling adequacy.

\subsection*{Path Family}

\textbf{Integrated Gradients~\cite{sundararajan2017axiomatic}.} Cmp by
fundamental theorem of calculus (Result~\ref{thm:ig-axioms}); Sa, Sb,
II, Lin, Sym by Theorem 1 of the cited paper. Cont follows from the
integrability of the gradient.

\textbf{Expected Gradients~\cite{erion2021expectedgradients}.} Same as
IG, with the baseline replaced by an expectation. Cmp, Sa, Sb, II,
Lin, Sym all preserved by linearity of the expectation operator.

\textbf{Guided IG~\cite{kapishnikov2021guidedig}.} Cmp preserved
because the integration is still over a path. II is conditional
because the path depends on the gradient magnitude, which can break
strict invariance under network reparameterization. Sym is not
preserved.

\textbf{Blur IG~\cite{xu2020blurig}.} Cmp on the scale-space path;
inherits II, Lin from IG. Sym fails because the path is not
permutation-equivariant.

\textbf{Integrated Hessians~\cite{janizek2021integratedhessians}.} Cmp
over feature pairs; inherits II, Sym, Lin from the IG construction.
Sa, Sb at the interaction level.

\subsection*{Gradient and Backpropagation Family}

\textbf{Saliency~\cite{simonyan2013saliency}.} Sb by definition
($\phi_i = 0$ if $\partial f / \partial x_i \equiv 0$). II by
invariance of partial derivatives. Cmp fails (sum of gradients
$\neq \Delta f$). Cont conditional ($\circ$) due to ReLU
non-smoothness.

\textbf{Gradient $\times$ Input.} Cmp is satisfied only when $\model$
is linear ($\circ$). II by invariance of $\nabla \model$. Sa
conditional on nonzero gradient.

\textbf{SmoothGrad~\cite{smilkov2017smoothgrad}.} Smoothing preserves
II and Sb; gives Cont by construction (averaging Gaussian kernel).
Cmp not satisfied.

\textbf{Guided BP~\cite{springenberg2015guidedbp} / Deconvnet~\cite{zeiler2014visualizing}.}
Do not satisfy II because the rule-modified backward pass is
non-standard. Their model-randomization behaviour is reported
separately in Table~\ref{tab:empirical-evidence}.

\textbf{DeepLIFT~\cite{shrikumar2017deeplift}.} Cmp by summation-to-delta
(Axiom~\ref{ax:deeplift-completeness}). Sa, Sb, Sym, Lin by
construction. II conditional on Rescale rule and architecture.

\textbf{LRP~\cite{bach2015lrp}.} Cmp by conservation
(Axiom~\ref{ax:lrp-conservation}). Sa, Sb conditional on the
propagation rule. II conditional on standard rules and absence of
bias terms.

\textbf{FullGrad~\cite{srinivas2019fullgrad}.} Cmp by explicit bias
accounting. Sa, Sb, II, Cont by the smooth differentiability of the
total gradient plus biases.

\textbf{Grad-CAM~\cite{selvaraju2017gradcam}.} Cmp not satisfied at
the pixel level. II conditional on the architecture having global
average pooling.

\textbf{HiResCAM~\cite{draelos2020hirescam}.} Cmp at the layer level
($\circ$). Sa, Sb by direct gradient application (no pooling). II is
conditional ($\circ$): the map is defined on a chosen internal layer,
and functionally equivalent networks need not share internal
representations.

\textbf{Score-CAM~\cite{wang2020scorecam}.} Does not require
gradients, but the weights are computed from selected activation maps
of a target layer, so II is conditional ($\circ$): strict
implementation invariance holds only for methods that depend on
input--output queries alone. Sa by construction (perturbation-based).

\subsection*{Perturbation Family}

\textbf{Occlusion~\cite{zeiler2014visualizing}.} Sa for any patch with
nonzero marginal contribution. II by black-box queries. Sym by
permutation-equivariance of patch sliding. Cmp fails (single
coalition, not exhaustive).

\textbf{LIME~\cite{ribeiro2016lime}.} Cmp conditional ($\circ$) on
the kernel; satisfied only when the kernel is the Shapley
kernel~\eqref{eq:shapley-kernel}. II by black-box queries. Cont by
smoothness of the local linear fit.

\textbf{Anchors~\cite{ribeiro2018anchors}.} Anchors output rules, not
real-valued attributions; Cmp, Lin, Cont are N/A. Sa and Sb apply
analogously (rule-based).

\textbf{Meaningful Perturbations~\cite{fong2017meaningful}.} Sa via
optimization objective. II by black-box queries. Cont conditional on
the regularization $\lambda$.

\textbf{RISE~\cite{petsiuk2018rise}.} Sa by construction. Sym by mask
distribution being permutation-symmetric. II by black-box queries.

\subsection*{Notes on Conditional Entries}

The $\circ$ marks in Table~\ref{tab:axiom-matrix} signal that the
axiom holds under specific conditions: choice of rule (LRP,
DeepLIFT), architectural restrictions (Grad-CAM family, DeepLIFT),
sample adequacy (KernelSHAP, GradientSHAP), or method variant
(CAM family). Readers extracting machine-readable claims from the
matrix should consult the per-method paragraph above for the precise
qualification.

\subsection*{Interpretation of the Matrix}

The matrix entries above summarize mathematical claims and
implementation-dependent empirical statuses reported in the cited
papers. Where a method's behaviour depends on a configuration
(e.g., LRP-$\epsilon$ vs.\ LRP-composition), the matrix records the
most commonly used configuration as the default and the alternative as
a $\circ$.

\section{Checklist Crosswalk for the Analytic Corpus}
\label{app:compliance-audit}

This appendix connects the proposed reporting checklist of
Section~\ref{sec:reporting} to the source literature used throughout
the survey. The goal is to show how each checklist item follows from
established attribution practice, axiomatic characterization, or
documented failure modes. The crosswalk also provides a compact guide
for applying the checklist across method families.

\subsection*{Item-Level Source Anchors}

Table~\ref{tab:checklist-crosswalk} records, for each item R1--R10,
the specification target, representative source anchors in the
analytic corpus, and the item's role in the checklist. Each anchor is
a paper that either motivated the item or documented the consequence
of leaving it unreported.

\begin{table*}[t]
\centering
\caption{Source anchors for the reporting checklist across the analytic
corpus. Each row identifies representative papers that motivate the
corresponding checklist item.}
\label{tab:checklist-crosswalk}
\footnotesize
\begin{tabularx}{\textwidth}{@{}p{0.08\textwidth}p{0.27\textwidth}p{0.35\textwidth}X@{}}
\toprule
Item & Specification target & Representative source anchors & Role in the standard \\
\midrule
R1 & Model and scalar output &
Saliency~\cite{simonyan2013saliency}, Grad-CAM~\cite{selvaraju2017gradcam},
ERASER~\cite{deyoung2020eraser} &
Fixes whether the attribution explains a logit, probability, loss, or
task-specific score. \\
R2 & Feature unit and display transform &
LIME~\cite{ribeiro2016lime}, TCAV~\cite{kim2018tcav},
hierarchical attribution~\cite{singh2018hierarchical},
transformer attribution~\cite{chefer2021transformer} &
Prevents direct comparison of pixels, superpixels, tokens, spans,
concepts, and grouped variables. \\
R3 & Baseline or reference distribution &
Integrated Gradients~\cite{sundararajan2017axiomatic},
DeepLIFT~\cite{shrikumar2017deeplift},
baseline studies~\cite{sturmfels2020baselines} &
Defines the counterfactual reference point against which feature
contributions are measured. \\
R4 & Value function or perturbation distribution &
Shapley values~\cite{shapley1953value}, SHAP~\cite{lundberg2017unified},
conditional Shapley~\cite{aas2021conditional},
manifold Shapley~\cite{frye2020manifold},
causal attribution~\cite{janzing2020causal},
QII~\cite{datta2016qii} &
Specifies feature absence, locality, and whether correlated features
are treated interventionally or conditionally. \\
R5 & Path, coalition strategy, or sampling design &
IG~\cite{sundararajan2017axiomatic},
Guided IG~\cite{kapishnikov2021guidedig},
KernelSHAP~\cite{lundberg2017unified,covert2021kernelshap},
RISE~\cite{petsiuk2018rise} &
Determines numerical approximation, sampling variance, and the
applicable axiom guarantees. \\
R6 & Axioms satisfied and not satisfied &
Shapley~\cite{shapley1953value},
Banzhaf~\cite{banzhaf1965weighted},
Aumann-Shapley~\cite{aumann1974values},
IG axioms~\cite{sundararajan2017axiomatic},
IG characterization~\cite{lundstrom2025ig},
impossibility results~\cite{bilodeau2024impossibility} &
States which mathematical guarantees the attribution can legitimately
support. \\
R7 & Approximation budget and stochasticity &
SmoothGrad~\cite{smilkov2017smoothgrad},
KernelSHAP analysis~\cite{covert2021kernelshap},
Shapley algorithms~\cite{chen2023algorithms},
RISE~\cite{petsiuk2018rise} &
Makes sampled, noisy, or discretized attributions reproducible. \\
R8 & Sanity-check behaviour &
Sanity checks~\cite{adebayo2018sanity},
sanity-check extensions~\cite{tomsett2020sanity},
HiResCAM~\cite{draelos2020hirescam} &
Tests whether the attribution depends on learned model parameters and
training labels. \\
R9 & Faithfulness or stability metric &
Infidelity~\cite{yeh2019infidelity}, ROAR~\cite{hooker2019roar},
ERASER~\cite{deyoung2020eraser}, RISE metrics~\cite{petsiuk2018rise},
robustness~\cite{alvarezmelis2018robustness},
shortcut tests~\cite{bastings2022shortcuts},
Quantus~\cite{hedstrom2023quantus} &
Connects attribution values to model behaviour under controlled
perturbation or stability tests. \\
R10 & Known failure modes &
Input invariance~\cite{kindermans2019unreliability},
fragility~\cite{ghorbani2019fragile},
manipulation~\cite{dombrowski2019manipulated},
fooling attacks~\cite{slack2020fooling},
disagreement~\cite{krishna2024disagreement},
explanation constraints~\cite{ross2017rightreasons,rieger2020penalizing} &
Links each attribution claim to the failure modes most relevant to its
method family and data setting. \\
\bottomrule
\end{tabularx}
\end{table*}

\subsection*{Cross-Family Reporting Profile}

Table~\ref{tab:family-reporting-profile} groups the same ten items by
method family, separating the items that are usually necessary to make
the attribution mathematically specified from the items that support
its empirical use.

\begin{table*}[t]
\centering
\caption{Checklist emphasis by method family. ``Primary'' items are
usually necessary to make the attribution mathematically specified;
``evidence'' items support empirical use of the attribution.}
\label{tab:family-reporting-profile}
\footnotesize
\begin{tabularx}{\textwidth}{@{}p{0.18\textwidth}p{0.25\textwidth}p{0.27\textwidth}X@{}}
\toprule
Method family & Primary specification items & Evidence items & Representative sources \\
\midrule
Shapley and value-function methods &
R1--R5, especially R4 for conditional/interventional choice &
R6, R7, R9, R10 &
\cite{shapley1953value,strumbelj2010efficient,datta2016qii,
lundberg2017unified,lundberg2020treeshap,covert2021kernelshap,
aas2021conditional,frye2020manifold,janzing2020causal} \\
Path-integral methods &
R1--R3 and R5 for baseline and path &
R6, R7, R8, R10 &
\cite{sundararajan2017axiomatic,erion2021expectedgradients,
kapishnikov2021guidedig,xu2020blurig,lundstrom2025ig} \\
Gradient and backpropagation methods &
R1, R2, R6 and the propagation rule &
R8, R9, R10 &
\cite{simonyan2013saliency,springenberg2015guidedbp,bach2015lrp,
shrikumar2017deeplift,ancona2018gradient,srinivas2019fullgrad} \\
CAM-style spatial methods &
R1, R2 and target layer / projection rule &
R8, R9, R10 &
\cite{zhou2016cam,selvaraju2017gradcam,chattopadhyay2018gradcampp,
wang2020scorecam,draelos2020hirescam,zheng2022shapcam} \\
Perturbation and surrogate methods &
R1, R2, R4 and R5 for perturbation design &
R7, R9, R10 &
\cite{ribeiro2016lime,ribeiro2018anchors,fong2017meaningful,
petsiuk2018rise,plumb2018maple,chang2019counterfactual,
schulz2020information} \\
Transformer and NLP attribution &
R1 and R2 for scalar target and token/span unit &
R8, R9, R10 &
\cite{arras2017rnnlrp,murdoch2018contextual,serrano2019attention,
jain2019attentionnot,abnar2020flow,chefer2021transformer,
deyoung2020eraser,jacovi2020faithfulness,bastings2022shortcuts} \\
\bottomrule
\end{tabularx}
\end{table*}

\subsection*{Use of the Crosswalk}

The crosswalk gives the reporting checklist a literature-backed
interpretation. For a new attribution study, the relevant method family
in Table~\ref{tab:family-reporting-profile} identifies the items that
must be specified before the attribution is mathematically defined. The
source anchors in Table~\ref{tab:checklist-crosswalk} then identify the
literature that explains why each item matters. For example, a
KernelSHAP study requires the scalar output (R1), feature grouping
(R2), the interventional or conditional value function (R4), the
coalition sampling design (R5), the approximation budget (R7), and the
correlated-feature failure mode (R10). An Integrated Gradients study
requires the scalar output (R1), baseline (R3), path and integration
budget (R5, R7), axioms used (R6), sanity-check behaviour (R8), and
baseline/path sensitivity (R10).

\bibliographystyle{IEEEtran}
\bibliography{bibliography}

\end{document}